\documentclass[9pt,twocolumn]{extarticle}
\usepackage[a4paper,top=17mm,bottom=18mm,left=15mm,right=15mm,columnsep=6mm]{geometry}
\usepackage[T1]{fontenc}\usepackage{amsmath,amssymb}\usepackage{newtxtext}\usepackage{newtxmath}
\usepackage[scaled=0.92]{helvet}
\usepackage{graphicx}\usepackage{xcolor}\usepackage{booktabs,array}\usepackage{placeins}
\newcolumntype{P}[1]{>{\raggedright\arraybackslash}p{#1}}
\usepackage[superscript,biblabel]{cite}
\usepackage[hidelinks]{hyperref}
\usepackage{caption}\DeclareCaptionLabelSeparator{pipe}{ $|$ }
\usepackage{titlesec}
\titleformat{\section}{\sffamily\bfseries\fontsize{11}{13}\selectfont}{}{0pt}{}
\titlespacing*{\section}{0pt}{9pt}{3pt}
\titleformat{\subsection}{\sffamily\bfseries\fontsize{9}{11}\selectfont}{}{0pt}{}
\titlespacing*{\subsection}{0pt}{7pt}{2pt}
\titleformat{\paragraph}[runin]{\sffamily\bfseries\fontsize{8.5}{10}\selectfont}{}{0pt}{}
\makeatletter
\renewenvironment{figure}[1][tp]{\@dblfloat{figure}[tp]}{\end@dblfloat}
\renewenvironment{table}[1][tp]{\@dblfloat{table}[tp]}{\end@dblfloat}
\renewcommand\@biblabel[1]{#1.}
\makeatother
\graphicspath{{figs/}}

\newenvironment{plainsays}{\par\medskip\begin{list}{}{\leftmargin=2em\rightmargin=2em}\item\relax}{\end{list}\medskip}
\newtheorem{theorem}{Theorem}
\newtheorem{corollary}{Corollary}
\newtheorem{proposition}{Proposition}
\usepackage{enumitem}\setlist{leftmargin=*,itemsep=1pt,topsep=2pt,parsep=0pt}
\usepackage{etoolbox}\AtBeginEnvironment{table}{\sffamily\footnotesize}
\usepackage{microtype}\usepackage{multicol}
\renewcommand{\FloatBarrier}{}
\begin{document}
\twocolumn[{%
\noindent{\sffamily\footnotesize\bfseries\color[HTML]{C00000} Article}\par\vspace{5mm}
\noindent{\sffamily\bfseries\fontsize{19}{23}\selectfont\raggedright\hyphenpenalty=10000 Reproducible quantum clustering for single-cell transcriptome atlases\par}\vspace{4mm}
\noindent{\sffamily\fontsize{10}{12}\selectfont Henry Han}\par\vspace{1mm}
\noindent{\sffamily\scriptsize Data Science and AI Innovation Laboratory, Baylor University, Waco, Texas 76798, USA.\quad e-mail: Henry\_Han@baylor.edu}\par\vspace{5mm}
\noindent\begin{minipage}{\textwidth}\bfseries\fontsize{9.5}{12.2}\selectfont Current single-cell clustering pipelines discard most genes and can yield inconsistent clusters across random seeds. We introduce KQE-2, a training-free hybrid quantum--classical framework for reproducible whole-transcriptome clustering. KQE-2 maps cells into entanglement-free quantum states, enabling exact feature evaluation on classical hardware and direct checks on a quantum processor. Theoretically, we prove that removing entanglement is necessary and sufficient to bound single-qubit feature variance locally. Practically, while retraining scVI shifted 14\% of cells and altered rare-type markers and drug targets, KQE-2 delivered identical biology across runs. On standard highly variable genes, the quantum encoding alone raised accuracy on a developing cortex from 0.61 to 0.79. KQE-2 scales to 11.4 million nuclei on one server, outperforming the compared methods on eight of ten complete atlases (and within the best method's run-to-run range on two), and leads on six even under standard downstream clustering.\end{minipage}\par\vspace{6mm}
}]

Single-cell atlases now span millions of cells, and clustering is the foundation for reading them \cite{regev2017,tabula2022,cellxgene2025}. Every downstream discovery, from cell types and marker genes to pathways and drug targets, is read from these clusters. Yet standard workflows discard most of the transcriptome upfront: scanpy and Seurat retain about 2,000 highly variable genes (HVGs), compress them by principal component analysis (PCA) and partition a nearest-neighbor graph \cite{wolf2018,hao2024}. Deep generative (scVI) and foundation (scGPT) models embed restricted gene sets after GPU training \cite{lopez2018,cui2024}, while dedicated clustering algorithms have been applied to only tens of thousands of cells \cite{kim2025scice,wang2023scdcca,vinh2025scagcl,an2024scdac}.

\textit{What atlases need from clustering.} Reference atlases demand more than a high summary score against an annotation. Standard metrics such as adjusted mutual information (AMI) \cite{vinh2010} are dominated by abundant types: a method can score well while swallowing a rare type into a neighbor, erasing that population and its markers. Worse, partitions that shift with the training seed alter the resulting markers, pathways and drug targets, making the biology irreproducible across laboratories. Extended Data Table~1 defines nine requirements for trustworthy atlas clustering; KQE-2 meets all nine.

\textit{The problem.} Can we cluster a million-cell atlas across its whole transcriptome, every measured gene, so that the same data always yield the same clusters and biology? Two obstacles stand in the way. First, HVG selection removes most genes upfront. Favoring variance across abundant populations, it systematically discards markers of rare types ($<1\%$ of cells) \cite{zhao2025hvg,wang2025genecover,dollinger2025}. Yet modern atlases hold 15 to 251 times more cells than genes, a regime in which the full transcriptome can be reliably estimated \cite{vershynin2012}; discarding genes discards biology. Second, learned embeddings can drift across runs: stochastic gradient descent reaches different local optima across seeds, scGPT samples 1,200 genes per cell at inference, and GPU floating-point accumulation is non-deterministic. Graph workflows are deterministic under a fixed seed, but can fracture abundant, homogeneous types along technical axes \cite{grabski2023}.

\textit{Our idea.} We encode the whole transcriptome into entanglement-free quantum states. Quantum feature maps embed data in exponentially large Hilbert spaces \cite{schuld2018,perezsalinas2020,havlicek2019,bose2026}, but entangling gates make exact evaluation \#P-hard (intractable in general) \cite{valiant1979,aaronson2013} and leave hardware estimates vulnerable to shot noise and drift \cite{schuld2022goal,bowles2024}. Without entanglement, every quantum feature reduces to an exact closed form: bit-for-bit identical on every run, computable for millions of cells on one GPU and checkable on a quantum processor. Requiring no training, this encoding draws its resolving power from the whole transcriptome and bounded quantum-state geometry.

\textit{What is new.} Existing quantum kernel methods rely on sampled, entangled maps \cite{havlicek2019}, and prior quantum single-cell tools clustered around 1,000 cells \cite{kubacki2023,wang2025qbm}; classical HVG-free methods binarize detection or fit sparse Bayesian models \cite{qiu2020,nwizu2026}. KQE-2 (kernel quantum encoding) evaluates every state exactly, retains the whole transcriptome and, to our knowledge, is the first quantum framework to cluster complete, multi-million-cell atlases reproducibly.

KQE-2 meets all nine requirements, ordered by importance:
\begin{enumerate}[leftmargin=*,itemsep=2pt,topsep=3pt]
\item \textit{Bit-for-bit reproducible biology.} KQE-2 returns identical clusters, markers, pathways and drug targets on every rerun, and gave identical clusters across servers for all 611,034 nuclei of a cerebellum atlas; retraining scVI with a new seed moved 14\% of cells.
\item \textit{Rare types and subtypes in imbalanced atlases.} KQE-2 recovers rare populations more accurately than scanpy, Seurat, scVI and scGPT (stratified Wilcoxon test, $P \le 0.043$; $P \le 2.1 \times 10^{-8}$ against the first three) and isolates the notochord, Purkinje cells and endocardium in an 11.4-million-nucleus embryo atlas (F1 of 0.85 to 0.90).
\item \textit{Full-scale coverage with leading accuracy.} KQE-2 scales to human, mouse and zebrafish atlases of up to 11.4 million nuclei, the only method to partition the largest, leading on eight of ten complete atlases and within the best method's run-to-run range on two.
\item \textit{Whole-transcriptome resolution.} KQE-2 retains every gene (switching to 5,000 HVGs, by a label-free rule based on data entropy and the requested number of clusters, on only three of ten atlases) and rescues types whose markers HVG selection strips, raising the share of lens fiber cells in their primary cluster from 0.40 to 1.00.
\item \textit{Zero training.} KQE-2 requires neither stochastic gradient descent nor a pretraining corpus.
\item \textit{Theoretical guarantees.} We prove that removing entanglement is necessary and sufficient to keep single-qubit feature variance local against data and hardware noise. When two randomized decompositions yielded different scores, the encoding preserved the clusters (ARI of 0.90 and 0.99, versus 0.43 and 0.81).
\item \textit{Exact, auditable quantum-state geometry.} Theorem~\ref{thm:state} certifies exact nearest quantum states for 888,263 nuclei in 100 s. KQE-2 clusters on this certified graph, which also gave the first exact quantum-state map of a complete 11.4-million-nucleus atlas.
\item \textit{Quantum hardware check.} On an IBM quantum processor, measured features across 380 cells matched exact values with a correlation of 0.978.
\item \textit{Deterministic donor correction.} Quantum alignment removes donor batch effects without random seeds, raising AMI on a 79-donor heart atlas from 0.92 to 0.96 (versus 0.81 for Harmony).
\end{enumerate}
Together, these results establish that an exact, entanglement-free quantum representation can cluster complete single-cell atlases of over ten million nuclei with leading accuracy on most complete atlases and bit-for-bit reproducibility.
\begin{figure}[t]
\centering
\includegraphics[width=\textwidth]{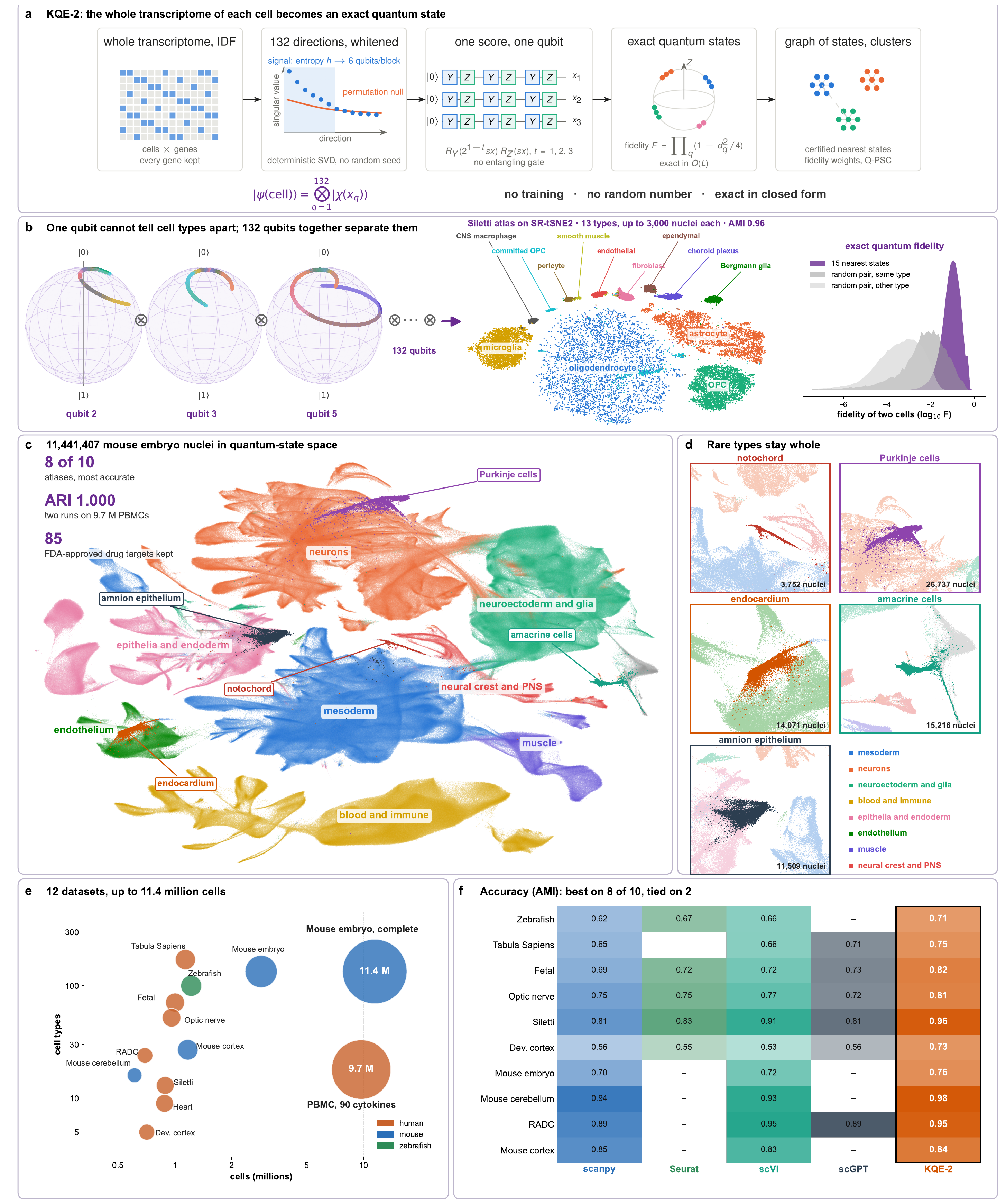}
\caption{\textbf{KQE-2 turns the whole transcriptome of each cell into an exact quantum state and clusters complete atlases.} \textbf{a}, The pipeline. Counts of all genes are weighted by inverse document frequency (IDF). A deterministic singular value decomposition (SVD) with no random seed gives 132 directions; a gene-permutation null separates signal from noise and sets the number of qubits, and the scores are scaled to unit variance (whitened). Each score rotates its own qubit through three layers of $R_Y$ and $R_Z$ rotations with no entangling gate, so each cell becomes a product of exact single-qubit states and the fidelity of two cells is a product over qubits. Quantum precision shrinkage clustering (Q-PSC) joins each cell to its 15 certified nearest states, weights the edges by fidelity, divides the graph into about $1.5k$ clusters with Leiden and merges them to the $k$ annotated types. \textbf{b}, Siletti non-neuronal brain atlas. Left, Bloch vectors of three qubits: no single qubit separates the types. Middle, SR-tSNE2 map of the quantum features of up to 3,000 nuclei per type (13 types; AMI 0.96, Table~\ref{tab:main}): the 132 qubits together separate them. Right, exact fidelity between each nucleus and its 15 nearest states, and between random pairs of nuclei of the same or of different types. \textbf{c}, All 11,441,407 nuclei of the complete mouse embryo atlas, mapped from their certified 15-nearest-state graph and colored by eight lineages merged from the author annotations. \textbf{d}, On the same map, the notochord, Purkinje cells, endocardium, amacrine cells and amnion epithelium form islands of nearby quantum states. \textbf{e}, The 12 datasets of this study, up to 11.4 million cells; the 2.86-million-nucleus mouse embryo subset and the complete atlas are plotted separately. \textbf{f}, AMI on the ten complete atlases (Table~\ref{tab:main}): KQE-2 is the most accurate on eight and within the run-to-run range of the best method on the other two; dash, not run or not applicable.}
\label{fig:overview}
\end{figure}

\section*{Results}
\subsection*{KQE-2 maps every cell to an exact quantum state}
KQE-2 turns the whole transcriptome of each cell into an exact quantum state in four deterministic steps (Fig.~\ref{fig:overview}). First, the log-transformed counts of the whole transcriptome are weighted by inverse document frequency (IDF), which raises the weight of genes detected in few cells, the signature of a marker. Second, a deterministic singular value decomposition (SVD), started from a fixed vector, gives 132 directions with no random step. A null model that permutes each gene across cells separates the signal directions from noise. Their data entropy, the Shannon entropy of their normalized singular values \cite{han2024is}, measures how many independent expression programs they hold. Rounded up, it is five to seven across the atlases, and KQE-2 uses six qubits per block (Fig.~\ref{fig:quantum}c--e). Third, each of the 132 scores, scaled to unit variance (whitened), rotates its own qubit in three layers that each load the score again (re-uploading layers), with no entangling gate. Each cell then has 726 quantum features: the Bloch vector of each qubit, made of the three expectations that fully describe its state, and the pairwise correlations within each block. Fourth, the cells are clustered on the graph of their quantum states by quantum precision shrinkage clustering (Q-PSC; Methods). Each cell is joined to its 15 nearest quantum states, found exactly and certified by Theorem~\ref{thm:state}, with edges weighted by the state fidelity. The Leiden algorithm first divides this graph into $\mathrm{round}(1.5k)$ clusters, and the two clusters with the largest summed fidelity relative to their size are merged repeatedly until $k$ remain, the number of annotated types.

A gene-set rule, also based on data entropy, decides the input before any clustering; it uses the requested number of clusters $k$ and no label. KQE-2 keeps the whole transcriptome unless the atlas holds many more expression programs than requested clusters, in which case it uses 5,000 HVGs (Methods). A large excess means that most of the additional programs describe variation within the requested types, as in the developing cortex, annotated with five to eight broad classes; there, 5,000 HVGs focus the states on the differences between types. Labels are never used to build the graph; they set only the number of clusters $k$ and score the result.

\textit{Four guarantees.} Four theorems (Methods) set the number of qubits per block (Theorem~\ref{thm:qubits}), bound the variance of each feature independently of the number of qubits (Theorem~\ref{thm:var}), show that removing entanglement is necessary and sufficient for this bound on single-qubit features (Theorem~\ref{thm:local}) and bound feature distances by state distances (Theorem~\ref{thm:state}).

\subsection*{The quantum encoding resolves cell types that the whitened scores merge}
\textit{Gain on the standard input.} The quantum encoding raises accuracy even when nothing else in a standard pipeline changes. With the 2,000 HVGs and principal components of scanpy held fixed, adding the encoding raised the AMI with the annotated types from 0.605 to 0.788 on the complete developing human cortex. On the brain atlas of the Rush Alzheimer's Disease Center (RADC), it raised the AMI from 0.887 to 0.931.

\textit{Gain on the scores of KQE-2.} The gain is larger on the 132 whitened scores of KQE-2 (Extended Data Table~2). On the 146,598-cell subsample of the developing cortex (five types), clustering these scores directly puts 92\% of the cells in one cluster (AMI 0.18 to 0.20). Encoding the same scores resolves all five types (AMI 0.766 over three Leiden seeds). It also raises the F1 of oligodendrocyte precursor cells (OPCs) from 0.45 under scanpy to 0.96. (F1, the harmonic mean of precision and recall in the cluster holding most of a type's cells) On the Siletti non-neuronal brain atlas, the encoding gives the OPCs their own cluster (AMI 0.935, against 0.815 for the whitened scores).

\textit{Gain on a complete atlas.} The gain holds on complete atlases. On the complete developing cortex (693,544 nuclei, eight clusters, 5,000 HVGs by the rule), Leiden at $k$ on the 15-nearest-neighbor graph of the whitened scores gave an AMI of 0.580, and the same graph built on the quantum features of the same scores gave 0.731.

\textit{The gain is specific to the quantum encoding.} None of the five families of classical feature maps tested on the same 132 scores reached the accuracy of the encoding. The gap was widest on the developing cortex, where one type holds 78\% of the cells. There, random Fourier features, trigonometric, quadratic, $\tanh$ and clipping maps gave 0.03 to 0.53, against 0.766 for the encoding (Extended Data Fig.~1; class imbalance, Supplementary Note~1).

\subsection*{Clusters form in exact quantum-state space}
\begin{figure}[p]
\centering
\includegraphics[width=\textwidth,height=0.775\textheight,keepaspectratio]{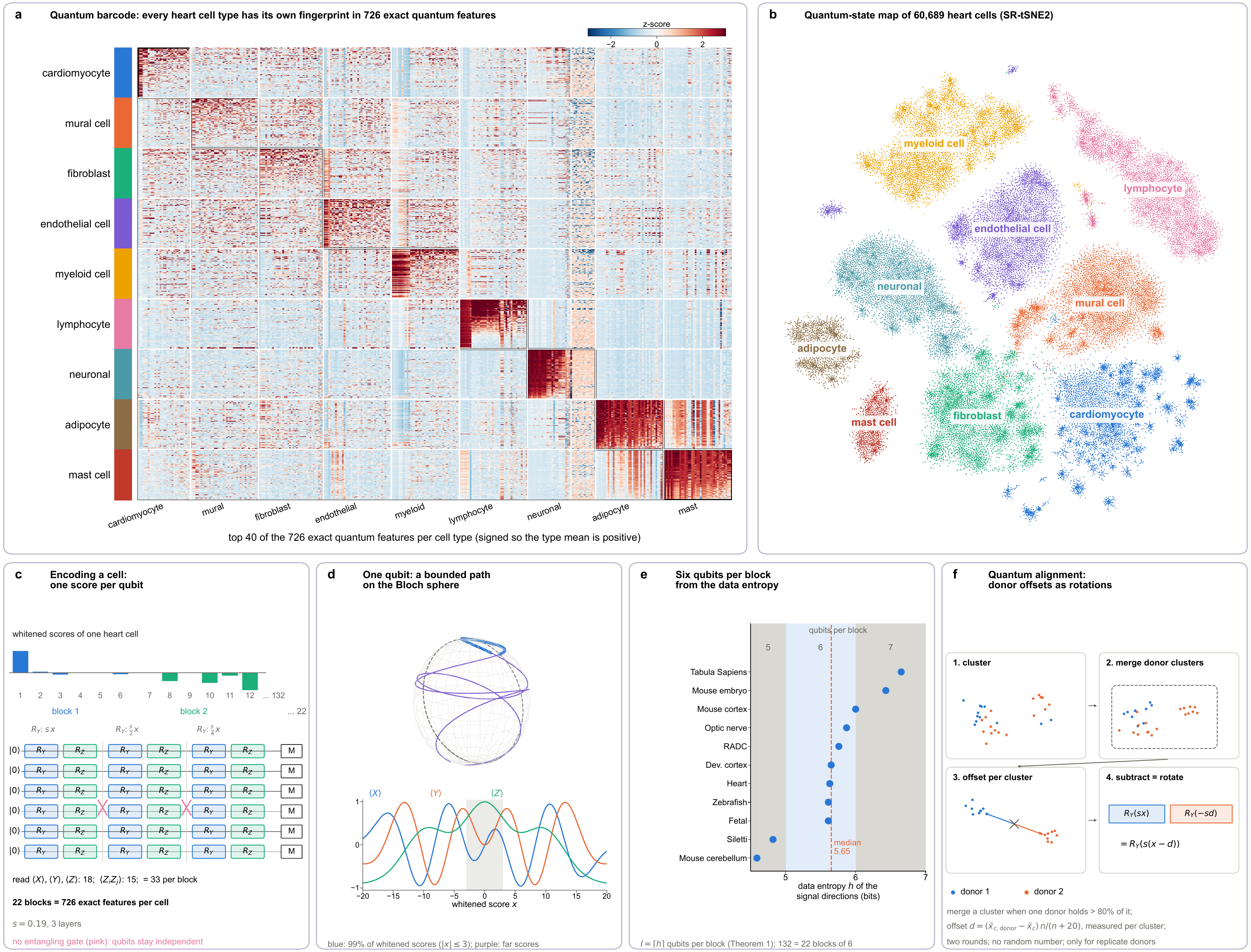}
\caption{\textbf{Each cell type has its own quantum barcode.} \textbf{a}, Quantum barcodes of the nine cell types of the cardiomyopathy heart atlas: for each type, the 40 of the 726 exact quantum features that mark it most, signed so that the type mean is positive (rows, cells; columns, features). \textbf{b}, SR-tSNE2 map of the quantum features of 60,689 heart cells, in which the nine types separate. \textbf{c}, The 132 whitened scores of a cell fill 22 blocks of six qubits. Each score rotates its own qubit by $R_Y(2^{-(t-1)}s\,x)$ and $R_Z(s\,x)$, $t = 1, 2, 3$, $s = 0.19$, with no entangling gate, giving 726 exact features per cell. \textbf{d}, The Bloch vector of one qubit traces a bounded path as its score grows, so one extreme score moves a cell by a bounded distance (Proposition~\ref{prop:sat}); blue, the 99\% of whitened scores with $|x| \le 3$. \textbf{e}, The data entropy of the signal directions of the 11 atlases sets six qubits per block at the median (Theorem~\ref{thm:qubits}). \textbf{f}, Quantum alignment removes donor offsets as donor-specific rotations, with no random number.}
\label{fig:quantum}
\end{figure}

The encoding turns the Leiden graph into a graph of quantum states. Without entanglement, the exact fidelity of two cells, the squared overlap of their states, is a product over qubits and costs linear time (Theorem~\ref{thm:state}). The same theorem bounds the feature distance above and below by the state distance, and Supplementary Note~2 works an example. Across 80,000 pairs of 20,000 Siletti cells, every pair satisfied the lower bound, and the feature distance and the exact state distance had a Spearman correlation of 0.9998. On the human heart atlas, each of the nine types has its own barcode across the 726 quantum features (Fig.~\ref{fig:quantum}a,b).

\textit{A complete atlas in quantum-state space.} The same geometry holds for every nucleus of a complete atlas (Fig.~\ref{fig:overview}b--d; Supplementary Fig.~2). For each of the 888,263 nuclei of the Siletti atlas, we found its 15 nearest neighbors under the exact state distance $-\log F$ over 132 qubits. With candidates taken by feature distance, for 888,146 nuclei (99.99\%) the lower bound of Theorem~\ref{thm:state} proves that no other nucleus can be closer; the remaining 117 nuclei were searched exhaustively, and the whole search took 100 s on one GPU (Methods). On average, 99.90\% of the 15 nearest states of a nucleus lie in its own KQE-2 cluster (Table~\ref{tab:main}). Nearest states have a median fidelity of 0.12, against 0.00053 for random nuclei of different clusters. This certified graph is the graph that Q-PSC clusters, so each KQE-2 cluster is a region of nearby quantum states. To our knowledge, this is the first map of a complete single-cell atlas in exact quantum-state space. The same search covered all 11,441,407 nuclei of the complete mouse embryo atlas in 3.1 h on one GPU, with 104 nuclei searched exhaustively. There, 93\% of the 15 nearest quantum states of a nucleus share its annotated type, and the notochord, Purkinje cells, endocardium, amacrine cells and amnion epithelium form islands of their own, with 83\% to 94\% of their nearest states in the same type (Fig.~\ref{fig:overview}c,d; Supplementary Fig.~2e,f).

\subsection*{KQE-2 leads on eight of ten complete atlases}
\textit{Benchmark.} KQE-2 reached the highest accuracy on most complete atlases. We compared KQE-2 with scanpy \cite{wolf2018}, Seurat \cite{hao2024}, scVI \cite{lopez2018} and the foundation model scGPT \cite{cui2024} on ten complete atlases of 611,034 to 2,860,465 cells \cite{duran2025,tabula2022,cao2020,opticnerve2025,siletti2023,velmeshev2023}. Every other method used the same graph construction and Leiden protocol, and KQE-2 is reported both with this protocol and in its full configuration with Q-PSC (Table~\ref{tab:main}). KQE-2 was the most accurate method on eight atlases and within the range of the best method's runs on the other two. With exactly the downstream protocol of the other methods, Leiden at $k$ on the feature graph, KQE-2 was the most accurate on six atlases and within 0.004 of the best method on three more (Table~\ref{tab:main}, column KQE-2), so the gain does not depend on the final clustering step. Its margin over the best other method ranged from 0.032 on the mouse cerebellum to 0.169 on the developing cortex. KQE-2 was also more accurate on the three atlases that scGPT may have seen in pretraining (Supplementary Note~3).

\textit{Atlases run after the gene-set rule was fixed.} Three atlases were run after the gene-set rule had been fixed: its threshold, and for each atlas the value of $\Delta$ and the chosen gene set, were recorded with a time stamp before the first run (Methods). On 2,860,465 nuclei of the mouse embryo from gastrulation to birth \cite{qiu2024embryo}, KQE-2 was the only method that partitioned the cells into the 134 annotated types, and it reached 0.761, against 0.717 for scVI and 0.703 for scanpy. On the mouse cerebellum (611,034 nuclei, 78\% granule cells) \cite{kozareva2021cereb}, KQE-2 reached 0.977, against 0.945 for scanpy. On the third, the mouse cortex \cite{yao2021cortex}, KQE-2 was within the range of the best method's seeds or runs, as it was on the RADC atlas.

\textit{Contribution of Q-PSC.} Both steps of Q-PSC add accuracy. The quantum-state graph alone raised Leiden at $k$ above the feature graph on the Siletti atlas (0.925 to 0.956), the mouse cerebellum (0.918 to 0.977) and the optic nerve (0.786 to 0.806). Merging then raised the AMI on six of the seven whole-transcriptome atlases, most on the fetal atlas (0.743 to 0.821) and the mouse embryo (0.722 to 0.761), and kept it on the mouse cerebellum (0.977). Merging joins the pieces of an abundant type, while rare types whose states are distinct keep their own clusters. On the three atlases for which the gene-set rule selects 5,000 HVGs, KQE-2 applies Leiden at $k$ directly (Supplementary Table~10). The merging step, adapted to the graphs of the other methods, is analysed in Supplementary Note~12. The accuracy of KQE-2 therefore comes from the quantum framework as a whole: the whole-transcriptome input, the exact quantum-state geometry and the merge, every step deterministic.

\textit{Statistical significance across atlases and rare types.} Two paired tests show that the gains are systematic (Extended Data Table~3). Across the ten atlases, KQE-2 exceeded the best other method by 0.050 AMI on average (median 0.042; bootstrap 95\% confidence interval 0.025 to 0.082; one-sided Wilcoxon signed-rank test, $P = 0.0049$). It also exceeded each method separately ($P = 0.0020$ to $0.031$). On rare types, the F1 of KQE-2 was higher than that of scanpy, Seurat, scVI and scGPT (332, 183, 480 and 260 rare types; mean gain 0.127, 0.110, 0.054 and 0.039; Wilcoxon signed-rank test stratified by atlas, $P = 1.1 \times 10^{-26}$, $1.1 \times 10^{-14}$, $2.1 \times 10^{-8}$ and 0.043) (Extended Data Table~3b). On all seven whole-transcriptome atlases, Leiden at $k$ on the quantum-state graph, before merging, already exceeded every other method.
\begin{table}[t]
\centering\small
\caption{\textbf{KQE-2 is the most accurate method on eight of ten complete atlases and within the range of the best method's runs on the other two.} AMI at the annotated number of types (Leiden seed 0); every other method and the column KQE-2 use the same graph construction and scoring. scVI and scGPT, mean of three runs (lowest in parentheses). KQE-2, Leiden at $k$ on the 15-nearest-neighbor graph of its quantum features, the same downstream protocol as every other method; KQE-2 + Q-PSC, clustering on the certified quantum-state graph (Methods). The gene-set rule fixes the configuration of KQE-2: Q-PSC on whole-transcriptome atlases, and Leiden at $k$ on 5,000 HVGs for the developing cortex, RADC and mouse cortex atlases. Margin, this configuration minus the best other method; w.r., within the range of the best method's seeds or runs (not a formal equivalence test). The developing cortex is scored at eight clusters. Bold, best per row. n.r., not run: $^{a}$the count matrix exceeds the $2^{31}$ nonzero entries that Seurat reads; $^{b}$Seurat was run on the first six atlases. $^{c}$At most 95 clusters within the resolution range. $^{d}$One training run, at most 112 clusters. n.a., not applicable (the scGPT checkpoint is for human data); n.u., not used, because the gene-set rule selects 5,000 HVGs.}
\label{tab:main}
\setlength{\tabcolsep}{3pt}
\resizebox{\textwidth}{!}{\begin{tabular}{lrcccccccr}
\toprule
Atlas & cells & scanpy & Seurat & scVI & scGPT & KQE-2 & KQE-2 + Q-PSC & Margin \\
\midrule
Zebrafish & 1,120,336 & 0.620 & 0.671 & 0.655 (0.654) & n.a. & 0.706 & \textbf{0.711} & $+0.039$ \\
Tabula Sapiens & 1,136,180 & 0.645 & n.r.$^{a}$ & 0.658 (0.655) & 0.708 (0.705) & 0.725 & \textbf{0.750} & $+0.042$ \\
Fetal & 978,128 & 0.686 & 0.716 & 0.717 (0.716) & 0.733 (0.729) & 0.750 & \textbf{0.821} & $+0.088$ \\
Optic nerve & 959,629 & 0.753 & 0.749 & 0.769 (0.757) & 0.721 (0.713) & 0.786 & \textbf{0.810} & $+0.041$ \\
Siletti non-neuronal & 888,263 & 0.807 & 0.832 & 0.908 (0.847) & 0.810 (0.777) & 0.925 & \textbf{0.963} & $+0.055$ \\
Developing cortex & 693,544 & 0.557 & 0.546 & 0.528 (0.515) & 0.561 (0.524) & \textbf{0.730} & n.u. & $+0.169$ \\
\midrule
Mouse embryo & 2,860,465 & 0.703$^{c}$ & n.r.$^{b}$ & 0.717$^{d}$ & n.a. & 0.717 & \textbf{0.761} & $+0.044$ \\
Mouse cerebellum & 611,034 & 0.945 & n.r.$^{b}$ & 0.926 (0.851) & n.a. & 0.918 & \textbf{0.977} & $+0.032$ \\
RADC brain & 693,677 & 0.886 & n.r.$^{b}$ & \textbf{0.950} (0.945) & 0.891 (0.857) & 0.946 & n.u. & $-0.004$ (w.r.) \\
Mouse cortex & 1,169,213 & \textbf{0.846} & n.r.$^{b}$ & 0.833 (0.825) & n.a. & 0.844 & n.u. & $-0.002$ (w.r.) \\
\bottomrule
\end{tabular}}
\end{table}

\begin{figure}[t]
\centering
\includegraphics[width=\textwidth]{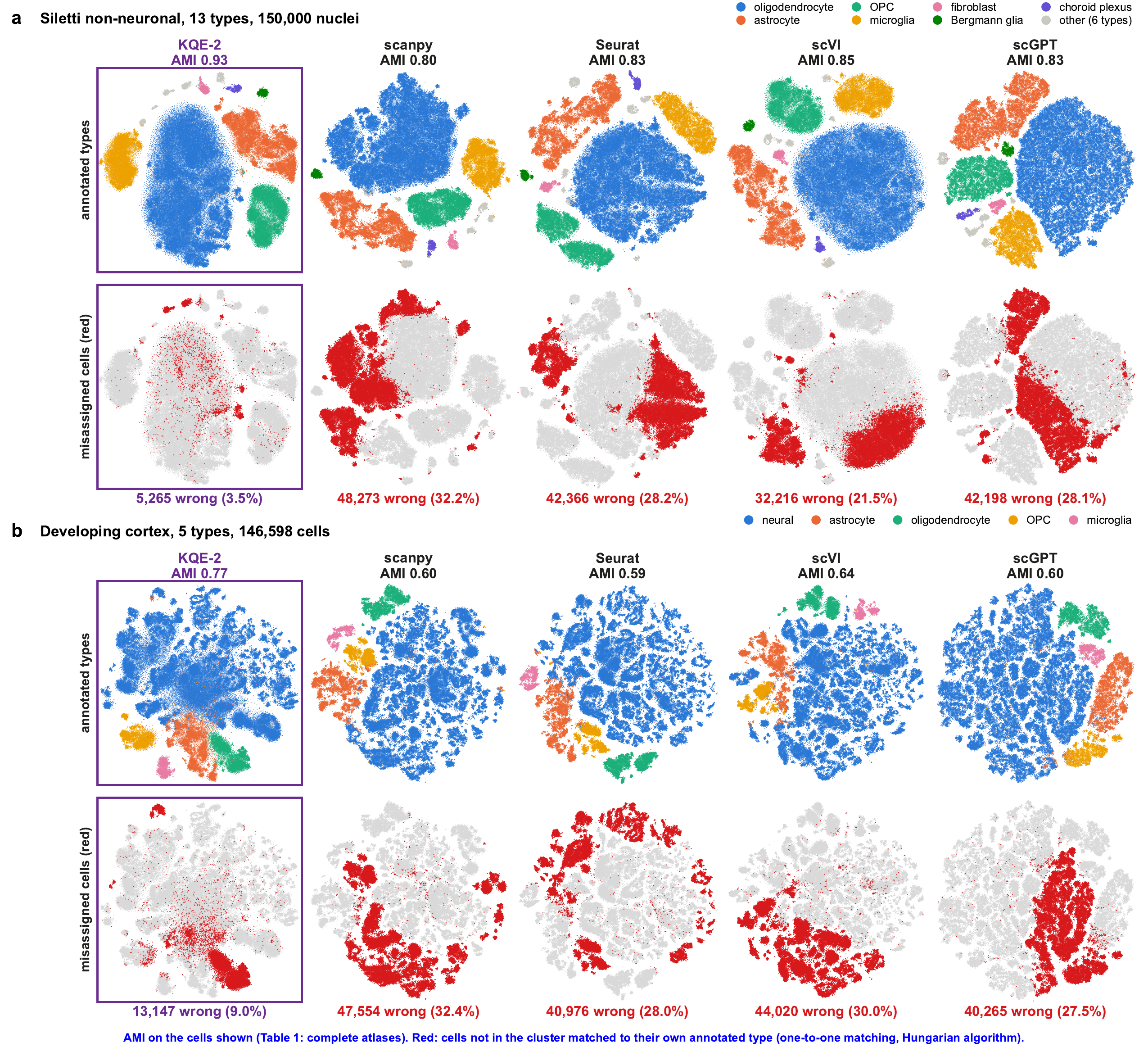}
\caption{\textbf{KQE-2 keeps abundant types whole and misassigns the fewest cells.} Five embeddings of the same cells, each clustered by Leiden at the annotated $k$ (seed 0) and drawn as an SR-tSNE2 map (Methods). \textbf{a}, Siletti non-neuronal brain atlas: 150,000 nuclei, 13 types. \textbf{b}, Developing cortex: 146,598 cells, five types. Top rows, cells colored by annotated type; titles give the AMI on these cells (Table~\ref{tab:main} reports the complete atlases). Bottom rows, misassigned cells in red: cells outside the cluster paired with their own type when clusters and types are paired one to one (Hungarian algorithm). KQE-2 misassigns 3.5\% and 9.0\% of the cells; scanpy, Seurat, scVI and scGPT misassign 21.5\% to 32.2\% and 27.5\% to 32.4\%.}
\label{fig:umap}
\end{figure}

\textit{Why KQE-2 leads.} KQE-2 keeps abundant types whole while it separates OPCs (Fig.~\ref{fig:umap}). On the complete Siletti atlas, scanpy, Seurat and scGPT divide the mature oligodendrocytes, 55\% of the cells, into two or three clusters. KQE-2 keeps 99.6\% of them in one cluster and gives the OPCs their own. When clusters are paired one to one with annotated types, KQE-2 misassigns 3.5\% of 150,000 Siletti nuclei and 9.0\% of 146,598 developing cortex cells; the other methods misassign 21.5\% to 32.2\% and 27.5\% to 32.4\% (Fig.~\ref{fig:umap}). The split of the other methods follows technical quality: in scanpy, the second oligodendrocyte cluster has 7.6 times the mitochondrial read fraction of the largest one (Extended Data Fig.~2). Every KQE-2 cluster carries the canonical markers of its type (Extended Data Fig.~3). An ablation shows that whitening and the encoding together prevent the split (Extended Data Table~4; Supplementary Note~4).

\textit{Scale without training.} On one GPU, it embedded all 1,120,336 zebrafish cells on all 32,031 genes in 973 s with a peak memory of 13.2 GB. Four dedicated clustering methods did not complete 146,598 cells within 3 hours and 150 GB (Extended Data Tables~1 and 5; Supplementary Table~12; Secuer, Supplementary Note~10). The largest run used the complete mouse embryo atlas \cite{qiu2024embryo}, 11,441,407 nuclei of 134 types and four times the subset of Table~\ref{tab:main}; its gene-set rule was recorded before the run. On one server (40 CPU cores, 502 GB of memory, one NVIDIA RTX 6000 Ada GPU), the deterministic decomposition of the whole transcriptome took 11.9 h and was bit-for-bit identical on rerun, the certified nearest-state graph took 3.1 h on one GPU, and Q-PSC took 12 min. KQE-2 partitioned the nuclei into the 134 annotated types with an AMI of 0.755, close to its accuracy on the 2.86-million-nucleus subset (0.761). On the same 502-GB server, scVI exceeded the memory, scanpy did not return an embedding and scGPT does not apply to mouse data, so KQE-2 was the only method that returned a clustering. To our knowledge, KQE-2 is the first method to cluster a complete atlas of 11.4 million nuclei on the whole transcriptome with bit-for-bit identical results.

\subsection*{Bit-for-bit reproducibility keeps markers and drug targets fixed}
\begin{figure}[t]
\centering
\includegraphics[width=\textwidth]{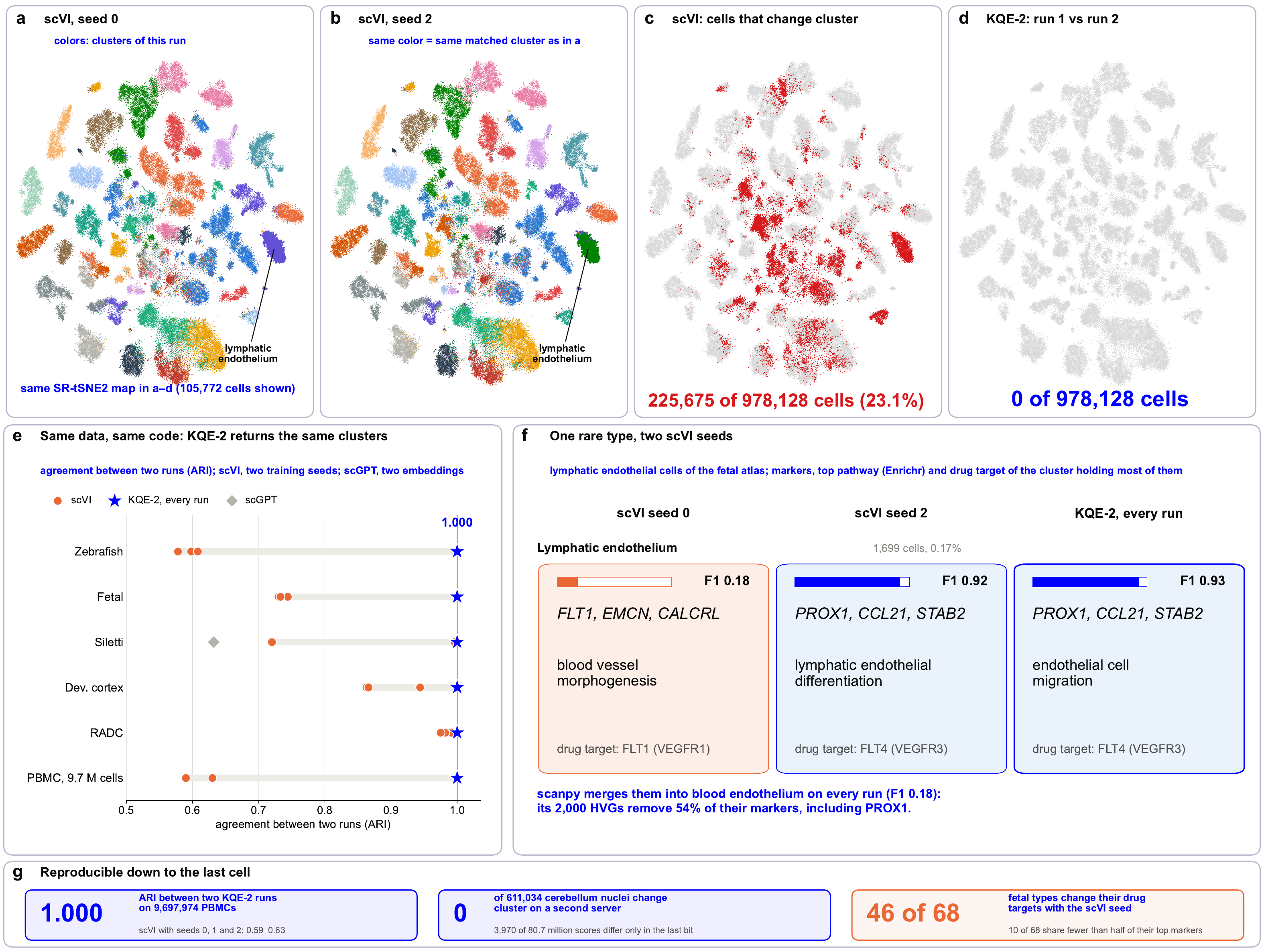}
\caption{\textbf{Same data, same code: KQE-2 returns the same clusters, markers and drug targets on every run.} \textbf{a}--\textbf{d}, One SR-tSNE2 map of 105,772 cells of the fetal atlas, colored four ways. \textbf{a}, scVI clusters with training seed 0. \textbf{b}, scVI clusters with seed 2; clusters paired one to one with those of seed 0 keep their color. The lymphatic endothelial cells move to another cluster. \textbf{c}, Cells that change cluster between the two seeds (red): 225,675 of the 978,128 cells of the complete atlas (23.1\%). \textbf{d}, KQE-2, run 1 against run 2: no cell changes cluster. \textbf{e}, ARI between the clusters of two runs on complete atlases: scVI with two training seeds (dots), two scGPT embeddings (diamond) and KQE-2 (star; ARI 1.000 on every atlas). \textbf{f}, The lymphatic endothelial cells of the fetal atlas (1,699 cells, 0.17\%) under scVI seeds 0 and 2 and under KQE-2: F1 of the cluster that holds most of them, with its markers, top Enrichr pathway and drug target. \textbf{g}, Two KQE-2 runs on 9,697,974 PBMCs agree at an ARI of 1.000 (three scVI seeds: 0.59 to 0.63); on a second server, all 611,034 nuclei of the mouse cerebellum atlas keep their cluster, although 3,970 of 80.7 million scores differ in the last bit; with the scVI seed, the drug targets of 46 of 68 fetal types change.}
\label{fig:repro}
\end{figure}

KQE-2 gives bit-for-bit identical results on every run on the same machine and software. On every complete atlas, the decomposition, the quantum features and the clusters were identical on rerun. On a second server with another software environment, the scores of the 611,034-nucleus mouse cerebellum atlas differed only in the last bit of 3,970 of 80.7 million entries, and the certified nearest states and the Q-PSC clusters of all nuclei were identical (Fig.~\ref{fig:repro}g). The intermediate scores therefore need not be bit-identical across machines for the partition to be. Corollary~\ref{cor:repro}(c) shows why for the feature graph: the $K$ nearest neighbors of a cell can change only if its neighbor margin $\delta_i$ (Methods) is below $4\Lambda \max_j \|\xi_j\|$, and a last-bit difference $\xi$ makes this bound of the order of the floating-point precision. The bounds of Theorems~\ref{thm:var} and \ref{thm:local} held in every perturbation and hardware test (Supplementary Note~5).

\textit{Stability under resampling and sequencing depth.} We clustered three random 80\% subsamples of the fetal and Siletti atlases, and both atlases with every count halved by binomial thinning, with the complete pipelines of KQE-2 and scanpy (Supplementary Note~11). On the cells shared by two subsamples, the KQE-2 partitions agreed at ARIs of 0.94 to 0.97 on the fetal atlas and 0.999 on the Siletti atlas, against 0.67 to 0.70 and 0.57 to 0.91 for scanpy. With half the reads, the fetal partition of KQE-2 agreed with its full-depth partition at an ARI of 0.90 (AMI 0.784), whereas that of scanpy fell to 0.45 (AMI 0.587). The quantum encoding contributes to this stability (Supplementary Note~11).

\textit{Trained models.} Trained models can change their clusters from run to run. On the complete Siletti atlas, scVI trained with seeds 0, 1 and 2 gave AMIs of 0.939, 0.938 and 0.847. Between seeds 0 and 2, 14\% of the cells changed cluster (adjusted Rand index, ARI, 0.72). On the complete fetal atlas, 225,675 of 978,128 cells (23.1\%) changed cluster between scVI seeds 0 and 2, whereas not one cell changed cluster between two runs of KQE-2 (Fig.~\ref{fig:repro}a--d). Three scGPT embeddings of the Siletti atlas gave AMIs of 0.777 to 0.827, and the first two partitions agreed at an ARI of 0.63. A stable score can hide this drift. On the zebrafish atlas, three scVI runs scored 0.654 to 0.656, yet their partitions agreed at an ARI of 0.58 to 0.61 (Fig.~\ref{fig:repro}e). This instability persists at ten million cells. On 9,697,974 peripheral blood mononuclear cells of 12 donors stimulated with 90 cytokines (Methods), two runs of KQE-2 gave identical scores and identical clusters for every cell (ARI 1.000), whereas the partitions of scVI with seeds 0, 1 and 2 agreed at ARIs of 0.59 to 0.63.

\textit{Biological consequences of seed drift.} The drift changes the biology read from the clusters (Fig.~\ref{fig:repro}f). On the fetal atlas under scVI seed 0, the lymphatic endothelial cells (1,699 cells) merged into blood endothelium, in a cluster marked by the pan-endothelial genes \textit{FLT1} and \textit{EMCN} (F1 0.18). Under seed 2, they formed their own cluster (F1 0.92), marked by \textit{CCL21}, \textit{STAB2} and \textit{PROX1}, the master regulator of lymphatic identity \cite{wigle1999}. The drug targets among the markers differed between the two seeds in 46 of the 68 fetal types. KQE-2 gives the lymphatic endothelial cells their own cluster on every run (F1 0.93), marked by the same three genes. scanpy is deterministic with a fixed Leiden seed (ARI 1.000 between replicate runs on three atlases). Its HVG selection, however, removes 54\% of the markers of these cells, including \textit{PROX1}, and scanpy merges them into blood endothelium on every run (F1 0.18).

\textit{Deterministic donor correction.} Harmony \cite{korsunsky2019} depends on a random seed: on the RADC atlas \cite{lee2024radc}, three seeds gave somatostatin interneurons an F1 of 0.98, 0.98 and 0.64. Quantum alignment subtracts the offset of each donor inside clusters, a compensating rotation of each qubit (Fig.~\ref{fig:quantum}f; Methods). On the 79-donor heart atlas \cite{reichart2022}, it raised the AMI of KQE-2 from 0.92 to 0.96, against 0.81 for Harmony, and the F1 of adipocytes from 0.03 to 0.99.

\textit{Quantum hardware.} On an IBM processor, circuits of the encoding with added controlled-$Z$ chains, a harder test than the entanglement-free circuit, matched the exact values for 380 cells with a Pearson correlation of 0.978 (Supplementary Note~5).

\subsection*{The whole transcriptome recovers rare cell types and unannotated subtypes}
Clustering on the whole transcriptome recovers the rare types whose markers HVG selection removes. On the human fetal atlas \cite{cao2020}, 62\% of the top-50 markers of each annotated type fall outside the 2,000 HVGs on average (6\% to 98\% across types). Selecting 5,000 or 10,000 HVGs still removes 49\% and 39\%.

Even before the quantum encoding, the deterministic decomposition of all 45,676 genes of the fetal atlas raised the share of a type's cells in its main cluster (Extended Data Fig.~4). The share rose from 0.40 to 1.00 for lens fiber cells, from 0.34 to 0.91 for amacrine cells, from 0.20 to 0.75 for neuroendocrine cells and from 0.52 to 0.98 for photoreceptors. The types that gained most had more of their markers outside the HVGs (Extended Data Fig.~4b).

\textit{Rare embryonic types.} The same holds for rare types that shape the embryo. On the 2.86-million-nucleus mouse embryo subset, KQE-2 gave the notochord, Purkinje cells, amnion epithelium, endocardium and amacrine cells, each 0.03\% to 0.24\% of the nuclei, an F1 of 0.83 to 0.91, against 0.04 to 0.28 for scanpy and scVI. On the complete atlas of 11.4 million nuclei, the same five types formed their own KQE-2 clusters (F1 0.83 to 0.90) and their own islands in quantum-state space (Fig.~\ref{fig:overview}d), and canonical markers confirm their identity. In the notochord cluster, 96\% of the nuclei express \textit{T} (Brachyury) and 64\% \textit{Shh}, against 1\% outside, and \textit{T} is the top marker of the cluster.

\textit{Subtypes beyond the annotation.} KQE-2 also resolves known subtypes that atlas annotations do not name, and it resolves them across species. On the human fetal atlas, without labels, it divided the annotated GABAergic neurons into stable clusters that match numbered, unnamed subclusters the authors obtained by re-clustering each type (Jaccard index 0.79 to 0.94), and their markers identify them as striatal projection neurons (\textit{EBF1}, \textit{FOXP1}, \textit{PDE10A}), striosomal neurons (\textit{OPRM1}; 529 cells) and interneurons of caudal ganglionic eminence origin (\textit{ADARB2}). Among the methods compared, only KQE-2 resolved the striatal projection neurons, the caudal ganglionic eminence interneurons and the striosomal neurons (Supplementary Note~8). On the complete mouse embryo atlas, which only KQE-2 clustered, the same lineages appeared independently, among nuclei that even the finest annotation of the authors, 190 cell types, calls cortical GABAergic interneurons; none of the 190 names a striatal neuron. KQE-2 separated striatal projection neurons from an indirect-pathway cluster and from interneurons of medial and caudal ganglionic eminence origin.

\subsection*{The genes removed by HVG selection carry pathways and drug targets}
\begin{figure}[t]
\centering
\includegraphics[width=\textwidth]{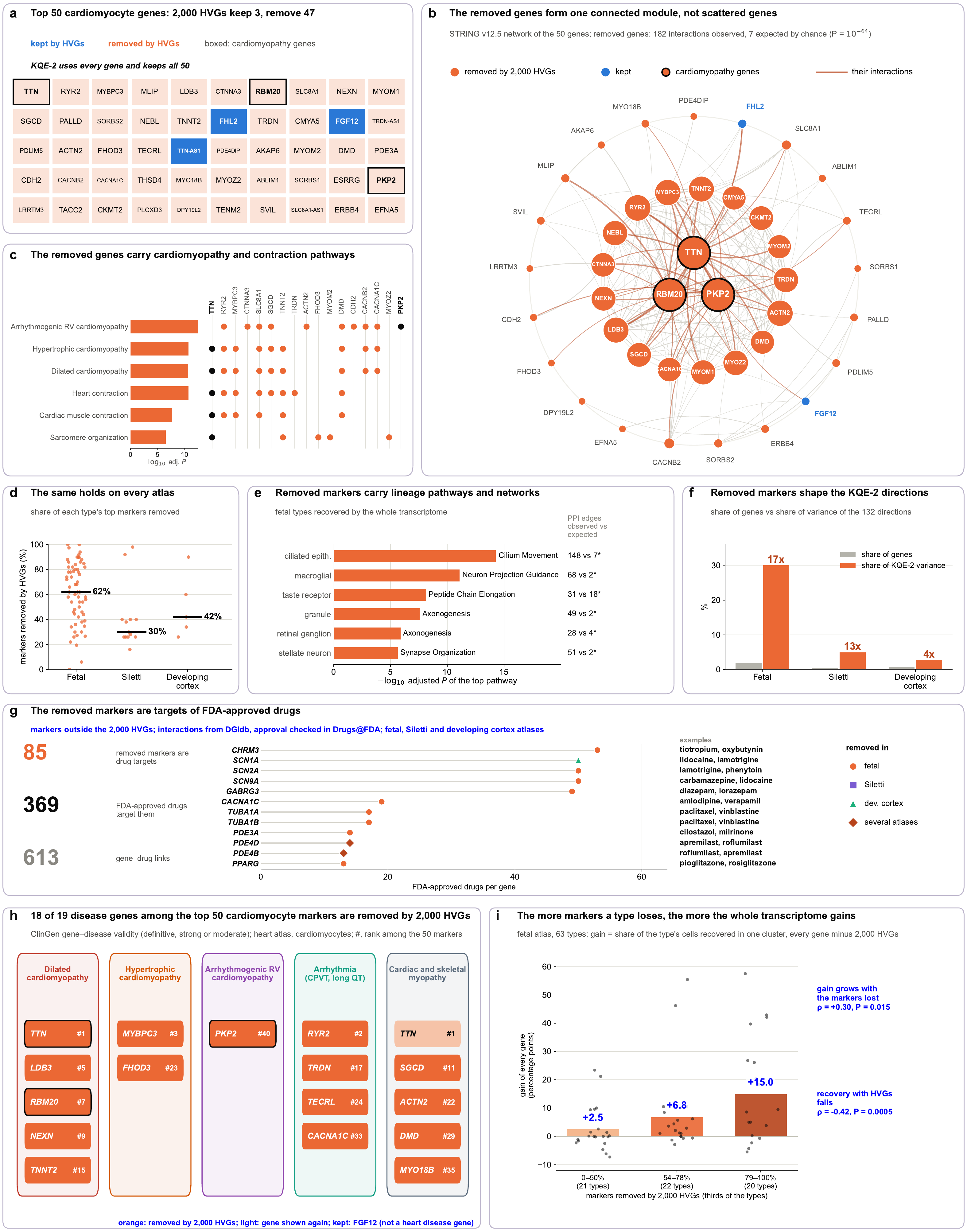}
\caption{\textbf{HVG selection removes disease genes, pathways, interaction networks and drug targets; KQE-2 keeps them.} Markers are the top-50 Wilcoxon markers of each type on the whole transcriptome; removed markers are those outside the 2,000 HVGs of scanpy. \textbf{a}, The 50 top markers of heart cardiomyocytes: 2,000 HVGs keep 3 and remove 47, whereas KQE-2 uses every gene and keeps all 50. Boxed, cardiomyopathy genes. \textbf{b}, STRING v12.5 network of the 50 genes: the removed genes form one connected module (182 interactions observed against 7 expected; $P = 10^{-64}$). \textbf{c}, Enrichr pathways of the removed genes: arrhythmogenic right ventricular, hypertrophic and dilated cardiomyopathy, heart and cardiac muscle contraction, and sarcomere organization; dots mark the genes in each pathway. \textbf{d}, Share of each type's markers removed on the fetal, Siletti and developing cortex atlases (150,000-cell subsamples; dots, types; bars, medians of 62\%, 30\% and 42\%). \textbf{e}, Top Enrichr pathway of the removed markers of fetal types that the whole transcriptome recovers, with observed and expected STRING edges (asterisk, $P < 0.05$). \textbf{f}, Share of the variance of the 132 KQE-2 directions carried by the removed markers, against their share of the genes. \textbf{g}, The removed markers of the three atlases include 85 targets of 369 FDA-approved drugs (613 gene--drug links; interactions from DGIdb, approval checked in Drugs@FDA); shown are the 12 genes with the most approved drugs, with examples. \textbf{h}, ClinGen gene--disease validity (definitive, strong or moderate): 2,000 HVGs remove 18 of the 19 disease genes among the 50 top cardiomyocyte markers; \#, rank among the 50. \textbf{i}, Fetal atlas, 63 types split into thirds by the share of markers removed. The more markers a type loses, the more it gains from the whole transcriptome (gain in the share of its cells recovered in one cluster, every gene minus 2,000 HVGs; Spearman $\rho = 0.30$, $P = 0.015$), and the less 2,000 HVGs recover it ($\rho = -0.42$, $P = 0.0005$). Bars, means; dots, types.}
\label{fig:bio}
\end{figure}

The genes that HVG selection removes carry core biology. Of the 50 top markers of heart cardiomyocytes, 2,000 HVGs keep only 3. The 47 they remove include the cardiomyopathy genes \textit{TTN}, \textit{RBM20} and \textit{PKP2} (Fig.~\ref{fig:bio}a), form one connected protein interaction module (STRING; 182 interactions observed against 7 expected; $P = 10^{-64}$) and carry the cardiomyopathy and contraction pathways (Fig.~\ref{fig:bio}b,c). They also include 18 of the 19 markers with a definitive, strong or moderate disease link in ClinGen (Fig.~\ref{fig:bio}h). KQE-2 uses every gene and keeps all 50. Across atlases, we tested the top-50 markers of every annotated type that fall outside the 2,000 HVGs of scanpy on the fetal, Siletti and developing cortex atlases. In total, HVG selection removes 830, 229 and 125 distinct marker genes on these atlases (Fig.~\ref{fig:bio}d).
\begin{itemize}[leftmargin=*,itemsep=1pt,topsep=2pt]
\item \textit{Pathways.} The removed markers are enriched for a pathway (Enrichr, adjusted $P < 0.05$) in 59 of 60 fetal types, 12 of 13 Siletti types and all five cortical types.
\item \textit{Interaction networks.} They form protein interaction networks denser than chance (STRING, $P < 0.05$) in 48 fetal, 9 Siletti and 4 cortical types.
\item \textit{Lineage functions.} In the fetal types that the whole transcriptome recovers, they carry lineage pathways, such as cilium movement in ciliated epithelial cells (Fig.~\ref{fig:bio}e).
\end{itemize}
These genes shape the quantum states of KQE-2. On the fetal atlas, the markers removed by 2,000 HVGs are 1.8\% of the whole transcriptome, yet they carry 30.1\% of the variance of the 132 directions, 16.5 times their share (Fig.~\ref{fig:bio}f). The more markers a fetal type loses to HVG selection, the more the whole transcriptome helps it: across 63 types, the gain in the share of a type's cells recovered in one cluster, from 2,000 HVGs to every gene, grew with the share of markers removed (Spearman $\rho = 0.30$, $P = 0.015$; Fig.~\ref{fig:bio}i).

\textit{Drug targets.} The removed genes include 85 targets of 369 drugs approved by the US Food and Drug Administration (FDA; interactions from DGIdb, approval checked in Drugs@FDA; Fig.~\ref{fig:bio}g) and 229 genes with a definitive, strong or moderate disease link in ClinGen. Among them are the sodium channels \textit{SCN1A}, \textit{SCN2A} and \textit{SCN9A}, targets of antiepileptic drugs, and the N-methyl-D-aspartate (NMDA) receptor subunit \textit{GRIN2B}, the target of esketamine (Supplementary Note~6).

\FloatBarrier
\section*{Discussion}
KQE-2 shows that atlas-scale clustering can be exact, reproducible and accurate at the same time. Exactness and reproducibility come from one design choice, an entanglement-free quantum encoding, which also raised accuracy on standard genes of the developing cortex and RADC atlases. Other quantum approaches model gene interactions with entangling circuits or Hamiltonians \cite{romero2025,sohail2026} or need a future fault-tolerant computer \cite{zhao2026}. KQE-2 needs neither, and it analyzes a complete atlas of 11.4 million nuclei today.

\textit{Why the whole transcriptome matters.} Keeping every gene makes clustering both more complete and more auditable. Genes discarded upfront cannot influence the clusters, and a type merged during clustering loses its specific markers in later analyses. Standard 2,000-HVG filtering removes a median of 46.9\% of the top differentially expressed genes of each cell type across 29 datasets \cite{han2026gov}. Restricting the input to 2,000 HVGs lowered the data entropy of the signal directions by 1.5 to 3.1 bits across seven atlases, so the effective number of expression programs fell below the number of annotated types on five of them. Because each qubit depends only on its own score (Theorem~\ref{thm:var}), no encoding can restore a program that selection has removed. On all six complete atlases compared, KQE-2 was more accurate on the whole transcriptome than on 2,000 HVGs. Given the whole transcriptome, scVI instead lost accuracy (0.771 against 0.908 on the Siletti atlas).

\textit{Why a quantum encoding helps, biologically.} Three properties of cells explain the gain. First, cell identity is written in many genes at once. Two cell types usually differ by small amounts along many directions, whereas technical factors such as sequencing depth, mitochondrial read fraction or the cell cycle concentrate on one or two. The encoding turns every direction into a bounded term (Proposition~\ref{prop:sat}). One technical direction can therefore separate two cells by at most one qubit's worth of distance, while many small differences add up. This mechanism keeps the Siletti oligodendrocytes in one cluster. Second, gene programs saturate like switches, and the bounded rotations of the encoding follow this saturation, whereas unbounded scores push cells with extreme expression away from their type. Third, differentiation is continuous. Neighboring cells along a lineage differ by small scores, on which the encoding is nearly linear, so a lineage stays connected. Cells of different types differ on many qubits, and their fidelity falls toward zero.

\textit{Exactness instead of entanglement.} Quantum machine learning has pursued advantage through deep, entangled and trained circuits \cite{havlicek2019}, which face barren plateaus \cite{mcclean2018} and shot noise \cite{schuld2022goal,bowles2024}. KQE-2 takes the opposite course. It separates representation from optimization: a fixed, exact entanglement-free encoding is followed by deterministic clustering on the certified graph of the quantum states (quantum k-means, Supplementary Note~9). The representation remains quantum: every feature is the expectation of a Pauli observable of a well-defined multi-qubit state. Algebraically, each single-qubit feature is a trigonometric polynomial of its score; what a classical trigonometric map does not carry is the geometry of the states. Every Bloch vector has unit length, the fidelity of two cells is the product $\prod_q (1 - d_q^2/4)$ over the chord lengths $d_q$ between their Bloch vectors, and the certified neighbors and the graph of Q-PSC are built on this fidelity (Theorem~\ref{thm:state}). The same circuit runs on superconducting processors. Removing entanglement also preserved accuracy (Supplementary Note~5). KQE-2 thereby sets a reproducibility standard for quantum methods in biology:
\begin{enumerate}[leftmargin=*,itemsep=1pt,topsep=2pt]
\item an exact closed-form reference value for every quantum feature;
\item a proven variance bound that does not grow with the number of qubits;
\item an audit of every quantum hardware execution against its exact reference.
\end{enumerate}

\textit{Scope and boundary conditions.} KQE-2 brings its largest gains to imbalanced atlases whose abundant type other methods divide along a technical axis, such as the developing cortex and the Siletti atlas, and to developmental atlases whose types meet at continuous boundaries, such as the zebrafish and fetal atlases (Proposition~\ref{prop:sat}). On mature brain atlases with many neuronal subtypes, such as the mouse cortex and the RADC atlas, KQE-2 matches the best method within the seed variation of that method and adds exact reproducibility. All benchmarks here cluster at the annotated number of types $k$. With the input configuration of Table~\ref{tab:main}, which uses the annotated $k$ through the gene-set rule, and a resolution chosen by a label-free stability criterion fixed before the runs, KQE-2 chose finer partitions that recovered rare types better on seven of nine atlases, while the other methods chose coarser partitions with a higher AMI on most atlases (Supplementary Note~13). Its value lies in exactness rather than speed of quantum hardware: its features are computed exactly on a GPU, and a quantum processor reproduces them.

\textit{Conclusion.} KQE-2 applies an exactly computable quantum-state representation to complete single-cell atlases. It clusters 11.4 million nuclei reproducibly on one server and reaches competitive accuracy on large atlases, with no pretraining corpus and no stochastic batch correction. Accuracy and reproducibility are the twin pillars of trustworthy single-cell science, yet the pursuit of benchmark accuracy has long left reproducibility in the shadow. Fixing the seed of a stochastic model freezes one of many local optima without showing which partition reflects the biology. KQE-2 meets the requirement of a diagnostic assay, the same result for the same sample, by construction. Its central innovation, an exact entanglement-free quantum encoding, shows that quantum states can serve biology as exact, recomputable representations rather than as stochastic samples. The same design can carry to spatial omics, bioimaging and other fields whose data must be analyzed reproducibly.

\section*{Methods}
\paragraph{Datasets and the high-dimensional regime.} We evaluated KQE-2 on twelve public single-cell RNA-sequencing (scRNA-seq) datasets (Supplementary Table~13). We define a single-cell dataset as large-sample high-dimensional when the cell count $n$ exceeds the measured gene count $p$ ($n > p$, with $p$ of the order of $10^4$). All eleven atlases satisfy $n \gg p$, with sample-to-feature ratios $n/p$ from 15.3 (Siletti non-neuronal brain atlas; $p = 58{,}232$ genes) to 251 (complete mouse embryo atlas; $p = 45{,}525$ genes). The eleven atlases comprise 0.61 to 11.4 million cells or nuclei with raw count matrices and author annotations retrieved from CZ CELLxGENE Discover \cite{cellxgene2025}. Seven of them, used throughout, are:
\begin{itemize}[leftmargin=*,itemsep=1pt,topsep=2pt]
\item \textit{Zebrafish developmental atlas} \cite{duran2025}: whole embryos at 11 developmental stages profiled by sci-RNA-seq3 (100 annotated types).
\item \textit{Human fetal atlas} \cite{cao2020}: 15 fetal organs of 28 fetuses, also by sci-RNA-seq3.
\item \textit{Human optic nerve atlas} \cite{opticnerve2025}: optic nerve and optic disc of 83 donors.
\item \textit{Siletti non-neuronal brain atlas} \cite{siletti2023}: 88 regions of four adult human brains, 13 non-neuronal types dominated by mature oligodendrocytes (55\% of all cells) and including their precursors (OPCs).
\item \textit{Human heart cardiomyopathy atlas} \cite{reichart2022}: three cardiac regions of 79 hearts with dilated cardiomyopathy, arrhythmogenic cardiomyopathy or non-failing controls.
\item \textit{Developing human cortex atlas} \cite{velmeshev2023}: ten cortical regions of 106 donors from prenatal to postnatal stages, annotated into five broad types.
\item \textit{RADC prefrontal cortex atlas} \cite{lee2024radc}: dorsolateral prefrontal cortex nuclei of 152 aged donors, many with Alzheimer disease.
\end{itemize}
Four further atlases enter the complete-atlas benchmark (Table~\ref{tab:main}); three of them (marked $\dagger$) were run after the gene-set rule, defined below, had been fixed. For these three atlases, the threshold of the rule, the value of $\Delta$ and the chosen gene set were recorded with a time stamp before any run, together with the rule that applies the merge on whole-transcriptome atlases and Leiden at $k$ on 5,000-HVG atlases (recorded before the results of the mouse embryo). The quantum-state graph of Q-PSC was introduced afterwards and is applied in the same way to every atlas:
\begin{itemize}[leftmargin=*,itemsep=1pt,topsep=2pt]
\item \textit{Tabula Sapiens} \cite{tabula2022,tabula2025}: the All Cells release of Tabula Sapiens 2.0, 28 tissues of 24 donors, single cells profiled by 10x 3$'$ v3, 10x 5$'$ v2, Smart-seq2 and Smart-seq3 (171 annotated types).
\item \textit{Mouse cortex and hippocampus}$^{\dagger}$ \cite{yao2021cortex}: 17 regions of 55 adult mice, single cells (27 types).
\item \textit{Mouse embryo}$^{\dagger}$ \cite{qiu2024embryo}: nuclei of 74 embryos from gastrulation to birth profiled by sci-RNA-seq3, the first of the four 2.86-million-nucleus subsets distributed by the authors (134 types). The complete atlas of 11,441,407 nuclei, all four subsets together, is the largest dataset of this study; it is used for the scale test, and its gene-set rule was recorded before the run.
\item \textit{Mouse cerebellum}$^{\dagger}$ \cite{kozareva2021cereb}: nuclei of six adult mice (16 types), granule cells 78\% of the nuclei.
\end{itemize}
Apart from the three sci-RNA-seq3 atlases and the two single-cell atlases (Tabula Sapiens and mouse cortex), all were generated with 10x Genomics 3$'$ single-nucleus chemistry. The reproducibility test at ten million cells used the Parse Biosciences dataset of 9,697,974 peripheral blood mononuclear cells from 12 donors, each stimulated with one of 90 cytokines or left untreated, profiled by split-pool combinatorial barcoding (18 annotated types; release of 3 February 2026, CC BY-NC 4.0) \cite{parse10m}. KQE-2 ran in the configuration chosen by the gene-set rule (5,000 HVGs, $\Delta = 1.245$) and was run twice from raw counts; scVI ran on the 2,000 HVGs of scanpy with its default training, which scvi-tools sets to one epoch at this size, and seeds 0, 1 and 2. Before evaluation, unannotated cells and types with fewer than 10 cells were excluded. Table~\ref{tab:main} evaluates the complete atlases; the mechanism analyses (Extended Data Tables~2 and 4), Fig.~\ref{fig:umap} and the marker analyses of the fetal, Siletti and developing cortex atlases (Fig.~\ref{fig:bio}d--g,i) use a fixed random subsample of 150,000 cells per atlas (seed 0), so that every method is benchmarked on the same cells. Dataset identifiers and release versions are given in Supplementary Table~14.

\paragraph{Whole-transcriptome input and IDF weighting.} Let $X \in \mathbb{R}^{m \times p}_{\ge 0}$ be the raw count matrix of $n$ cells and $p$ genes, and let $d_g = |\{i : X_{ig} > 0\}|$ be the number of cells in which gene $g$ is detected. Under inverse document frequency (IDF) weighting, gene $g$ receives the normalized weight
\begin{equation}
w_g = \frac{\log\big((n+1)/(d_g+1)\big)}{\max_{g'} \log\big((n+1)/(d_{g'}+1)\big)},
\label{eq:idf}
\end{equation}
and cell $i$ is represented by the unit-normalized vector $z_i = \tilde{z}_i / \|\tilde{z}_i\|_2$ with $\tilde{z}_{ig} = w_g \log(1 + X_{ig})$, $g = 1, \dots, p$. This adapts term frequency--inverse document frequency weighting, which has been used to cluster scRNA-seq data \cite{moussa2018} and which, paired with a singular value decomposition (SVD), forms the latent semantic indexing of single-cell chromatin data \cite{cusanovich2015,stuart2021}, to the whole transcriptome. In ablations without IDF weighting, $w_g = 1$ and the counts are first scaled to $10^4$ per cell. Depending on the gene-set rule defined below, the input spans all $p$ genes or the top 5,000 highly variable genes (HVGs).

\paragraph{Theory: IDF weighting as the data entropy of a gene.} For gene $g$, let $v_{ig} = \log(1 + X_{ig})$, $u_{ig} = v_{ig} / \sum_{i'} v_{i'g}$ and $H_g = -\sum_{i} u_{ig} \log_2 u_{ig}$, the data entropy \cite{han2024is} of the expression profile of gene $g$ across the atlas, and let $d_g$ be the number of cells that detect gene $g$.

\begin{plainsays}
In plain words, the IDF weight measures how concentrated a gene is across cells with the same entropy that sets the width of our quantum circuit. A gene expressed uniformly across every cell that detects it receives exactly its IDF weight, and a marker concentrated in a rare population receives a stronger weight. \emph{Biologically:} a housekeeping gene detected in nearly every cell counts for little, whereas a marker detected only in the cells of one type, such as \textit{PROX1} in lymphatic endothelium, counts for much, so the markers of rare types shape the clusters even when they are expressed in few cells.
\end{plainsays}

The two quantities are linked by the maximum-entropy property of the uniform distribution. A vector of shares with $d_g$ nonzero entries has entropy at most $\log_2 d_g$, with equality if and only if the shares are equal. Hence $H_g \le \log_2 d_g$, with equality if and only if $X_{ig}$ takes the same nonzero value in every cell $i$ that detects gene $g$, and the information gain relative to a uniform distribution over all cells satisfies $\log_2 n - H_g \ge \log_2 (n / d_g)$, the inverse document frequency.

\noindent Empirically, because single-cell counts are sparse (in 50,000-cell subsamples of the Siletti, developing cortex and RADC atlases, 54\% to 62\% of all nonzero entries equal 1), the entropy gain $\log_2 n - H_g$ and the closed-form IDF $\log_2(n / d_g)$ have a Spearman correlation of 0.9999 over the genes detected in at least 10 cells, with a largest difference of 0.53 bits. KQE-2 therefore uses the analytical IDF weight of equation~(\ref{eq:idf}) in place of the gene entropy.

\paragraph{Deterministic spectral decomposition and retained signal directions.} Let $Z = [z_1, \dots, z_n]^{\top} \in \mathbb{R}^{n \times p}$ be the normalized cell-by-gene matrix. We center its columns implicitly and compute its leading singular vectors deterministically by the implicitly restarted Lanczos method of ARPACK \cite{lehoucq1998}, started from a fixed vector of all ones, with a convergence tolerance of $10^{-6}$ and no landmark subsampling. Because no random number is drawn, two executions on the same matrix, machine and software give bit-for-bit identical singular values and cell scores. (The perturbation experiments that test the variance bounds compare this exact solver with a randomized SVD with 20 oversamples and two power iterations, whose seed induces controlled score perturbations.) To separate signal directions from noise we perform parallel analysis \cite{horn1965} against a gene-permutation null: each gene's values are permuted independently across cells, and the $r$ leading directions whose singular value exceeds the mean $i$th singular value of the permuted matrix are retained. On 10,000-cell subsamples of the seven atlases, $r$ ranges from 37 to 72. To treat all atlases uniformly while matching the block architecture, KQE-2 retains $L = 132$ directions, a multiple of the number of qubits per block that exceeds $r$ on every dataset. The scores are standardized; when whitening is active, each direction is scaled to unit variance across cells, whereas unwhitened runs keep the singular value scaling.

\paragraph{Label-free choice of the number of qubits per block and of the gene set.} Let $\sigma_1 \ge \dots \ge \sigma_r$ be the singular values of the $r$ signal directions retained by parallel analysis and $u_i = \sigma_i / \sum_{j} \sigma_j$. The data entropy \cite{han2024is} of the retained signal subspace is
\begin{equation}
h_{\mathrm{sig}} = -\sum_{i=1}^{r} u_i \log_2 u_i,
\label{eq:hsig}
\end{equation}
the base-2 logarithm of the effective rank \cite{roy2007}. Computed without labels on 10,000 cells, it sets the number of qubits per block by Theorem~\ref{thm:qubits} as $l = \lceil h_{\mathrm{sig}} \rceil$. Across the atlases studied, $\lceil h_{\mathrm{sig}} \rceil$ falls between 5 and 7; KQE-2 fixes the median, $l = 6$, for all runs.

Whether KQE-2 operates on the whole transcriptome or on 5,000 HVGs is likewise decided before clustering, without labels, by the gene-set rule, which uses the spectral excess $\Delta = h_{\mathrm{sig}} - \log_2 k$ relative to the requested cluster count $k$, where $h_{\mathrm{sig}}$ is evaluated on all $p$ genes. Because $2^{h_{\mathrm{sig}}}$ is the effective number of expression programs in the data, $\Delta$ compares the active programs with the requested clusters. When $\Delta > 1.15$, that is, when the effective programs outnumber the clusters by more than a factor of $2^{1.15} \approx 2.2$, the additional programs describe variation within types, and KQE-2 restricts the input to 5,000 HVGs (counts scaled to $10^4$ per cell, $w_g = 1$). Otherwise it encodes the whole transcriptome with IDF weighting. The threshold was calibrated on 11 datasets, selecting the more accurate gene set on 10 of them, and fixed before any run on new atlases.

\paragraph{Entanglement-free quantum encoding and exact GPU evaluation.} For each cell, the $L = 132$ whitened scores are partitioned into $B = L/l = 22$ contiguous blocks of $l = 6$ scores. Each block $x = (x_1, \dots, x_l)^{\top} \in \mathbb{R}^{l}$ is mapped onto an $l$-qubit register initialized in $|0\rangle^{\otimes l}$ by $T$ data re-uploading layers. In layer $t \in \{1, \dots, T\}$, each qubit $q$ undergoes a dyadic $Y$-rotation followed by a $Z$-rotation, with no entangling gate (Theorem~\ref{thm:var}):
\begin{equation*}
V(x_q) = \prod_{t=1}^{T} R_Z(s\, x_q)\, R_Y(2^{-(t-1)} s\, x_q).
\end{equation*}
(In the entangled control, a chain of controlled-$Z$ gates on neighboring qubits is inserted between the $R_Y$ and $R_Z$ rotations of each layer.) From each encoded block, KQE-2 reads the single-qubit expectations $\langle X_q \rangle, \langle Y_q \rangle, \langle Z_q \rangle$ (the Bloch vector of qubit $q$) and all two-point correlations $\langle Z_i Z_j \rangle$, $1 \le i < j \le l$, within the block. Concatenating the $B$ blocks gives $B\,(3l + l(l-1)/2)$ exact quantum features per cell, $22 \times (18 + 15) = 726$ for $L = 132$ and $l = 6$. The rotation scale $s \in \{0.10, 0.15, 0.19, 0.25, 0.30, 0.40\}$ and depth $T \in \{2, 3, 4, 5\}$ are selected without labels by the stability of the Leiden partition (mean pairwise adjusted Rand index over three seeds); all full-atlas runs use $T = 3$ and $s = 0.19$.

\textit{Exact GPU evaluation.} States are evaluated by a batched statevector simulator written in PyTorch on the GPU, which applies single-qubit rotations as batched $2 \times 2$ contractions and controlled-$Z$ gates (for the entangled control) as phase-sign masks. Its expectations match brute-force Pauli traces to within $2 \times 10^{-7}$. In the entanglement-free encoding, each qubit evolves independently on the Bloch sphere, so the three Pauli expectations of qubit $q$ are exact closed-form trigonometric functions of the single scalar $x_q$.

\paragraph{Deterministic donor correction by quantum alignment.} When donor labels are provided for biological replicates, KQE-2 corrects donor effects deterministically. The unaligned scores are first partitioned by Leiden clustering at $k$ (seed 0). Any cluster in which one donor holds more than 80\% of the cells is merged, repeatedly, into the cluster with the nearest centroid, because such a cluster is formed by the donor effect. Within each remaining cluster $c$, the offset of donor $d$ is the shrunken difference of means
\begin{equation*}
\delta_{cd} = \frac{n_{cd}}{n_{cd} + \lambda}\,(\bar{x}_{cd} - \bar{x}_{c}),
\end{equation*}
where $\bar{x}_{cd}$ and $\bar{x}_{c}$ are the mean score vectors of donor $d$'s cells in cluster $c$ and of all cells in $c$, $n_{cd}$ is the number of donor $d$'s cells in $c$, and $\lambda = 20$ shrinks the offsets of small donor-cluster groups toward zero. Each cell's offset is subtracted, the scores are re-standardized, and the procedure is applied twice. Because every gate has the form $R_P(a x_q)$ about a fixed Pauli axis $P \in \{Y, Z\}$ and coaxial rotations add, $R_P(a x_q)\, R_P(-a\delta_{cd,q}) = R_P(a(x_q - \delta_{cd,q}))$: subtracting a donor offset in score space is equivalent to appending an inverse, donor-specific rotation about the same axis to each gate. No step draws a random number. For comparison, Harmony \cite{korsunsky2019} was applied to the same 132 scores with donor identity as the batch variable, followed by the same encoding.

\paragraph{Theory: data entropy sets the number of qubits.} For any matrix $Y$, the data entropy $h(Y)$ is equation~(\ref{eq:hsig}) applied to all nonzero singular values of $Y$; $h_{\mathrm{sig}}$ is the case of the signal directions. Let an encoding $\varphi$ map each cell $x_i$ to an $l$-qubit pure state $|\varphi(x_i)\rangle \in \mathbb{C}^{2^{l}}$. Stacking the $n$ encoded states as the rows of $\Psi$ defines the quantum data entropy $h_Q(X; \varphi) = h(\Psi)$, which depends only on the state Gram matrix $G_{ij} = \langle \varphi(x_i) | \varphi(x_j) \rangle$.

\begin{plainsays}
The theorem below says that the data entropy of the input tells how many qubits an encoding needs. A circuit with fewer than $h(X)$ qubits must compress orthogonal biological signals, whereas qubits beyond $h(X)$ scatter the states across an exponentially large Hilbert space until all pairs of cells look equally far apart, the regime in which quantum kernels concentrate. KQE-2 therefore sets the number of qubits per block from the data entropy before any circuit is run and without labels. \emph{Biologically:} the entropy counts how many independent expression programs the atlas holds, such as lineage, differentiation stage and cell cycle; KQE-2 gives the encoding as many qubits per block as there are programs to separate, and no more, so related cell types are neither forced together nor scattered apart.
\end{plainsays}

\begin{theorem}[Data entropy sets the number of qubits]
\label{thm:qubits}
Let $X$ have rank $r$ and data entropy $h(X)$, and let $\varphi$ encode each row of $X$ on $l$ qubits.
(a) \emph{Capacity bound.} $h_Q(X; \varphi) \le \log_2 \operatorname{rank}(\Psi) \le \min\{l, \log_2 n\}$.
(b) \emph{Inner-product preservation.} If the rows of $X$ have equal norm and $\langle \varphi(x_i) | \varphi(x_j) \rangle = c\, \langle x_i, x_j \rangle$ for all $i, j$ and some $c > 0$, then $h_Q(X; \varphi) = h(X)$ and $l \ge \log_2 r \ge h(X)$.
(c) \emph{Hilbert-space dilution.} With the ensemble density operator $\rho = \tfrac{1}{n} \Psi^{\dagger} \Psi$, the mean squared fidelity satisfies $\tfrac{1}{n^{2}} \sum_{i,j} |G_{ij}|^{2} = \operatorname{Tr} \rho^{2} \ge 2^{-l}$, with equality if and only if $\rho = I / 2^{l}$.
\end{theorem}

\noindent Part (b) gives the design rule $l = \lceil h \rceil$ for encodings that preserve inner products. The encoding of KQE-2 preserves distances up to a bounded factor (Theorem~\ref{thm:state}), and we adopt the same rule for it as a design choice motivated by part (b), not as a consequence of it. In the data, $\lceil h_{\mathrm{sig}} \rceil$ was 5 to 7 across the 11 atlases, with median 6 (Fig.~\ref{fig:quantum}e), the number of qubits per block used throughout. A proof is given in Supplementary Note~7.

\paragraph{Theory: dimension-independent variance bounds without entanglement.} Encoded features fluctuate for two reasons: the input scores change slightly between runs, and the processor that evaluates them is noisy. Entangling gates couple qubits across the register, so both kinds of error propagate across the light cone of the circuit.

\begin{plainsays}
In short, in an entanglement-free circuit each quantum feature is a closed-form trigonometric function of at most two scores. Perturbing a score shifts the feature by at most a fixed Lipschitz multiple of the perturbation, and noise on one qubit shifts its expectation by at most that qubit's error rate. Neither bound grows with the number of qubits, so the features are as stable on 132 qubits as on 6. With entangling gates, every feature reads every score and every noisy gate within its light cone, and its variance bound grows with the width and depth of the circuit. \emph{Biologically:} a small change in how one cell was measured, or a noisy qubit on a processor, moves that cell's encoding by a small, bounded amount and leaves every other cell untouched, so the cluster a cell belongs to does not hinge on technical noise elsewhere in the atlas.
\end{plainsays}

\begin{theorem}[Variance of the entanglement-free encoding]
\label{thm:var}
Let a block $x \in \mathbb{R}^{l}$ be encoded by the product unitary $U(x) = \bigotimes_{q=1}^{l} V(x_q)$, with $V(x_q) = \prod_{t=1}^{T} R_Z(s x_q)\, R_Y(\beta_t s x_q)$, $\beta_t = 2^{-(t-1)}$, and let $\Lambda = s \big(T + \sum_{t=1}^{T} \beta_t\big)$.
\begin{enumerate}
\item[(a)] \emph{Exact linear-time evaluation.} For $P \in \{X, Y, Z\}$, $\langle P_q \rangle = g_P(x_q)$ with $g_P(x) = \langle 0 | V(x)^{\dagger} P\, V(x) | 0 \rangle$, a trigonometric polynomial in $x$, and the two-point correlations factorize exactly as $\langle Z_i Z_j \rangle = g_Z(x_i)\, g_Z(x_j)$. All $3l + l(l-1)/2$ features of a block are evaluated exactly in $O(Tl + l^2)$ operations.
\item[(b)] \emph{Stability under data perturbation.} If the scores are perturbed by a random vector $\xi$, then $\mathrm{Var}\big[g_P(x_q + \xi_q)\big] \le \Lambda^2\, \mathrm{Var}[\xi_q]$ and $\mathrm{Var}\big[g_Z(x_i + \xi_i)\, g_Z(x_j + \xi_j)\big] \le 2\Lambda^2 \big(\mathrm{Var}[\xi_i] + \mathrm{Var}[\xi_j]\big)$, independently of the register width $l$.
\item[(c)] \emph{Stability under hardware noise.} If each qubit $q$ experiences a noise channel $\mathcal{N}_q$ within diamond-norm distance $\tfrac{1}{2}\|\mathcal{N}_q - \mathrm{id}\|_{\diamond} \le \varepsilon_q$ of the identity, the measured $\langle P_q \rangle$ deviates from $g_P(x_q)$ by at most $2\varepsilon_q$, and the measured $\langle Z_i Z_j \rangle$ from $g_Z(x_i) g_Z(x_j)$ by at most $2(\varepsilon_i + \varepsilon_j)$. Each error involves one or two qubits, whatever the width of the circuit.
\end{enumerate}
\end{theorem}

\noindent In the data, the bound of (b) held with room to spare: between two randomized decompositions of 150,000 cells, every Bloch vector moved by at most 0.71 times $\Lambda|\xi_q|$, and under Gaussian perturbations the largest feature variance was 0.57 to 0.75 of its bound. On an IBM processor, the measured features of 380 cells matched the exact values of (a) with a correlation of 0.978 (Supplementary Note~5). The proof is given in Supplementary Note~7.

\paragraph{Theory: necessity and sufficiency of removing entanglement.} We say that a single-qubit feature $f_{q,P}(x) = \langle P_q \rangle$ has \emph{local variance} if two conditions hold. \emph{Data locality} (D): $f_{q,P}$ depends on the data only through its own score $x_q$, so perturbing any other score $x_{q'}$, $q' \ne q$, leaves it unchanged. \emph{Noise locality} (N): inserting any single-qubit noise channel on any other qubit $q' \ne q$, at any layer of the circuit, leaves it unchanged. Here an entangling gate is any two-qubit gate that is not a product of single-qubit gates.

\begin{plainsays}
The next theorem shows that a feature confines data and hardware errors to its own qubit exactly when no entangling gate couples that qubit to the rest of the register. Once a single non-product two-qubit gate is present, crosstalk is unavoidable: there are inputs for which an error on a neighboring qubit changes the measured feature. \emph{Biologically:} each qubit carries one expression direction; without entangling gates, an error in one direction, for example a technical axis such as sequencing depth, cannot leak into the directions that carry lineage or cell-type identity.
\end{plainsays}

\begin{theorem}[Entanglement and local variance]
\label{thm:local}
(a) \emph{Sufficiency.} If the circuit contains no entangling gate, every single-qubit Pauli feature satisfies (D) and (N). (b) \emph{Necessity.} If a layer applies a two-qubit gate $\Gamma$ to qubits $q$ and $q'$ that is not a product $\Gamma_q \otimes \Gamma_{q'}$ of single-qubit gates, then there are product input states of $q$ and $q'$ and a Pauli operator $P$ for which $\langle P_q \rangle$ after $\Gamma$ changes with the state of qubit $q'$ and changes under a Pauli error on qubit $q'$. (c) \emph{Failure in entangled re-uploading circuits.} For the encoding of Theorem~\ref{thm:var} with controlled-$Z$ chains ($l = 6$, $T = 3$, $s = 0.19$ or $0.25$), (D) and (N) fail on sets of inputs of nonzero measure.
\end{theorem}

\noindent The experiments match both parts. Without entanglement, perturbing the other five scores of a block or inserting an $X$ error on a neighboring qubit changed a qubit's features by at most $1.3 \times 10^{-15}$, numerical precision. With controlled-$Z$ chains, the same error moved them by 0.23 to 0.27 on average (Supplementary Note~5). The proof is given in Supplementary Note~7.

\paragraph{Theory: reproducibility of graph clustering.} The Lipschitz stability of Theorems~\ref{thm:var} and \ref{thm:local} carries over from the features to the neighbor sets of the $K$-nearest-neighbor graph, and identical scores give an identical weighted graph and Leiden partition. For cell $i$, let $f_i \in \mathbb{R}^{3L}$ collect its $3L$ single-qubit features and let $\delta_i > 0$ be the gap between the distances to its $K$th and $(K+1)$th nearest neighbors in feature space, where $K = 15$ is the number of neighbors of the graph.

\begin{plainsays}
What follows for clustering: when the scores of a cell move slightly between two runs, its encoded state moves by at most a bounded multiple, and so does its distance to every other cell. As long as that movement stays below the gap that separates each cell's $K$th neighbor from the next candidate, every cell keeps the same nearest neighbors in both runs. When the scores are identical, as for KQE-2 on the same machine and software, the weighted graph is identical too, and so is every clustering or learning result computed from it. \emph{Biologically:} two analyses of the same atlas return the same clusters, and therefore the same markers, pathways and drug targets for every cell type, which is the property that a trained model loses when its seed changes.
\end{plainsays}

\begin{corollary}[Reproducibility in learning and clustering]
\label{cor:repro}
Under the entanglement-free encoding, suppose that two runs give scores $x_i$ and $x_i + \xi_i$ for every cell $i$. Then (a) $\|f_i(x_i + \xi_i) - f_i(x_i)\| \le \Lambda \|\xi_i\|$; (b) every pairwise feature distance changes by at most $\Lambda(\|\xi_i\| + \|\xi_j\|)$; and (c) if $\delta_i > 4 \Lambda \max_j \|\xi_j\|$ for every cell $i$, the two runs give every cell the same $K$ nearest neighbors; and (d) if $\xi_i = 0$ for every cell, the two runs construct the same weighted $K$-nearest-neighbor graph on all features, single- and two-qubit, and any clustering or learning method that accesses the data only through this graph, such as Leiden with a fixed seed or a $K$-nearest-neighbor classifier, returns the same output in both runs.
\end{corollary}

\noindent For the deterministic decomposition of KQE-2, $\xi_i = 0$ on the same machine and software, so part (d) gives identical graphs and clusters, as observed on every complete atlas. When the scores of two randomized decompositions differ, the encoding still gave nearly the same clusters (ARI 0.90 and 0.99 against 0.43 and 0.81 for the scores; Supplementary Note~5). The proof is given in Supplementary Note~7.

With entangling gates, Theorem~\ref{thm:local} shows that the Lipschitz bound of (a) must accumulate the scores of every qubit within the light cone of the entanglement, and a single hardware error moves the features of up to $2T + 1$ qubits, so the margin $\delta_i$ required for the same guarantee grows with the reach of the entanglement. Together, Theorems~\ref{thm:var} and \ref{thm:local} give a design principle: a quantum feature map that must remain auditable qubit by qubit and stable independently of its width should contain no entangling gate. The exact classical value of Theorem~\ref{thm:var}(a) is available for any number of qubits without entanglement; shallow entangled blocks of a few qubits can also be evaluated exactly, but their features no longer have local variance. Lipschitz bounds for quantum models have been given before \cite{berberich2024}; Theorems~\ref{thm:var} and \ref{thm:local} add the variance form, its independence from the number of qubits and the necessity of removing entanglement. Product states still show exponential concentration when evaluated through a global fidelity kernel \cite{thanasilp2024}; KQE-2 instead reads local Pauli observables and, where it uses the fidelity, its logarithm, which is a sum over qubits (Theorem~\ref{thm:state}).

\paragraph{Theory: feature geometry is quantum-state geometry.}
\begin{plainsays}
In one sentence, the quantum features that KQE-2 reads out are the complete quantum state of the cell. Two cells are close in feature space exactly when their states have high fidelity, so the nearest-neighbor graph that Leiden partitions is a graph of quantum states. Without entanglement, the exact fidelity costs linear time; with entanglement, it is as hard as computing an output amplitude of a quantum circuit. \emph{Biologically:} two cells fall in the same cluster because their whole quantum states, built from all their directions, are close, not because they agree on a few selected genes; the overlap of two cells is computed exactly, so the similarity that defines a cell type can be checked cell by cell.
\end{plainsays}

\begin{theorem}[Quantum-state geometry of the encoding]
\label{thm:state}
Let a cell with scores $x \in \mathbb{R}^{L}$ be encoded without entanglement as the product state $|\psi(x)\rangle = \bigotimes_{q=1}^{L} |\chi(x_q)\rangle$ with $|\chi(x_q)\rangle = V(x_q)|0\rangle$, let $\mathbf{r}_q(x) = (\langle X_q\rangle, \langle Y_q\rangle, \langle Z_q\rangle) \in \mathbb{R}^{3}$ be the Bloch vector of qubit $q$, let $f(x) = (r_1(x), \dots, r_L(x)) \in \mathbb{R}^{3L}$, and for two cells write $d_q = \|\mathbf{r}_q(x) - \mathbf{r}_q(y)\|$.
\begin{enumerate}
\item[(a)] \emph{The features are the state.} $\|\mathbf{r}_q(x)\| = 1$ and $|\psi(x)\rangle\langle\psi(x)| = \bigotimes_q \tfrac{1}{2}(I + \mathbf{r}_q(x)\cdot\boldsymbol{\tau})$ with $\boldsymbol{\tau} = (X, Y, Z)$ the vector of Pauli matrices, so $f(x)$ determines the $L$-qubit state and the state determines $f(x)$.
\item[(b)] \emph{Exact fidelity in linear time.} $F(x, y) = |\langle \psi(x) | \psi(y) \rangle|^2 = \prod_{q} \big(1 - d_q^2/4\big)$, computed with $O(L)$ operations.
\item[(c)] \emph{Feature distance is state distance.} $\tfrac{1}{2} d_q^2 = \|\rho_q(x) - \rho_q(y)\|_{\mathrm{HS}}^2$, and, if $d_q < 2$ for every qubit,
\begin{equation}
\tfrac{1}{4}\|f(x) - f(y)\|^2 \;\le\; -\log F(x, y) \;\le\; \sum_{q} \frac{d_q^2/4}{1 - d_q^2/4}.
\label{eq:state}
\end{equation}
If $d_q = 2$ for some qubit, then $F(x, y) = 0$ and $-\log F(x, y) = +\infty$.
\item[(d)] \emph{Complexity.} For a circuit $U(x)$ with entangling gates, $F(x, y) = |\langle 0 | U(x)^{\dagger} U(y) | 0\rangle|^2$ is a transition amplitude of $U(x)^{\dagger} U(y)$, whose exact evaluation is \#P-hard in general \cite{valiant1979,aaronson2013}. Removing entanglement reduces this cost to $O(L)$.
\end{enumerate}
\end{theorem}

\noindent The data confirm (a) to (c). Every one of the $146{,}598 \times 132$ Bloch vectors of the developing cortex has unit norm to six decimals. Across 80,000 pairs of Siletti cells, every pair satisfied the lower bound of (c), and the feature distance and $-\log F$ had a Spearman correlation of 0.9998 (Extended Data Fig.~5). The proof is given in Supplementary Note~7.

\paragraph{Theory: saturation of single-direction differences.} Let two cells have scores $x, y \in \mathbb{R}^{L}$ and write $\eta_q = x_q - y_q$. In score space their distance is $\|x - y\|^2 = \sum_q \eta_q^2$, which one direction can dominate. The encoding replaces each term by the chord between two points of the Bloch sphere.

\begin{plainsays}
Intuitively, the encoding turns every direction into a bounded, saturating term. A large difference along a single direction, such as a technical axis of sequencing depth or mitochondrial fraction, can move two cells apart by at most one qubit's worth of distance, however large it is. Many small differences spread over many directions, the signature of a different cell type, add up almost linearly and are kept. Within-type spread is compressed and between-type separation is preserved, which is why abundant types stay whole and continuous lineages stay connected. \emph{Biologically:} the oligodendrocytes of the Siletti atlas, which other methods split by mitochondrial read fraction, stay in one cluster, and cells along a differentiation path, such as OPCs maturing into oligodendrocytes, stay connected rather than being cut at an arbitrary point.
\end{plainsays}

\begin{proposition}[Saturation of single-direction differences]
\label{prop:sat}
For the encoding of Theorem~\ref{thm:var}, with $d_q = \|\mathbf{r}_q(x) - \mathbf{r}_q(y)\|$ and $\Lambda = s\,(T + \sum_{t} \beta_t)$:
\begin{enumerate}
\item[(a)] \emph{Bounded contribution of every direction.} $d_q \le 2$ for every qubit $q$, so $\|f(x) - f(y)\|^2 = \sum_q d_q^2 \le 4L$, whereas $\sum_q \eta_q^2$ is unbounded.
\item[(b)] \emph{Linear for small and saturating for large differences.} $d_q \le \min\{\Lambda |\eta_q|,\, 2\}$. For a single layer of $R_Y$ rotations the chord is exact, $d_q = 2\,|\sin(s\eta_q/2)|$, so that $d_q \ge (2s/\pi)\,|\eta_q|$ whenever $s|\eta_q| \le \pi$: the map is linear up to the constant $2s/\pi$ on differences below $\pi/s$ and saturates beyond.
\item[(c)] \emph{Many small differences outweigh one large difference.} If $x$ and $y$ differ only along one direction, by any amount $\eta_0$, then $\|f(x) - f(y)\| \le 2$. If they differ by $\eta \le \pi/s$ along each of $M$ directions, then for the single-layer map $\|f(x) - f(y)\| \ge (2s/\pi)\,\eta\sqrt{M}$. The ratio of the second distance to the first is therefore at least $s\eta\sqrt{M}/\pi$, which grows with $M$, whereas in score space the same ratio is $\eta\sqrt{M}/|\eta_0|$, which vanishes as $|\eta_0|$ grows.
\item[(d)] \emph{Fidelity form.} For the single-layer map, $F(x, y) = \prod_q \cos^2(s\eta_q/2)$ and $-\log F(x, y) = \sum_q -\log\cos^2(s\eta_q/2) \ge \sum_q (s\eta_q)^2/4$: the state distance is a sum over directions of terms that are quadratic for small differences, so neighbors along a continuous lineage keep a fidelity near one and cells that differ on many qubits have a fidelity near zero.
\end{enumerate}
\end{proposition}

\noindent \textit{Proof.} (a) Bloch vectors have unit norm (Theorem~\ref{thm:state}a), so the chord between two of them is at most 2. (b) The upper bound is the Lipschitz bound of Theorem~\ref{thm:var}(b) applied to the Bloch map. For one $R_Y$ layer, $\mathbf{r}_q(x) = (\sin s x_q, 0, \cos s x_q)$, so $d_q = 2|\sin(s\eta_q/2)|$, and $\sin(u/2) \ge u/\pi$ for $0 \le u \le \pi$. (c) The first statement is (a) for one qubit; the second sums the lower bound of (b) over the $M$ differing qubits. (d) For one $R_Y$ layer, $|\langle \chi(x_q) | \chi(y_q) \rangle|^2 = \cos^2(s\eta_q/2)$, the product follows from Theorem~\ref{thm:state}(b), and $-\log\cos^2(u/2) \ge u^2/4$ for $|u| \le \pi$. \hfill$\square$

\noindent The upper bounds hold as stated for the three-layer circuit of KQE-2. The exact chord and the lower bounds are written for one $R_Y$ layer, and for the three-layer circuit they hold with point-dependent constants. The measured separations agree: on the Siletti atlas, the encoding raised the separation of every major glial type, for example oligodendrocytes from 0.24 to 0.40, and KQE-2 kept 99.6\% of the oligodendrocytes in one cluster (Results).

\paragraph{Embedding geometry metrics.} Given an embedding $E$ with rows $E_i$ and an annotated type $c$ with $n_c$ cells, centroid $\mu_c$ and within-type scatter $t_c = n_c^{-1}\sum_{i \in c} \|E_i - \mu_c\|^2$, the separation of $c$ is $\min_{c' \neq c} \|\mu_c - \mu_{c'}\| / \sqrt{t_c}$, and its neighborhood purity is the mean fraction of the 15 nearest neighbors of its cells that share its annotation. Both are computed identically on the unencoded whitened scores and on the quantum features of the same 150,000 cells; labels are used only for these measurements.

\paragraph{Exact quantum-state neighbors of a complete atlas.} For each nucleus, the Bloch vectors $\mathbf{r}_q$ of its 132 qubits are computed in closed form from its whitened scores ($T = 3$, $s = 0.19$). Candidate neighbors are the $M = 200$ nuclei nearest in Euclidean distance between the Bloch-vector features $f$, found on the GPU by matrix products. Their exact state distance $-\log F = -\sum_q \log\big((1 + \mathbf{r}_q \cdot \mathbf{r}'_q)/2\big)$ is computed, and the 15 smallest are kept. By Theorem~\ref{thm:state}(c), every nucleus outside the candidate list has $-\log F \ge \tfrac{1}{4}\|f - f'\|^2 \ge \tfrac{1}{4} e_M$, where $e_M$ is the $M$th smallest squared feature distance. When $\tfrac{1}{4} e_M$ is at least the 15th smallest exact distance among the candidates, the 15 neighbors are therefore exact; for the nuclei without this certificate, all distances are computed exhaustively. The map of Supplementary Fig.~2b is a UMAP of the resulting 15-nearest-neighbor graph (random seed 0), colored by the KQE-2 partition of Table~\ref{tab:main}. Fidelities of random pairs use 200,000 pairs drawn from a stratified sample of 30,909 nuclei (at most 2,500 per cluster). The same procedure was applied to all 11,441,407 nuclei of the complete mouse embryo atlas; the colors of Fig.~\ref{fig:overview}c and Supplementary Fig.~2e merge the 26 major cell clusters of the authors into eight lineages.

\paragraph{Computational complexity.} With $n$ cells, $p$ genes, nnz nonzero counts, $L = 132$ directions and $K = 15$ neighbors, every step of KQE-2 except the exact neighbor search is linear in $n$ (Supplementary Table~15). IDF weighting and normalization cost $O(\mathrm{nnz})$ and keep the matrix sparse. The decomposition uses only sparse matrix--vector products, $O(\mathrm{nnz}\, L + (n + p) L^2)$ per Lanczos restart, so no dense $n \times p$ matrix is formed. Without entanglement, each qubit is evaluated in closed form, so the encoding costs $O(nL)$ instead of the $2^{l}$-dimensional state vector of an entangled block (Theorem~\ref{thm:var}), and the fidelity of two cells costs $O(L)$ (Theorem~\ref{thm:state}). The exact neighbor search is quadratic, $O(n^2 \cdot 3L)$ multiply-adds for the candidates on the GPU and $O(nML)$ for the exact fidelities of the $M = 200$ candidates; the lower bound of Theorem~\ref{thm:state} certifies the neighbors without an exhaustive search for 99.99\% of the Siletti nuclei. Leiden and the merge scale with the $nK$ edges and with $k'^3$, $k' = \mathrm{round}(1.5k)$. Memory is $O(\mathrm{nnz} + n \cdot 726)$, 33 GB of features for 11.4 million nuclei. On that atlas, the decomposition took 11.9 h, the certified neighbor search 3.1 h and Q-PSC 12 min. The compared methods are limited by training (scVI, scGPT and the dedicated methods), by dense intermediate matrices (scanpy and Seurat scale an $n \times 2{,}000$ matrix; scAGCL builds an $n \times n$ matrix) or by a hard limit of $2^{31}$ nonzero entries (Seurat).

\paragraph{Rare embryonic types.} On the mouse embryo atlas, the F1 of each annotated type is that of the cluster holding most of its nuclei (Leiden seed 0, annotated $k$; KQE-2 with Q-PSC). For the marker check on the complete atlas, 3,000 nuclei of that KQE-2 cluster and 3,000 nuclei outside it were drawn at random (seed 0); counts were normalized to $10^4$ per nucleus and log-transformed, the share of nuclei with a nonzero count of each canonical marker was compared inside and outside, and the top 10 markers of the cluster were ranked by the Wilcoxon test of scanpy.

\paragraph{Classical feature maps.} To isolate the contribution of the quantum map, we benchmarked it against five families of transformations applied to the same whitened scores and clustered with the same protocol: block-wise random Fourier features $\cos(s\, x W + b)$ with the output width of the quantum block; deterministic cosines and sines at the frequencies of the $T$-layer circuit; a quadratic map per block; the bounded maps $\tanh(x/a)$ with $a \in \{1, 2, 3\}$ and $\max(-c, \min(c, x))$ with $c \in \{1.5, 2.5\}$; and the circuit with controlled-$Z$ chains. For the rare-population titration, the OPCs of the complete Siletti atlas and the microglia of the complete RADC atlas were subsampled to 2\% and 0.5\% of the cells (seed 1) with every other cell kept; each representation was clustered at the annotated $k$ with three Leiden seeds, and recovery was measured by the F1 of the type against its best-matching cluster.

\paragraph{Reproducibility and perturbation experiments.} Run-to-run perturbations are the differences between the scores of two randomized decompositions with seeds 0 and 1, after aligning the sign of each direction; Gaussian perturbations of standard deviation $\nu \in \{0.05, 0.1, 0.2\}$ are added to the scores of 2,000 cells, 30 draws each. Local variance (Theorem~\ref{thm:local}) is tested on 1,000 cells with the statevector of one six-qubit block, perturbing the other five scores ($\nu = 0.2$, 20 draws) or inserting a Pauli-$X$ error on qubit 2 before the last layer. State-space graphs use $-\log F$ or the Euclidean distance between Bloch vectors as the distance of the 15-nearest-neighbor graph. For the training-seed test, scVI was trained with seeds 0, 1 and 2 on the complete Siletti, zebrafish, fetal, developing cortex and RADC atlases with the benchmark settings; a second run with seed 0 gave an identical embedding. scGPT embeddings of the complete Siletti atlas were computed three times with the benchmark settings (whole-human checkpoint, which samples at most 1,200 genes per cell). To measure how a training seed changes the biology on the fetal atlas, each annotated type was matched to the scVI cluster holding most of its cells under each seed; markers were the top-50 Wilcoxon markers of that cluster against all other cells (log-normalized counts of the whole transcriptome, up to 3,000 cells per group, log fold change above 1 and Benjamini--Hochberg adjusted $P < 0.01$), pathways were taken from Enrichr (GO Biological Process 2023 and KEGG 2021) and established drug targets were defined as in the drug-target analysis below. Each embedding was clustered at the annotated $k$ with Leiden seed 0, and the share of cells that changed cluster between two runs is one minus the size of the largest one-to-one matching between the two partitions, as a fraction of all cells.

\paragraph{Pathway, interactome and drug-target analyses.} On each 150,000-cell subsample, markers were called on the log-normalized counts of the whole transcriptome for every type with at least 20 cells (up to 3,000 cells per type) with the Wilcoxon test of scanpy against all other cells, keeping the top 50 per type with log fold change above 1 and Benjamini--Hochberg adjusted $P < 0.01$. Pathway enrichment used Enrichr \cite{kuleshov2016} (Gene Ontology (GO) Biological Process 2023, KEGG 2021 Human and Reactome 2022), protein interaction density used STRING v12 \cite{szklarczyk2023} (\textit{Homo sapiens} interaction enrichment test), and drug targets were queried through the GraphQL interface of DGIdb 5.0 \cite{cannon2024}, all on 15 October 2026. An established drug target is a gene with at least one interaction with an approved drug that states a mode of action (for example inhibitor, antagonist, agonist or blocker) from ChEMBL, Guide to Pharmacology, TTD, TEND, DrugBank or the FDA. Disease genes have a definitive, strong or moderate gene--disease validity classification in ClinGen \cite{strande2017}, downloaded on 15 October 2026. Approval by the FDA was checked for every drug in Drugs@FDA (openFDA interface, 8 November 2026), matching drug names with their salt forms. For the 52,888 cardiomyocytes of the 141,017-cell benchmark sample of the heart atlas, the top-50 Wilcoxon markers were called on the whole transcriptome against the 2,000 HVGs of scanpy (flavor `seurat'); the network of the 50 genes and the interaction enrichment of the removed genes used STRING v12.5 (combined score at least 0.4), and their pathways used Enrichr, both on 8 November 2026. The share of retained variance carried by gene $g$ is $\sum_{i} \sigma_i^2 V_{gi}^2 / \sum_{i} \sigma_i^2$, where $\sigma_i$ are the 132 retained singular values and $V$ the gene loadings of the IDF-weighted matrix.

\paragraph{Baseline embedding and clustering methods.} All competing methods were run with their standard published pipelines. scanpy \cite{wolf2018}: normalization to $10^4$ counts, log1p, 2,000 HVGs, unit-variance scaling clipped at 10, 50 principal components. Seurat 5.3.0 \cite{hao2024}: NormalizeData, FindVariableFeatures (2,000 genes), ScaleData, 50 principal components. scVI 1.4.2 \cite{lopez2018}: raw counts of 2,000 HVGs, 30 latent dimensions, default training. scGPT \cite{cui2024}: the pretrained whole-human checkpoint applied to raw counts. Dedicated clustering methods were run from their official repositories on the same 150,000-cell subsamples (seed 0) under a 3-hour wall-clock limit and a 150-GB memory watchdog: scICE \cite{kim2025scice} (scLENS preprocessing on the GPU, UMAP, consistency clustering over 1 to $\max(20, k+5)$ clusters), scDCCA \cite{wang2023scdcca} (2,000 HVGs, size-factor normalization, latent dimension 32, encoder layers 256 and 64, 70 pretraining and 100 training epochs), scAGCL \cite{vinh2025scagcl} (its own graph preprocessing, 500 epochs, $k$-means) and scDAC \cite{an2024scdac} (raw counts of 4,000 Seurat-v3 HVGs, 500 epochs).

\paragraph{Maps.} The maps of Figs.~1b, 2b, 3 and 4a--d were drawn with SR-tSNE2, a fast implementation of the rank-preserving method SR-tSNE \cite{han2026srtsne}. It computes a t-SNE embedding \cite{vandermaaten2008} with openTSNE \cite{policar2024} (perplexity 30, PCA initialization, seed 42, 250 early-exaggeration and 750 further iterations) and then refines it, pulling each point toward its 90 nearest neighbors in the input space (weight $\lambda = 50$, 10 neighbors per step). Each map is built from the embedding of the method it shows; Fig.~4a--d colors one map of the KQE-2 features by the clusters of each run. The maps of Fig.~1c,d are the UMAP of the certified 15-nearest-state graph of the complete mouse embryo atlas (see below). In Fig.~3, each embedding was clustered by Leiden at the annotated $k$ (seed 0) on its 15-nearest-neighbor graph, the protocol of Table~\ref{tab:main}. A cell is misassigned when its cluster is not the one paired with its annotated type in the one-to-one pairing of clusters and types that matches the most cells (Hungarian algorithm on the contingency table).

\paragraph{Unified graph construction and Leiden scoring.} Every embedding was evaluated through the same downstream protocol. An exact 15-nearest-neighbor graph was built on the GPU, weighted with the fuzzy simplicial set of UMAP \cite{mcinnes2018} and partitioned with Leiden \cite{traag2019}. To score at the annotated number of types $k$, the resolution was found by bisection from 0.02 (at most 12 steps) until the partition had $k$ clusters, for Leiden seeds 0, 1 and 2. Table~\ref{tab:main}, Fig.~\ref{fig:umap} and Extended Data Tables~3 and 4 report seed 0; the other tables report the mean over the three seeds. Labels are used only to set $k$ and to score.

\paragraph{Q-PSC: quantum precision shrinkage clustering on the quantum-state graph.} On whole-transcriptome atlases, KQE-2 clusters on the graph of its quantum states (Supplementary Fig.~1). The 15 neighbors of each cell are its certified exact nearest states under $-\log F$ (previous paragraphs; Theorem~\ref{thm:state}), and the edge between cells $i$ and $j$ is weighted by their fidelity $F_{ij} = \prod_q \tfrac{1}{2}(1 + \mathbf{r}_{iq} \cdot \mathbf{r}_{jq})$; the weight matrix is symmetrized by the larger of the two directions. Leiden (seed 0) is bisected to $k' = \mathrm{round}(1.5k)$ clusters (Python rounding, which sends halves to the even integer). With $\Pi$ the $n \times k'$ membership matrix and $A$ the fidelity-weighted adjacency matrix, $C = \Pi^{\top} A \Pi$ holds the summed fidelity between clusters; the pair $(a, b)$ that maximizes $C_{ab} / \sqrt{v_a v_b}$, with $v_a = \sum_{b'} C_{ab'}$, is merged, $C$ and $v$ are updated, and the step is repeated until $k$ clusters remain. Every step is deterministic and needs no parameter beyond the factor 1.5. Given the scores, the certified neighbors and the clusters were bit-for-bit identical on rerun and on a second server with another software environment (mouse cerebellum). On the complete mouse embryo atlas, the bisection toward $k' = 201$ reached 202 clusters, and that partition was merged to $k = 134$.

\paragraph{Quantum hardware.} Circuits were executed through the IBM Quantum Open plan on the 156-qubit superconducting processor ibm\_marrakesh (Heron r2) with Qiskit 2.5.2 and qiskit-ibm-runtime 0.50.0 (EstimatorV2, 1,024 shots, readout-error mitigation at resilience level 1). Qubits with readout error above 0.15 were excluded, and disjoint 6-qubit paths were found on the coupling graph. Jobs db82lm6il99s73fp68ag (50 cells) and db82u8mr6oqs7388l1mg (380 cells), 15 October 2026 (Supplementary Note~5).

\paragraph{Statistics.} AMI \cite{vinh2010} and the adjusted Rand index \cite{hubert1985} were computed with scikit-learn; values follow the seed convention of the scoring protocol above, with the range of the seeds given where it is reported. Agreement between hardware and exact features is measured by the Pearson correlation and distance preservation by the Spearman correlation. Because every method was evaluated on the same cells, graph construction and Leiden protocol, comparisons between methods are paired. The AMI of KQE-2 in the configuration of Table~\ref{tab:main} was compared with that of the best other method and of each method on the atlases where it ran, by the one-sided Wilcoxon signed-rank test over atlases (SciPy), with a bootstrap 95\% confidence interval of the mean margin (10,000 resamples of the atlases, seed 0). Rare types are annotated types below 1\% of the cells, and the F1 of a type is that of the cluster holding most of its cells (Leiden seed 0, annotated $k$). The F1 of KQE-2 and of each other method were compared over the rare types by the one-sided Wilcoxon signed-rank test stratified by atlas (van Elteren), which ranks the differences within each atlas and weights each atlas by $1/(n_a + 1)$ for its $n_a$ rare types, so that no atlas dominates the test.

\section*{Data availability}
All atlases are public on CZ CELLxGENE Discover \cite{cellxgene2025}; the dataset versions are listed in Supplementary Table~14. The cytokine-stimulated PBMC dataset is distributed by Parse Biosciences under a CC BY-NC 4.0 license \cite{parse10m}.

\section*{Code availability}
The Python package scKQE2 (imported as sckqe2; version 2.1.0) implements every step of KQE-2, including the gene-set rule, Q-PSC on the certified quantum-state graph, quantum alignment, the certified exact search of nearest quantum states and the circuits run on quantum hardware. It comes with a tutorial, a Google Colab notebook, tests against a brute-force statevector simulation and scripts that download the atlases and reproduce the results, and is available at \url{https://github.com/hank08819/scKQE-2}.

{\fontsize{7.5}{8.8}\selectfont\setlength{\itemsep}{0pt}
}

\section*{Acknowledgements}
The concept of kernel quantum encoding (KQE) and its initial quantum machine learning results originated in September 2021. KQE-2 represents the culmination of more than five years of sustained exploration, evolving through successive iterations of the quantum encoding, whole-transcriptome modeling and mathematical theory across ever-larger single-cell atlases, and ultimately arriving at the exact, entanglement-free framework presented here. We thank Dr. Erica Baker for insightful discussions on quantum computing in fall 2021. Above all, we pay tribute to the memory of Professor Zhang Han (1938--2024), whose unwavering encouragement and lifelong guidance made this journey possible.

\section*{Funding}
This work was supported by the start-up funds of the McCollum Family Chair in Data Sciences, Baylor University.

\section*{Author contributions}
H.H. conceived the study, developed the method, proved the theorems, performed the experiments and wrote the manuscript.

\section*{Competing interests}
The author declares no competing interests.

\onecolumn
\makeatletter
\renewenvironment{figure}[1][]{\par\medskip\begin{minipage}{\textwidth}\centering\def\@captype{figure}}{\end{minipage}\par\bigskip}
\renewenvironment{table}[1][]{\par\medskip\begin{minipage}{\textwidth}\centering\sffamily\footnotesize\def\@captype{table}}{\end{minipage}\par\bigskip}
\makeatother
\section*{Extended Data}
\begin{figure}[h]
\centering
\includegraphics[width=0.85\textwidth]{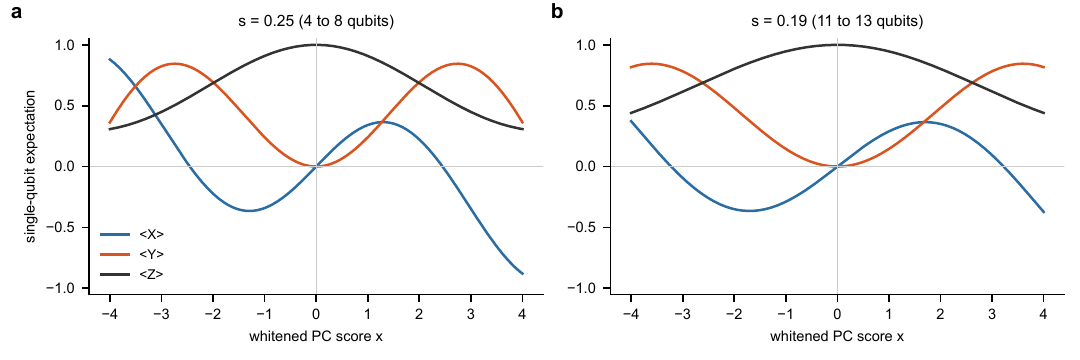}
\caption*{\textbf{Extended Data Fig. 1 $|$ The single-qubit map is a bounded function of the whitened score.} $\langle X\rangle$, $\langle Y\rangle$ and $\langle Z\rangle$ of one qubit after three layers $R_Y(\beta_t s x)$, $R_Z(s x)$, $\beta_t = 1, 1/2, 1/4$, against the whitened score $x$; \textbf{a}, $s = 0.25$; \textbf{b}, $s = 0.19$. On the developing cortex (mean of three Leiden seeds), all quantum features reach an AMI of 0.740, the single-qubit features 0.651, $\langle Z\rangle$ alone 0.677 and the whitened scores 0.192.}
\label{fig:gfun}
\end{figure}

\begin{figure}[h]
\centering
\includegraphics[width=\textwidth]{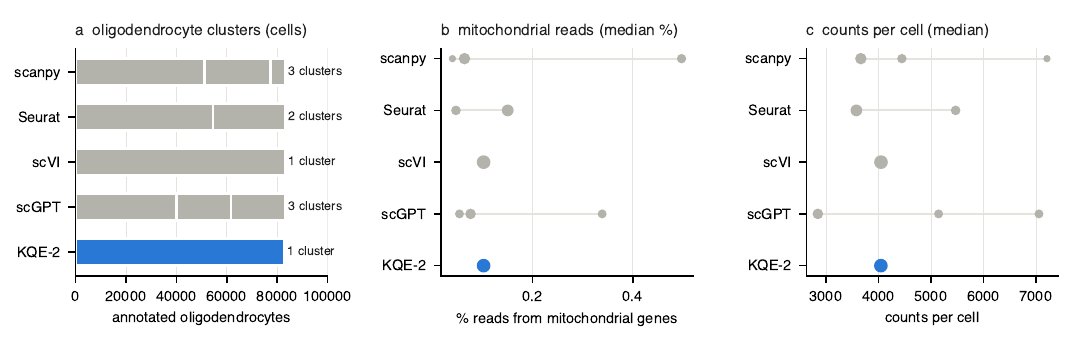}
\caption*{\textbf{Extended Data Fig. 2 $|$ The oligodendrocyte clusters of other methods follow technical quality.} Siletti non-neuronal brain atlas, the 150,000 cells of Fig.~\ref{fig:umap}a with the complete-atlas partitions; for each method, the clusters that hold at least 5\% of the annotated oligodendrocytes. \textbf{a}, Number of oligodendrocytes in each cluster: KQE-2 keeps 99.6\% of them in one cluster, and scVI keeps them whole in this run, whereas scanpy, Seurat and scGPT split them. \textbf{b}, Median percentage of counts from the 37 mitochondrial genes: the second scanpy cluster has 7.6 times the mitochondrial fraction of the largest one, and the clusters of Seurat and scGPT differ 3.1- and 4.4-fold. \textbf{c}, Median counts per cell, which differ 1.5- and 2.5-fold between the clusters of Seurat and of scGPT. Dot size is proportional to the number of cells in the cluster.}
\label{fig:oligo}
\end{figure}

\begin{figure}[h]
\centering
\includegraphics[width=\textwidth]{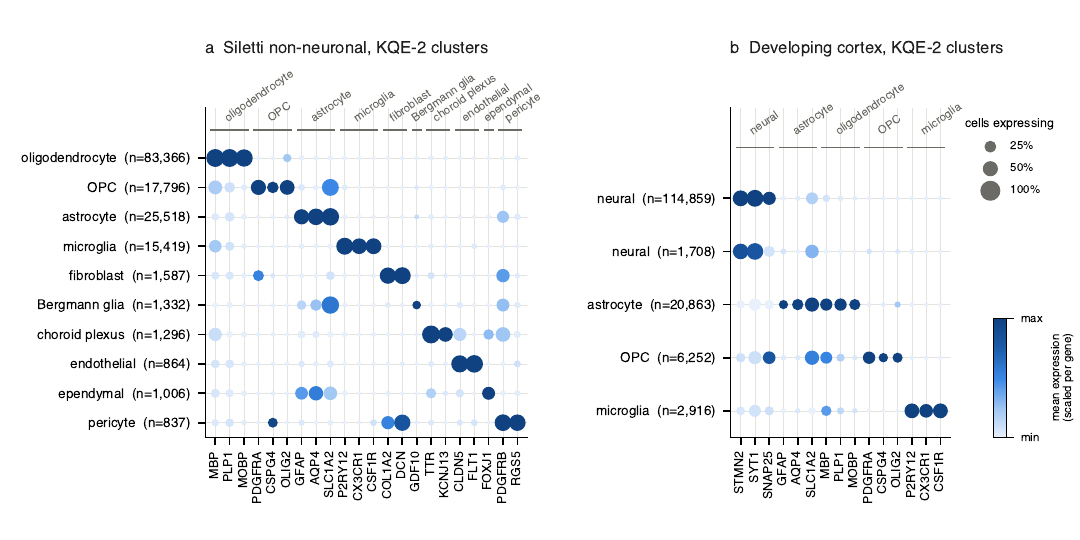}
\caption*{\textbf{Extended Data Fig. 3 $|$ Every KQE-2 cluster carries the canonical markers of its type.} Dot plots of the KQE-2 clusters with at least 0.5\% of the cells, on the cells of Fig.~\ref{fig:umap}; rows, majority annotated type and cell count; dot size, fraction of expressing cells; color, scaled mean expression. \textbf{a}, Siletti non-neuronal brain atlas, complete-atlas KQE-2 partition (Q-PSC), 10 clusters: OPCs express \textit{PDGFRA}, \textit{CSPG4} and \textit{OLIG2}, microglia \textit{P2RY12}, \textit{CX3CR1} and \textit{CSF1R}, and oligodendrocytes \textit{MBP}, \textit{PLP1} and \textit{MOBP}. \textbf{b}, Developing cortex, 5 clusters.}
\label{fig:markers}
\end{figure}

\begin{figure}[h]
\centering
\includegraphics[width=\textwidth]{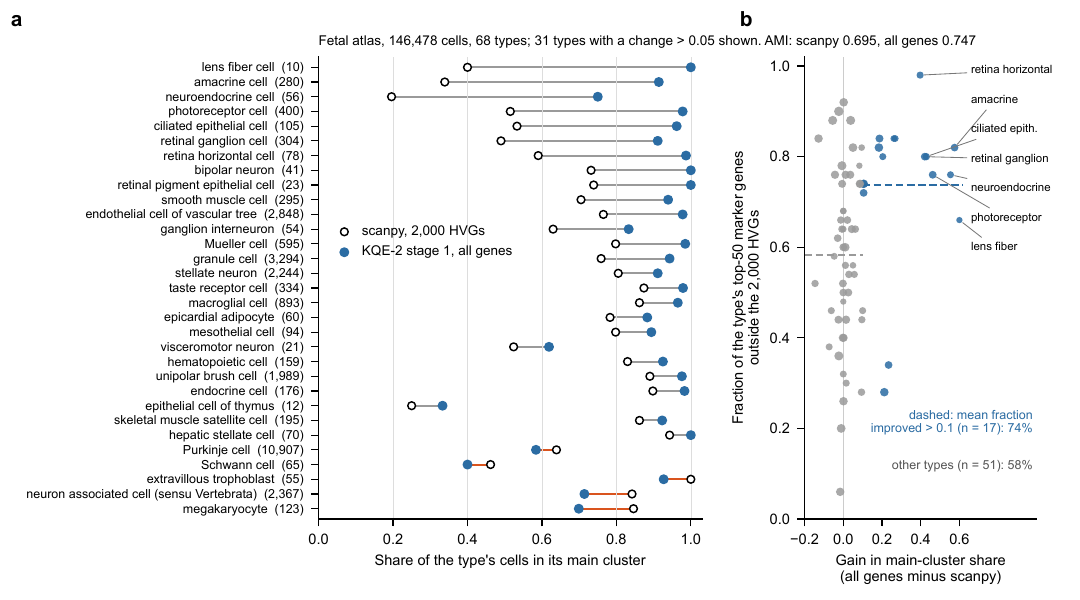}
\caption*{\textbf{Extended Data Fig. 4 $|$ Clustering on the whole transcriptome recovers the rare types whose markers HVG selection removes.} Human fetal atlas, 146,478 cells, 68 annotated types (used only for scoring). \textbf{a}, Share of each type's cells in its main Leiden cluster under scanpy (2,000 HVGs, 50 principal components; open circles) and under the first stage of KQE-2, the deterministic decomposition of all 45,676 genes before the quantum encoding (130 directions, library-size normalized log counts; filled circles), for the 31 types whose share changed by more than 0.05. Leiden at the annotated $k$, seed 0. AMI: scanpy 0.695, whole transcriptome 0.747. \textbf{b}, Gain in main-cluster share against the share of each type's top-50 markers outside the 2,000 HVGs. The 17 types that gained more than 0.1 (blue) have 74\% of their markers outside the HVGs, against 58\% for the other 51 types (dashed lines).}
\end{figure}

\begin{figure}[h]
\centering
\includegraphics[width=\textwidth]{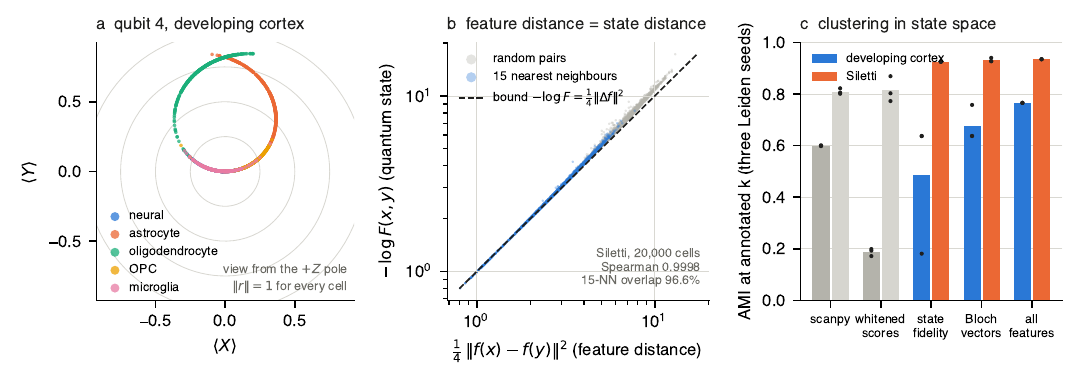}
\caption*{\textbf{Extended Data Fig. 5 $|$ The clusters are formed in quantum-state space.} \textbf{a}, Bloch vectors of one qubit for 2,000 developing-cortex cells (400 per type): every vector has unit length, and the types occupy different arcs. \textbf{b}, For 80,000 Siletti cell pairs (gray, random; blue, 15 nearest neighbors), the log-infidelity $-\log F$ against the feature distance $\frac{1}{4}\|\Delta f\|^2$; every pair lies above the lower bound of Theorem~\ref{thm:state} (dashed). \textbf{c}, AMI at the annotated number of types on the developing cortex (blue) and the Siletti atlas (orange) for graphs built from scanpy's principal components and the whitened scores (gray) and from the exact 132-qubit fidelity, the Bloch vectors and all quantum features (bars, mean of three Leiden seeds; dots, seeds; Extended Data Table~2). All quantum features reach 0.766 and 0.935, against 0.188 and 0.815 for the whitened scores.}
\label{fig:qstate}
\end{figure}

\begin{table}[h]
\centering\footnotesize
\caption*{\textbf{Extended Data Table 1 $|$ KQE-2 runs on every organism, assay and scale of this study and gives the same result on every run.} Every entry refers to an observation of this study (Results; Extended Data Table~5; Supplementary Table~12). Dedicated methods: scICE, scDCCA, scAGCL and scDAC.}
\label{tab:scope}
\setlength{\tabcolsep}{3pt}
\begin{tabular}{P{2.1cm}P{2.45cm}P{2.45cm}P{2.45cm}P{2.45cm}P{2.3cm}}
\toprule
 & KQE-2 & scGPT & scVI & scanpy, Seurat & Dedicated methods \\
\midrule
Organism & human, mouse, zebrafish & human only (whole-human checkpoint) & any, retrained per atlas & any & any \\
Assay & 10x 3$'$ and 5$'$, sci-RNA-seq3, Smart-seq; cells and nuclei & set by pretraining & any & any & any \\
Scale & 0.61 to 11.4 million cells, one server & 29 to 39 min per atlas for the embedding & retrained on every atlas; one run on 2.86 million nuclei & scanpy runs; Seurat cannot read matrices above $2^{31}$ nonzeros & 146,598 cells not completed in 3 h and 150 GB \\
Genes & whole transcriptome & 1,200 sampled per cell & 2,000 HVGs; the whole transcriptome lowers the accuracy (0.908 to 0.771) & 2,000 HVGs & 2,000 to 4,000 HVGs \\
Training & none & pretrained; four atlases predate the snapshot & retrained for every atlas & none & trained on every dataset \\
Same result on rerun & yes, bit for bit & no (ARI 0.63 between embeddings) & no (ARI 0.58 to 0.75 between seeds) & yes with a fixed Leiden seed; HVG selection removes the same markers each time & no (random initialization) \\
Donor correction & deterministic (quantum alignment) & none & batch covariate in training & Harmony, seed-dependent (F1 0.98, 0.98, 0.64) & none \\
Quantum hardware & runs, audited against exact values & n.a. & n.a. & n.a. & n.a. \\
Labels used & number of clusters only & same & same & same & same \\
\bottomrule
\end{tabular}
\end{table}

\begin{table}[h]
\centering\small
\caption*{\textbf{Extended Data Table 2 $|$ Clustering in quantum-state space is more accurate than clustering the whitened scores.} AMI at the annotated number of types, mean over three Leiden seeds (range in parentheses), on the 150,000-cell subsamples. Every graph uses 15 nearest neighbors and the same Leiden protocol; only the distance changes. State fidelity: $-\log F$ between the 132-qubit states of Theorem~\ref{thm:state}. Bloch vectors: Euclidean distance between the single-qubit Bloch vectors, which equals $\sqrt{2}$ times the Hilbert--Schmidt distance of the states. All features: Bloch vectors and within-block $\langle Z_i Z_j \rangle$, the encoding of Table~\ref{tab:main}. On the Siletti atlas, the exact fidelity graph reaches 0.925 against 0.815 for the whitened scores.}
\label{tab:state}
\setlength{\tabcolsep}{5pt}
\begin{tabular}{lcc}
\toprule
Graph built from & Developing cortex ($k = 5$) & Siletti ($k = 13$) \\
\midrule
scanpy principal components & 0.600 (0.598--0.601) & 0.810 (0.801--0.823) \\
Whitened scores & 0.188 (0.172--0.199) & 0.815 (0.773--0.869) \\
\midrule
Quantum-state fidelity & 0.485 (0.181--0.638) & 0.925 (0.925--0.926) \\
Bloch vectors & 0.678 (0.637--0.758) & 0.932 (0.927--0.941) \\
All features of the encoding & \textbf{0.766} (0.766--0.766) & \textbf{0.935} (0.935--0.935) \\
\bottomrule
\end{tabular}
\end{table}

\begin{table}[h]
\centering\footnotesize
\caption*{\textbf{Extended Data Table 3 $|$ Paired tests show that the gains of KQE-2 are systematic.} \textbf{a}, AMI margin of KQE-2 (Table~\ref{tab:main}) over the best other method and over each method, on the atlases where that method ran; one-sided Wilcoxon signed-rank test over atlases. \textbf{b}, Macro-averaged F1 of the rare types of each atlas; the F1 of a type is taken in the cluster holding most of its cells; n.a., no per-type result. The developing cortex has no rare type. Bottom rows, one-sided Wilcoxon signed-rank test over the rare types, stratified by atlas (van Elteren).}
\label{tab:tests}
\setlength{\tabcolsep}{4pt}
\begin{tabular}{lrrrr}
\multicolumn{5}{l}{\textbf{a} \quad AMI over atlases}\\
\toprule
Comparison & atlases & mean margin & atlases higher & $P$ \\
\midrule
Best other method & 10 & $+0.050$ & 8 (2 within range) & 0.0049 \\
scanpy & 10 & $+0.087$ & 9 & 0.0020 \\
Seurat & 5 & $+0.104$ & 5 & 0.031 \\
scVI & 10 & $+0.065$ & 9 & 0.0020 \\
scGPT & 6 & $+0.099$ & 6 & 0.016 \\
\bottomrule
\end{tabular}

\vspace{8pt}
\begin{tabular}{lrccccc}
\multicolumn{7}{l}{\textbf{b} \quad Rare types}\\
\toprule
Atlas & rare types & scanpy & Seurat & scVI & scGPT & KQE-2 \\
\midrule
Zebrafish & 82 & 0.281 & 0.278 & 0.330 & n.a. & \textbf{0.387} \\
Tabula Sapiens & 148 & n.a. & n.a. & 0.274 & 0.278 & \textbf{0.282} \\
Fetal & 53 & 0.241 & 0.276 & 0.334 & 0.355 & \textbf{0.476} \\
Optic nerve & 40 & \textbf{0.567} & 0.565 & 0.552 & 0.494 & 0.558 \\
Siletti non-neuronal & 8 & \textbf{0.675} & 0.561 & 0.672 & 0.563 & 0.658 \\
Mouse embryo & 119 & 0.316 & n.a. & 0.392 & n.a. & \textbf{0.486} \\
Mouse cerebellum & 8 & 0.893 & n.a. & 0.897 & n.a. & \textbf{0.972} \\
RADC brain & 11 & 0.549 & n.a. & 0.551 & \textbf{0.553} & 0.547 \\
Mouse cortex & 11 & 0.550 & n.a. & 0.535 & n.a. & \textbf{0.600} \\
\midrule
Rare types compared & & 332 & 183 & 480 & 260 & \\
Mean F1 gain of KQE-2 & & $+0.127$ & $+0.110$ & $+0.054$ & $+0.039$ & \\
$P$ & & $1.1 \times 10^{-26}$ & $1.1 \times 10^{-14}$ & $2.1 \times 10^{-8}$ & 0.043 & \\
\bottomrule
\end{tabular}
\end{table}

\begin{table}[h]
\centering\small
\caption*{\textbf{Extended Data Table 4 $|$ Whitening and the quantum encoding together keep the oligodendrocytes in one cluster.} Siletti non-neuronal brain atlas, 150,000-cell subsample, 83,112 annotated oligodendrocytes. $R^2$: five-fold cross-validated ridge regression from the embedding to the percentage of mitochondrial counts and to $\log_{10}$ counts per cell, within oligodendrocytes (20,000 cells). Clusters: Leiden at the annotated $k$ (seed 0), clusters holding at least 5\% of the oligodendrocytes; MT, mitochondrial; MT ratio, median mitochondrial percentage of the second-largest over the largest. AMI over all 13 types (Leiden seed 0). KQE-2 keeps one oligodendrocyte cluster and reaches the highest AMI, 0.935.}
\label{tab:ablation}
\setlength{\tabcolsep}{4pt}
\begin{tabular}{lccccc}
\toprule
Embedding & $R^2$ (\%MT) & $R^2$ (counts) & Clusters & MT ratio & AMI \\
\midrule
scanpy (2,000 HVGs) & 0.68 & 0.56 & 3 & 8.16 & 0.801 \\
scanpy, no MT genes & 0.14 & 0.55 & 3 & 1.42 & 0.838 \\
Scores, IDF, unwhitened & 0.73 & 0.99 & 2 & 0.23 & 0.797 \\
Scores, IDF, whitened & 0.73 & 0.99 & 1 & -- & 0.803 \\
Scores, no IDF, whitened & 0.69 & 0.98 & 1 & -- & 0.846 \\
KQE-2 (IDF, whitened, encoding) & 0.74 & 0.98 & 1 & -- & \textbf{0.935} \\
Encoding, no IDF, whitened & 0.76 & 0.98 & 1 & -- & 0.866 \\
\bottomrule
\end{tabular}
\end{table}

\begin{table}[h]
\centering\small
\caption*{\textbf{Extended Data Table 5 $|$ KQE-2 clusters 146,598 cells in 11 minutes, beyond the reach of dedicated clustering methods.} Wall-clock time of one run from raw counts to cluster labels. Limits: 3 hours and 150 GB of memory per run. Dedicated clustering methods ran on a shared server (NVIDIA RTX A6000, 40 CPU cores, 502 GB memory); KQE-2 on one NVIDIA RTX 6000 Ada GPU, whole transcriptome, deterministic decomposition, from raw counts to cluster labels at the annotated number of types. Genes: input used by each method.}
\label{tab:scale}
\setlength{\tabcolsep}{5pt}
\begin{tabular}{llccc}
\toprule
Method & Genes & 20,000 cells & 50,000 cells & 146,598 cells \\
\midrule
scICE & whole transcriptome (scLENS) & 71 min & 2.7 h & beyond 3 h \\
scDCCA & 2,000 HVGs & 13 min & 1.8 h & beyond 3 h \\
scAGCL & 2,000 HVGs & 51 min & 2.3 h & dense $n \times n$ matrix \\
scDAC & 4,000 HVGs & 6.4 h & beyond 150 GB & beyond 150 GB \\
\midrule
KQE-2 & whole transcriptome & 1.9 min & 4.3 min & 11 min \\
\bottomrule
\end{tabular}
\end{table}

\clearpage
\section*{Supplementary Information}
\begin{multicols}{2}
\end{multicols}
\begin{figure}
\centering
\includegraphics[width=\textwidth]{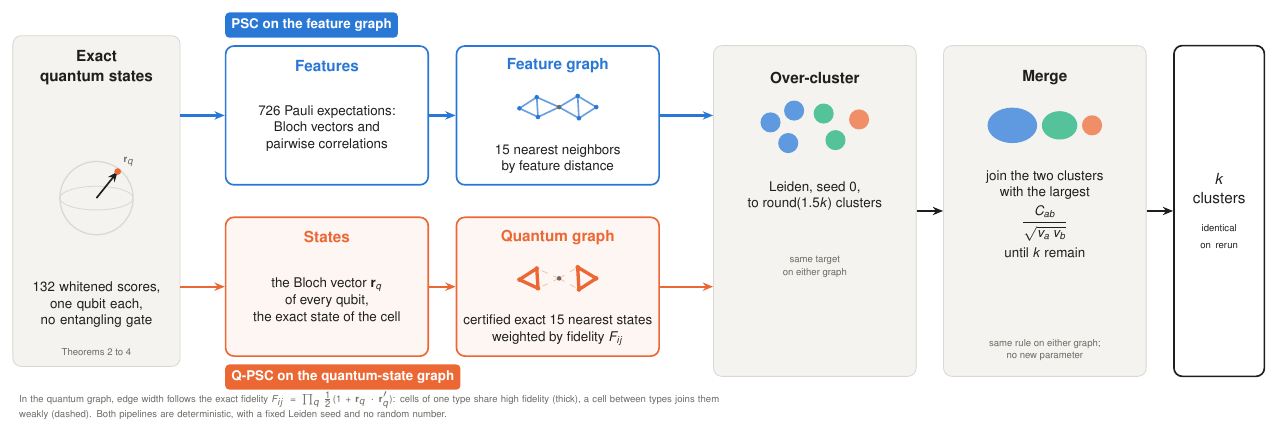}
\caption*{\textbf{Supplementary Fig. 1 $|$ Quantum precision shrinkage clustering (Q-PSC).} Both pipelines start from the exact quantum states of KQE-2: 132 whitened scores, one qubit each, no entangling gate (Theorems~\ref{thm:var} to \ref{thm:state}). Top, for comparison, precision shrinkage clustering (PSC): the same steps on the feature graph, the 15 nearest neighbors by distance between the 726 features. Bottom, Q-PSC on the quantum-state graph, the certified exact 15 nearest states with edges weighted by the fidelity $F_{ij}$; KQE-2 uses it on whole-transcriptome atlases (Table~\ref{tab:main}). On either graph, Leiden (seed 0) divides the cells into $\mathrm{round}(1.5k)$ clusters, and the pair of clusters with the largest $C_{ab}/\sqrt{v_a v_b}$ is merged until $k$ clusters remain (Methods). Every step is deterministic.}
\end{figure}
\begin{figure}[h]
\centering
\includegraphics[width=\textwidth]{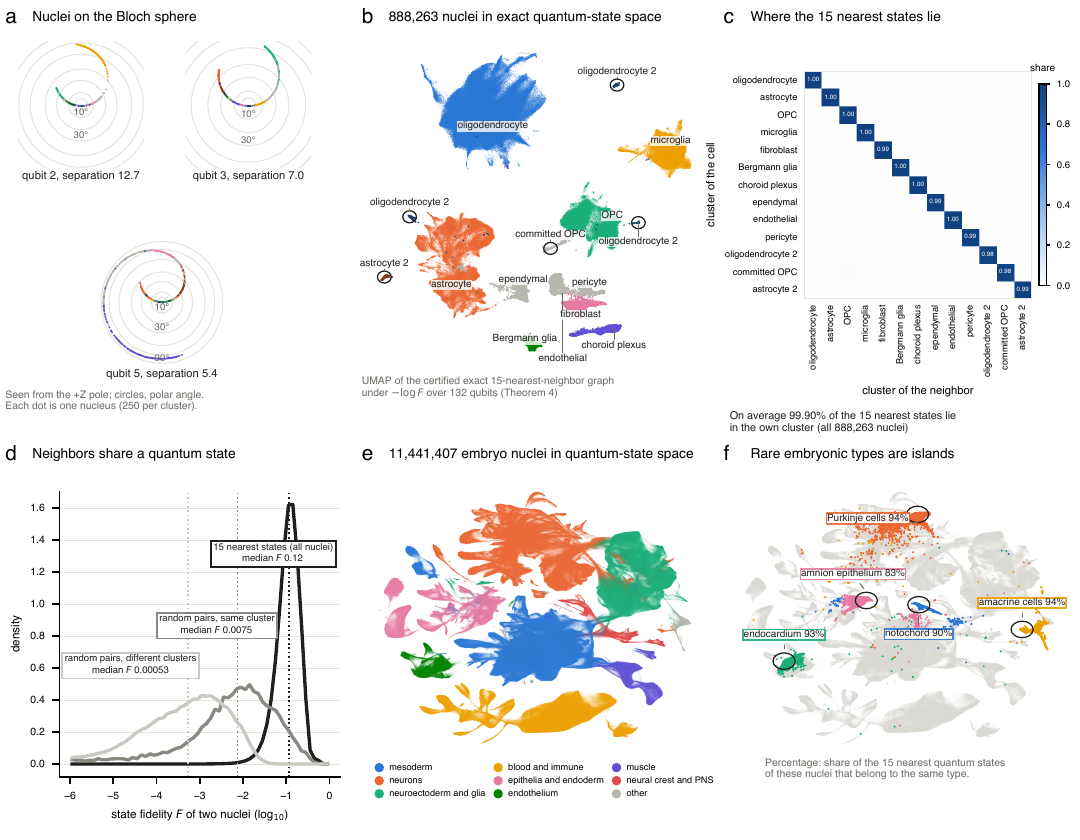}
\caption*{\textbf{Supplementary Fig. 2 $|$ A complete atlas in exact quantum-state space: each KQE-2 cluster is a region of nearby quantum states.} Siletti non-neuronal brain atlas, 888,263 nuclei, 13 KQE-2 clusters (Table~\ref{tab:main}, AMI 0.963); every quantity is computed from the exact 132-qubit states. \textbf{a}, Bloch vectors of the three qubits that separate the clusters most (between-cluster over within-cluster variance), seen from the $+Z$ pole; 250 nuclei per cluster. Each qubit traces a bounded path, and the clusters occupy different arcs of it. \textbf{b}, Map of all 888,263 nuclei, built from the exact 15-nearest-neighbor graph under the state distance $-\log F$; circles mark the three smallest clusters. \textbf{c}, Share of the 15 nearest states of the nuclei of each cluster that lie in each cluster: on average, 99.90\% lie in the own cluster. \textbf{d}, State fidelity of nearest states (median 0.12) and of random pairs of nuclei in the same cluster (0.0075) and in different clusters (0.00053). \textbf{e}, All 11,441,407 nuclei of the complete mouse embryo atlas, mapped from their exact 15-nearest-neighbor graph and colored by eight lineages merged from the author annotations. \textbf{f}, The same map: the notochord, Purkinje cells, endocardium, amacrine cells and amnion epithelium form islands of nearby quantum states; percentages, share of the 15 nearest states of their nuclei that belong to the same type.}
\label{fig:qspace}
\end{figure}
\begin{multicols}{2}
\paragraph{Supplementary Note 1: the encoding under class imbalance.} To test the encoding under class imbalance, we lowered the share of OPCs in the Siletti atlas from their native 11.9\% (105,734 cells) to 2\% and 0.5\% (3,932 cells) while keeping every other cell. The encoding gave the highest AMI at every share (0.928, 0.916 and 0.911), ahead of the same whitened scores (0.915, 0.902 and 0.887), $\tanh$ (0.909 to 0.887), RFF (0.849 to 0.907) and scanpy (0.817 to 0.782). The margin of the encoding over its own whitened scores widened from 0.013 to 0.024 as the OPCs became rarer. The gain comes from the abundant types, which the encoding keeps from fragmenting, while the OPCs themselves stayed well recovered at 2\% and 0.5\% ($F_1$ 0.95 and 0.83). On the complete RADC atlas, lowering the share of microglia from 4.4\% to 2\% and 0.5\% gave 0.951, 0.946 and 0.944 for the encoding, against 0.938 to 0.940 for the whitened scores, 0.941 to 0.947 for RFF and 0.887 to 0.932 for scanpy.

\paragraph{Supplementary Note 2: a worked example of the state distances.}
Take two cells with two scores each, $x_A = (0.5, -1.2)$ and $x_B = (1.4, -1.0)$, and the encoding of the Methods with $T = 3$ and $s = 0.19$. Each score is turned into a single-qubit state $(a, b)$ by three layers of $R_Y$ and $R_Z$ rotations, and its Bloch vector is $\mathbf{r} = (2\,\mathrm{Re}(\bar a b),\, 2\,\mathrm{Im}(\bar a b),\, |a|^2 - |b|^2)$. Supplementary Table~1 lists every intermediate quantity. On qubit 1, $d_1^2 = 0.09675$; the Hilbert--Schmidt distance of the two density matrices is $0.04838 = d_1^2/2$, and the fidelity of the two single-qubit states, computed from the amplitudes, is $0.97581 = 1 - d_1^2/4$. On qubit 2 the same identities hold with $d_2^2 = 0.00521$. The two-qubit fidelity is the product $F = 0.97454$, so $-\log F = 0.02579$. The feature distance gives the lower bound $\tfrac{1}{4}\|\Delta f\|^2 = 0.02549$, and the upper bound of Eq.~(\ref{eq:state}) is $0.02609$; the state distance lies between them, as Theorem~\ref{thm:state} states.

\end{multicols}
\begin{table}
\centering\small
\caption*{\textbf{Supplementary Table 1 $|$ In a worked example, the state distance lies between the two bounds of Theorem~\ref{thm:state}.} Two cells, two qubits, $T = 3$, $s = 0.19$; $0.02549 \le -\log F = 0.02579 \le 0.02609$. HS, squared Hilbert--Schmidt distance $\|\rho_A - \rho_B\|^2_{\mathrm{HS}}$; $F_q$, single-qubit fidelity from the amplitudes.}
\label{tab:example}
\setlength{\tabcolsep}{4pt}
\begin{tabular}{lcccccc}
\toprule
Qubit & $x_A$, $x_B$ & $\mathbf{r}_A$ & $\mathbf{r}_B$ & $d_q^2$ & HS & $F_q$ \\
\midrule
1 & 0.5, 1.4 & (0.161, 0.038, 0.986) & (0.350, 0.268, 0.898) & 0.09675 & 0.04838 & 0.97581 \\
2 & $-1.2$, $-1.0$ & ($-0.325$, 0.203, 0.924) & ($-0.289$, 0.145, 0.946) & 0.00521 & 0.00261 & 0.99870 \\
\midrule
\multicolumn{7}{l}{$F = 0.97581 \times 0.99870 = 0.97454$; \ $\tfrac{1}{4}\|\Delta f\|^2 = 0.02549 \le -\log F = 0.02579 \le 0.02609$} \\
\bottomrule
\end{tabular}
\end{table}
\begin{multicols}{2}

\paragraph{Supplementary Note 3: atlases released before the pretraining of scGPT.}
The whole-human model of scGPT was pretrained on the normal cells of the CZ CELLxGENE Census, version of 8 May 2023 \cite{cui2024,cellxgene2025}. Supplementary Table~2 gives, for each human atlas of this study, the date on which its collection was first released on CELLxGENE Discover (curation interface, accessed 17 October 2026). Four atlases were public before the pretraining snapshot, so scGPT may have seen their normal cells during pretraining, which can only raise its scores. The other three were released after the snapshot. KQE-2 is not pretrained and sees every atlas for the first time.

\end{multicols}
\begin{table}
\centering\small
\caption*{\textbf{Supplementary Table 2 $|$ Four human atlases were public before the pretraining snapshot of scGPT (8 May 2023), and KQE-2 was more accurate than scGPT on the three of them benchmarked in Table~\ref{tab:main}.} First release of the CELLxGENE collection; for the fetal and Siletti atlases the evaluated file is a subset whose cells are in the collection's primary datasets.}
\label{tab:leak}
\begin{tabular}{llll}
\toprule
Atlas & First release & Before snapshot & Cells that scGPT may have seen \\
\midrule
Tabula Sapiens \cite{tabula2022} & 19 Jul 2021 & yes & all cells (healthy donors) \\
Fetal \cite{cao2020} & 3 Jun 2021 & yes & normal cells \\
Heart \cite{reichart2022} & 5 Aug 2022 & yes & cells of control donors \\
Siletti non-neuronal \cite{siletti2023} & 9 Dec 2022 & yes & all cells (normal) \\
Developing cortex \cite{velmeshev2023} & 19 Sep 2023 & no & none \\
Optic nerve \cite{opticnerve2025} & 5 May 2025 & no & none \\
RADC \cite{lee2024radc} & 29 Jul 2025 & no & none \\
\bottomrule
\end{tabular}
\end{table}
\begin{multicols}{2}

\paragraph{Supplementary Note 4: markers and ablation of the oligodendrocyte split.} Each KQE-2 cluster is marked by the canonical genes of its type (Extended Data Fig.~3). The precursor clusters express \textit{PDGFRA}, \textit{CSPG4} and \textit{OLIG2}, the microglia clusters \textit{P2RY12}, \textit{CX3CR1} and \textit{CSF1R}, and the oligodendrocyte clusters \textit{MBP}, \textit{PLP1} and \textit{MOBP}. The small Siletti clusters of 837 to 1,332 cells carry \textit{GDF10} (Bergmann glia), \textit{TTR} and \textit{KCNJ13} (choroid plexus), \textit{FOXJ1} (ependymal cells), \textit{CLDN5} and \textit{FLT1} (endothelial cells) and \textit{PDGFRB} and \textit{RGS5} (pericytes). The oligodendrocyte clusters into which the other methods divide this type differ in technical quality rather than in identity (Extended Data Fig.~2). On the complete Siletti atlas, the second scanpy cluster has a median mitochondrial read fraction 7.6 times that of the largest one (0.497\% against 0.065\%). The oligodendrocyte clusters of Seurat and scGPT differ 3.1- and 4.4-fold in mitochondrial reads and 1.5- and 2.5-fold in counts per cell. KQE-2 keeps 99.6\% of the oligodendrocytes of the displayed cells (82,746 of 83,112) in one cluster. Significance analysis can detect and merge such clusters after clustering \cite{grabski2023}. KQE-2 avoids the split at the embedding step. An ablation shows how (Extended Data Table~4). Deleting the 11 mitochondrial genes from scanpy's 2,000 HVGs still leaves three oligodendrocyte clusters. KQE-2 keeps the full quality information in its scores (a ridge regression recovers log counts with $R^2 = 0.99$, against 0.56 for scanpy), and whitening keeps it from dominating the nearest-neighbor graph: unwhitened scores divide the oligodendrocytes into two clusters, and whitened scores keep them in one. Whitening alone merges the OPCs into the oligodendrocytes (AMI 0.803), and the quantum encoding separates them again while keeping one oligodendrocyte cluster (AMI 0.935).

The encoding widens the margin between types, as Proposition~\ref{prop:sat} predicts: on the Siletti atlas, the distance from each major glial type to its nearest other type, relative to its spread, rose from 0.24--0.33 to 0.40--0.46 (Methods).

\paragraph{Supplementary Note 5: details of the reproducibility analyses.}
\textit{Tests of the theorems.} We tested each part on 150,000 cells of the developing cortex and the Siletti atlas. We compared two runs of a randomized decomposition with different seeds, over 19.4 and 19.8 million cell-qubit pairs. Every Bloch vector moved by at most 0.70 and 0.71 times the bound $\Lambda\,|\xi|$ of Theorem~\ref{thm:var}. Under Gaussian perturbations of the scores, the largest variance of any feature is 0.57 to 0.75 of its bound. Without entanglement, perturbing the other five scores of a block or inserting an $X$ error on a neighboring qubit changes a qubit's features by at most $1.3 \times 10^{-15}$. With controlled-$Z$ chains, the same perturbation moves them with a standard deviation of $7 \times 10^{-4}$ to $9 \times 10^{-4}$. The same error moves them by a mean of 0.23 to 0.27, as Theorem~\ref{thm:local} states.

\textit{From features to clusters.} We then asked whether the guarantee reaches the clusters (Corollary~\ref{cor:repro}). The two decompositions give scores that differ in the directions beyond the signal, yet the encoding gives nearly the same clusters in both runs. The adjusted Rand index between the runs is 0.90 and 0.99, against 0.43 and 0.81 for the scores themselves. Across 45 runs with three Leiden seeds each, the AMI of the encoding varied across seeds by a median of 0.005, against 0.011 for scanpy and 0.016 for the scores. The encoding was more stable than scanpy in 32 of the 45 runs. On the complete fetal atlas, three scVI runs scored an AMI of 0.716 to 0.720 (Leiden seed 0), and two runs agreed at an adjusted Rand index of 0.73 to 0.74.

\textit{Megakaryocytes and the allocation of clusters.} For 10 of 68 types the two runs shared fewer than half of their top markers, and these were mostly rare types. Under seed 0 the megakaryocytes (832 cells, 0.09\% of the atlas) fell into a cluster marked by hemoglobin genes (\textit{HBG2}, \textit{HBA1}), whose top pathway was gas transport. Under seed 2 they formed their own cluster (F1 0.86 against 0.11), marked by \textit{ITGA2B}, \textit{PF4} and \textit{PLEK}, with platelet aggregation and activation as top pathways. Under seed 0 the lymphatic endothelial cells (1,699 cells) merged with blood endothelium, marked by \textit{FLT1} and \textit{EMCN}. Under seed 2 they formed their own cluster (F1 0.92 against 0.18), marked by \textit{PROX1}, \textit{CCL21} and \textit{STAB2}, with lymphatic endothelial differentiation as a top pathway. The megakaryocyte cluster of seed 2 carries \textit{ITGA2B}, the target of the antiplatelet drugs abciximab, eptifibatide and tirofiban. It also carries \textit{PDE3A}, the target of anagrelide; the cluster of seed 0 carries neither gene. KQE-2 gives its megakaryocytes their own cluster (F1 0.78). scanpy merges both the megakaryocytes and the lymphatic endothelial cells (F1 0.10 and 0.18). HVG selection removes 58\% of the megakaryocyte markers, including \textit{PDE3A} (150,000-cell subsample). In addition, at the annotated $k$ scanpy spends 28 of its 71 clusters on glutamatergic neurons, the largest type, against 13 for KQE-2. As a result, KQE-2 gives six more types a cluster of their own than scanpy. In scanpy, the megakaryocytes fall into a cluster of 13,901 myeloid and lymphoid cells, and the lymphatic endothelial cells into vascular endothelium.

\textit{Hardware.} On IBM ibm\_marrakesh we ran one 114-qubit circuit per cell for 380 developing-cortex cells (76 per type). The measured features reproduce the exact ones with correlation 0.978 (mean absolute error 0.050), and a 50-cell run on 132 qubits gives correlation 0.953. These runs include controlled-$Z$ chains, a harder test than the entanglement-free encoding, which needs no two-qubit gate. Because the features of the encoding are exact on a classical computer, every hardware number can be checked against an exact reference, and atlas-scale results use the exact evaluation.

Across 42 comparisons, adding chains of controlled-$Z$ entangling gates stayed within 0.01 AMI of the entanglement-free circuit in 31 cases and was ahead in 6.

\paragraph{Supplementary Note 6: which marker genes HVG selection removes.}
The loss matters most when a whole type disappears into a neighboring cluster, because a type without a cluster has no markers to find. On the fetal atlas, scanpy places fewer than half of the cells of five types in one cluster, and KQE-2 on the whole transcriptome places more than half. These types are amacrine cells, retinal ganglion cells, inhibitory interneurons, lens fiber cells and neuroendocrine cells (5,327 cells). Their markers include 15 established drug targets, 12 of them outside the HVGs, among them \textit{GRIA4} (perampanel) and \textit{PDE4D} (apremilast).

In the developing cortex, HVG selection removes \textit{SCN1A}, the Dravet syndrome gene, and the glutamate receptors \textit{GRIA2} and \textit{GRIK2} from the markers of OPCs.

\paragraph{Supplementary Note 7: proofs.}
\textit{Proof of Theorem~\ref{thm:qubits}.} (a) The shares $u_i$ of $\Psi$ form a probability vector with $\operatorname{rank}(\Psi)$ nonzero entries, and the Shannon entropy of such a vector is at most the logarithm of the number of its nonzero entries. $\Psi$ has $n$ rows and $2^{l}$ columns, so $\operatorname{rank}(\Psi) \le \min\{n, 2^{l}\}$. (b) The assumption gives $\Psi \Psi^{\dagger} = c\, X X^{\top}$. The nonzero eigenvalues of both sides are the squared singular values, so $\sigma_i(\Psi) = \sqrt{c}\, \sigma_i(X)$ for every $i$ and $\operatorname{rank}(\Psi) = r$. A common factor cancels in $u_i$, hence $h(\Psi) = h(X)$, and (a) gives $h(X) \le \log_2 r \le l$. Equal row norms make $c \|x_i\|^2 = 1$ consistent with unit states. (c) $\operatorname{Tr} \rho^{2} = \tfrac{1}{n^{2}} \operatorname{Tr}(\Psi^{\dagger} \Psi \Psi^{\dagger} \Psi) = \tfrac{1}{n^{2}} \|\Psi \Psi^{\dagger}\|_F^{2} = \tfrac{1}{n^{2}} \sum_{i,j} |G_{ij}|^{2}$. The eigenvalues $\lambda_k \ge 0$ of $\rho$ sum to one over at most $2^{l}$ indices, so by the Cauchy--Schwarz inequality $\sum_k \lambda_k^{2} \ge 2^{-l}$, with equality only when all $2^{l}$ eigenvalues equal $2^{-l}$.

\textit{Proof of Theorem~\ref{thm:var}.} (a) The state is a tensor product, so the expectation of an operator acting on qubit $q$ depends only on $V(x_q)$, and the expectation of $Z_i \otimes Z_j$ factorizes. The entries of each rotation are cosines and sines of half its angle, so $g_P$ is a trigonometric polynomial. (b) The circuit $V(x)$ is a product of $2T$ gates $e^{-i a_k x G_k / 2}$ with Pauli generators $G_k$ and coefficients $a_k \in \{s, \beta_1 s, \dots, \beta_T s\}$. Differentiating gate by gate gives $|g_P'(x)| \le \sum_k |a_k|\, \|P\| = \Lambda$, so $g_P$ is $\Lambda$-Lipschitz. For an independent copy $\xi'$ of $\xi$, $\mathrm{Var}[f(\xi)] = \tfrac{1}{2}\mathbb{E}\big[(f(\xi) - f(\xi'))^2\big] \le \tfrac{1}{2}\Lambda^2\, \mathbb{E}\big[(\xi_q - \xi_q')^2\big] = \Lambda^2 \mathrm{Var}[\xi_q]$. The product $g_Z(x_i) g_Z(x_j)$ has gradient norm at most $\sqrt{2}\Lambda$ because $|g_Z| \le 1$, and the same identity gives the second bound. (c) For any state $\rho$ and Pauli operator $P$ with $\|P\| = 1$, $|\mathrm{Tr}(P\, \mathcal{N}(\rho)) - \mathrm{Tr}(P \rho)| \le \|\mathcal{N}(\rho) - \rho\|_1 \le 2\varepsilon$. Without entangling gates the noisy state remains a product $\bigotimes_q \mathcal{N}_q(\rho_q)$, so $\langle P_q \rangle$ sees only $\mathcal{N}_q$, and $|ab - a'b'| \le |a - a'| + |b - b'|$ for $|a|, |b'| \le 1$ gives the bound on $\langle Z_i Z_j \rangle$. $\square$

\textit{Proof of Theorem~\ref{thm:local}.} (a) is Theorem~\ref{thm:var}: the state is a product, so $f_{q,P}$ is a function of $x_q$ and of the channel acting on qubit $q$ alone. (b) Write $\Phi(O) = \Gamma^{\dagger}(O \otimes I)\Gamma$ for operators $O$ on qubit $q$. Suppose $\Phi(O) = O' \otimes I$ for every $O$. Then $\Phi$ is a unital $*$-homomorphism from the $2 \times 2$ matrices into themselves, so by the Skolem--Noether theorem $\Phi(O) = V^{\dagger} O V \otimes I$ for a unitary $V$. The unitary $\Gamma (V^{\dagger} \otimes I)$ then commutes with every $O \otimes I$, so it lies in the commutant $I \otimes M_2$ and equals $I \otimes \Gamma_{q'}$; hence $\Gamma = V \otimes \Gamma_{q'}$, a product. Therefore, if $\Gamma$ is not a product, some Pauli operator $P$ has $\Phi(P) = \sum_{\tau} O_{\tau} \otimes \tau$ with $O_{\tau} \neq 0$ for some $\tau \in \{X, Y, Z\}$. For product inputs $\rho_q \otimes \rho_{q'}$, $\langle P_q \rangle = \sum_{\tau} \mathrm{Tr}(O_{\tau} \rho_q)\, \mathrm{Tr}(\tau \rho_{q'})$, where $\tau$ runs over $I, X, Y, Z$. Choose $\rho_q$ as an eigenvector of $O_{\tau}$ with nonzero eigenvalue, so that $\mathrm{Tr}(O_{\tau}\rho_q) \neq 0$. Then $\langle P_q \rangle$ varies with the Bloch component $\mathrm{Tr}(\tau \rho_{q'})$ of qubit $q'$, which violates (D) when $\rho_{q'}$ is set by $x_{q'}$. A Pauli error on $q'$ that anticommutes with $\tau$ reverses the sign of that component, which violates (N). For the controlled-$Z$ gate, $\mathrm{CZ}\,(X \otimes I)\,\mathrm{CZ} = X \otimes Z$ gives $\langle X_q \rangle = \langle X \rangle_{q} \langle Z \rangle_{q'}$ explicitly. (c) Every feature is a trigonometric polynomial in $x$, hence real analytic, so a dependence on $x_{q'}$ at one point holds on a set of nonzero measure. We evaluated the circuit exactly with 50 significant digits at $x = (1.3, -0.7, 2.1, 0.4, -1.6, 0.9)$. Raising $x_2$ by 1 changes $\langle X_1 \rangle$ by $3.0 \times 10^{-3}$ ($s = 0.19$) and $5.3 \times 10^{-3}$ ($s = 0.25$) with controlled-$Z$ chains, and by less than $10^{-50}$ without them. A Pauli-$X$ error on qubit 2 before the last controlled-$Z$ layer changes $\langle X_1 \rangle$ by $-0.598$ and $-0.594$ with the chains, and by less than $10^{-50}$ without them. $\square$

\textit{Proof of Corollary~\ref{cor:repro}.} (a) Each gate $e^{-i a_k x G_k/2}$ rotates the Bloch vector of its qubit by the angle $a_k x$ about a fixed axis, so $\|\mathrm{d} \mathbf{r}_q / \mathrm{d} x_q\| \le \sum_k |a_k| = \Lambda$, and $\mathbf{r}_q$ is $\Lambda$-Lipschitz in its own score. Without entanglement, $f_i$ is the concatenation of the $\mathbf{r}_q(x_{iq})$, which gives (a). (b) follows from (a) and the triangle inequality. (c) By (b), every distance from cell $i$ changes by at most $2\Lambda \max_j \|\xi_j\| < \delta_i / 2$, so no neighbor outside the first $K$ can overtake one inside, and the neighbor set of $i$ is unchanged. (d) With $\xi_i = 0$, every feature, single- or two-qubit, is the same function of the same scores, so the distances, the neighbor sets and the edge weights, fuzzy or fidelity, are identical; Leiden with a fixed seed and a $K$-nearest-neighbor classifier are deterministic functions of the weighted graph. $\square$

\textit{Proof of Theorem~\ref{thm:state}.} (a) A pure single-qubit state is determined by its Bloch vector, which has unit length, and a product state is the tensor product of its factors. (b) $\langle \psi(x)|\psi(y)\rangle = \prod_q \langle \chi(x_q)|\chi(y_q)\rangle$, and for two single-qubit pure states $|\langle \chi|\chi'\rangle|^2 = \mathrm{Tr}(\rho\rho') = \tfrac{1}{2}(1 + \mathbf{r}\cdot\mathbf{r}')$. For unit vectors $\mathbf{r} \cdot \mathbf{r}' = 1 - \tfrac{1}{2}\|\mathbf{r} - \mathbf{r}'\|^2$, which gives $1 - d_q^2/4$. (c) $\rho_q(x) - \rho_q(y) = \tfrac{1}{2}(\mathbf{r}_q(x) - \mathbf{r}_q(y))\cdot\boldsymbol{\tau}$ and $\mathrm{Tr}(\tau_a\tau_b) = 2\,[a = b]$ give the Hilbert--Schmidt identity. If $d_q = 2$ for some $q$, the factor $1 - d_q^2/4$ vanishes and $F = 0$. Otherwise, with $u_q = d_q^2/4 \in [0, 1)$, $-\log F = -\sum_q \log(1 - u_q)$, and $u \le -\log(1 - u) \le u/(1-u)$ gives Eq.~(\ref{eq:state}), since $\sum_q u_q = \tfrac{1}{4}\|f(x) - f(y)\|^2$. (d) The overlap $\langle 0 | U(x)^{\dagger}U(y)|0\rangle$ is the amplitude of the all-zero outcome of the circuit $U(x)^{\dagger}U(y)$. $\square$

\paragraph{Supplementary Note 8: subtypes within annotated types.}
We asked whether KQE-2 separates cells that the annotation of an atlas groups together. On the fetal atlas (978,128 cells, 71 annotated types), we took every annotated type that is the majority of at least two KQE-2 clusters (Table~\ref{tab:main} partition, Q-PSC, Leiden seed 0) of at least 300 cells. For each such cluster we recorded its purity (share of cells of its majority type); its stability, the largest Jaccard index with a cluster of the KQE-2 partition obtained by Leiden at $k$ on the feature graph, a partition reached by a different clustering step; the finer annotations of the authors inside the cluster, namely their 657 numbered subclusters (\texttt{sub\_cluster\_name}), obtained by re-clustering each type, and the name of the matched mouse cell type (\texttt{Matched\_MCA\_cell\_name}); and its markers. Markers were found by the Wilcoxon test (scanpy) between up to 3,000 cells of the cluster and up to 3,000 other cells of the same annotated type (random seed 0), on library-size normalized log counts of the whole transcriptome, keeping genes with a log fold change above 1 and an adjusted $P$ below $10^{-10}$, ranked by the difference in the share of expressing cells. Supplementary Table~3 lists the clusters with stability of at least 0.7 whose markers identify a cell type. KQE-2 recovered five author subclusters at the atlas level, without labels and without re-clustering any type (Jaccard index 0.79 to 0.94). The authors left these subclusters numbered; the matched mouse names identify the MGE-derived interneurons (\emph{GABAergic neuron Lhx6 high}), the ventricular cardiomyocytes (\emph{left ventricle cardiomyocyte}) and the OPCs, but not the striatal projection neurons, the striosomal neurons, the CGE-derived interneurons or the \textit{OTX2}-positive neurons, whose identities follow here from canonical markers.

\end{multicols}
\begin{table}
\centering\small
\caption*{\textbf{Supplementary Table 3 $|$ Stable KQE-2 clusters of the fetal atlas within annotated types.} Percentages, share of expressing cells in the cluster and in the other cells of the same annotated type. Stability, largest Jaccard index with a cluster of the Leiden-at-$k$ partition. Author subcluster, the numbered author subcluster that holds most of the cells, with its Jaccard index; matched name, the authors' matched mouse cell type, given when it names the cluster.}
\setlength{\tabcolsep}{3pt}
\resizebox{\textwidth}{!}{\begin{tabular}{lrcclll}
\toprule
Annotated type & cells & purity & stability & author subcluster (Jaccard); matched name & markers & consistent with \\
\midrule
GABAergic neuron & 12,631 & 1.00 & 0.98 & Inhibitory neurons-1 (0.94) & \textit{EBF1} 86/12\%, \textit{RARB} 76/12\%, \textit{FOXP1} 83/27\%, \textit{PDE10A} 77/25\% & striatal projection neurons \\
GABAergic neuron & 529 & 0.99 & 0.96 & Inhibitory neurons-19 (0.82) & \textit{OPRM1} 75/5\%, \textit{FIGN} 88/11\%, \textit{KCNMB2} 98/20\% & striosomal projection neurons \\
GABAergic neuron & 12,502 & 1.00 & 0.97 & Inhibitory neurons-7 (0.83); \emph{Lhx6 high} & \textit{NXPH1} 81/14\%, \textit{SOX6} 60/9\%, \textit{ERBB4} 96/43\% & MGE-derived interneurons \\
GABAergic neuron & 8,989 & 0.98 & 0.78 & Inhibitory neurons-3 (0.79) & \textit{PDZRN3} 80/10\%, \textit{DLX6-AS1} 75/16\%, \textit{ADARB2} 70/22\% & CGE-derived interneurons \\
GABAergic neuron & 5,414 & 1.00 & 0.94 & Inhibitory neurons-9 (0.92) & \textit{OTX2-AS1} 97/2\%, \textit{CHRM2} 72/11\%, \textit{ZFPM2} 64/15\% & \textit{OTX2}-positive GABAergic neurons \\
cardiac muscle cell & 14,147 & 1.00 & 0.84 & Cardiomyocytes-2 (0.64); \emph{left ventricle} & \textit{MYL2} 68/8\%, \textit{MYH7} 80/23\%, \textit{MYL3} 70/16\% & ventricular cardiomyocytes \\
oligodendrocyte & 7,429 & 0.99 & 0.89 & several (0.24); \emph{OPC} & \textit{PDGFRA} 70/6\%, \textit{PCDH15} 92/39\% & oligodendrocyte precursor cells \\
\bottomrule
\end{tabular}}
\end{table}
\begin{multicols}{2}

\textit{Comparison with other methods.} On the same fetal cells, we clustered the embeddings of scanpy, Seurat and scVI with their standard protocol of Table~\ref{tab:main}, Leiden at $k$ (seed 0); scGPT exceeded the GPU memory in this run. For each KQE-2 cluster of Supplementary Table~3 we report the largest Jaccard index with any cluster of each partition (Supplementary Table~4). No other partition reached 0.7 for the striatal projection neurons (at most 0.55), the CGE-derived interneurons (at most 0.64) or the 529 striosomal neurons (at most 0.03). The MGE-derived interneurons, the \textit{OTX2}-positive neurons, the ventricular cardiomyocytes and the OPCs were also resolved by at least one other method.

\end{multicols}
\begin{table}
\centering\small
\caption*{\textbf{Supplementary Table 4 $|$ Largest Jaccard index of each KQE-2 subtype cluster with a cluster of the other methods (fetal atlas).} Leiden at $k$, seed 0.}
\begin{tabular}{lrccc}
\toprule
KQE-2 cluster & cells & scanpy & Seurat & scVI \\
\midrule
striatal projection neurons & 12,631 & 0.47 & 0.55 & 0.48 \\
striosomal neurons (\textit{OPRM1}) & 529 & 0.02 & 0.03 & 0.02 \\
MGE-derived interneurons & 12,502 & 0.74 & 0.84 & 0.50 \\
CGE-derived interneurons & 8,989 & 0.58 & 0.64 & 0.38 \\
\textit{OTX2}-positive GABAergic neurons & 5,414 & 0.64 & 0.85 & 0.82 \\
ventricular cardiomyocytes & 14,147 & 0.84 & 0.84 & 0.84 \\
OPCs & 7,429 & 0.83 & 0.88 & 0.84 \\
\bottomrule
\end{tabular}
\end{table}
\begin{multicols}{2}

\textit{Independent replication in the mouse embryo.} We repeated the search on the complete mouse embryo atlas (11,441,407 nuclei; KQE-2 with Q-PSC, 134 clusters), and compared each cluster with both the cell ontology label and the 190 finer cell-type labels of the authors (\texttt{author\_cell\_type}) \cite{qiu2024embryo}; a cluster is unnamed when both labels are shared with the rest of its type. Markers were computed on up to 3,000 nuclei per side as above. Of 78 clusters that split an annotated type, 32 had a stability of at least 0.7 and a purity of at least 0.8. Several followed developmental time rather than identity (for example, hepatocytes divided by embryonic day) and are not listed. Supplementary Table~5 lists the clusters whose markers identify a cell type and that no author label matches. The five neuronal clusters are labeled \emph{GABAergic cortical interneurons} for 95\% to 100\% of their nuclei even at the finest author level, whose only interneuron labels are this one and \emph{cortical interneurons (Prox1+)}; none of the 190 labels names striatal neurons. The two mesodermal clusters are labeled \emph{lateral plate and intermediate mesoderm} for 92\% and 98\% of their nuclei. Because no other method clustered the complete atlas, these clusters have no counterpart in another partition.

\end{multicols}
\begin{table}
\centering\small
\caption*{\textbf{Supplementary Table 5 $|$ Stable KQE-2 clusters of the complete mouse embryo atlas that neither the ontology label nor the 190 finer author labels name.} Annotated type, the ontology label. Percentages, share of expressing nuclei in the cluster and in the other nuclei of the same annotated type.}
\setlength{\tabcolsep}{3pt}
\resizebox{\textwidth}{!}{\begin{tabular}{lrcll}
\toprule
Annotated type & nuclei & stability & markers & consistent with \\
\midrule
cerebral cortex GABAergic interneuron & 100,222 & 0.95 & \textit{Ebf1} 76/11\%, \textit{Rarb} 66/9\%, \textit{Foxp2} 80/29\% & striatal projection neurons \\
cerebral cortex GABAergic interneuron & 48,046 & 0.94 & \textit{Drd2} 49/2\%, \textit{Adora2a} 14/0\%, \textit{Penk} 9/1\% & indirect-pathway striatal projection neurons \\
cerebral cortex GABAergic interneuron & 117,308 & 0.93 & \textit{Nxph1} 93/19\%, \textit{Lhx6} 55/6\%, \textit{Maf} 50/4\%, \textit{Sst} 12/1\% & MGE-derived interneurons \\
cerebral cortex GABAergic interneuron & 31,198 & 0.90 & \textit{Adarb2} 96/46\%, \textit{Nr3c2} 55/8\%, \textit{Zbtb16} 49/9\% & CGE-derived interneurons \\
cerebral cortex GABAergic interneuron & 32,505 & 0.77 & \textit{Lhx8} 68/3\%, \textit{Gbx1} 38/1\% & \textit{Lhx8}-positive basal forebrain lineage \\
lateral mesodermal cell & 74,402 & 0.81 & \textit{Wt1} 84/16\%, \textit{Msln} 32/0\%, \textit{Upk3b} 50/4\%, \textit{Muc16} 61/0\% & mesothelial cells \\
lateral mesodermal cell & 51,070 & 0.91 & \textit{Abcc9} 66/7\%, \textit{Rgs5} 36/4\%, \textit{Kcnj8} 22/2\%, \textit{Pdgfrb} 68/20\% & pericytes \\
\bottomrule
\end{tabular}}
\end{table}
\begin{multicols}{2}

On the complete mouse embryo atlas, KQE-2 separated striatal projection neurons (\textit{Ebf1}, \textit{Rarb}, \textit{Foxp2}) from an indirect-pathway cluster (\textit{Drd2} in 49\% of nuclei against 2\%, \textit{Adora2a} 14\% against 0\%) and from interneurons of medial (\textit{Lhx6}, \textit{Nxph1}) and caudal (\textit{Adarb2}) ganglionic eminence origin.

On the complete mouse embryo atlas, the endocardium cluster expresses \textit{Npr3} and \textit{Nfatc1} (59\% and 45\% of the nuclei, against 4\% and 4\%), and the amacrine cluster expresses \textit{Tfap2b} (63\% against 5\%).

Striatal projection neurons control movement and reward; they degenerate in Huntington's disease and lose their dopamine input in Parkinson's disease.

\paragraph{Supplementary Note 9: quantum k-means.}
Quantum k-means estimates the distance between each cell state and each centroid state on a quantum processor and assigns every cell to the nearest centroid \cite{kerenidis2019qmeans}. In the limit of exact distances, it returns the partition of classical k-means on the same features. We therefore ran k-means (k-means++ initialization, Lloyd iterations, scikit-learn) on the 726 quantum features of KQE-2 at the annotated $k$, with random seeds 0, 1 and 2. On the Siletti atlas, k-means reached an AMI of 0.646 to 0.652 and a macro-averaged rare-type F1 of 0.231 to 0.241, against 0.963 and 0.658 for KQE-2; its largest cluster held 19\% to 22\% of the nuclei, so the oligodendrocytes, 55\% of the atlas, were divided, and its partitions agreed between seeds at ARIs of 0.61 to 0.81. On the fetal atlas, k-means reached 0.652 to 0.660 (rare-type F1 0.227 to 0.289), against 0.821 (0.476) for KQE-2, with ARIs of 0.68 to 0.78 between seeds. We also ran quantum k-means with PennyLane 0.45 (default.qubit) on 1,040 Siletti nuclei (up to 80 per type), each represented by its first 6-qubit block, with the distance estimated by the swap test on 13 qubits. One swap test with 1,000 shots took 23 ms, against about 0.003 ms for the closed-form fidelity of the same pair, and its estimate of the fidelity varied with a standard deviation of 0.012 between shot draws. With 1,000 shots, one clustering of the 1,040 nuclei (13 centroids, 10 iterations, 135,200 swap tests) took 1.5 to 1.6 h, against 0.3 to 0.6 s with exact fidelities; three runs that differed only in the shot seed agreed at ARIs of 0.79 to 0.90 with the exact partition and 0.81 between one another, and 100 shots lowered the agreement to 0.69. One clustering of the Siletti atlas with 10 iterations would need $1.2 \times 10^{8}$ swap tests, 31 days for a single 6-qubit block, and the swap test of the complete 132-qubit states needs 265 qubits, beyond any simulator.

\paragraph{Supplementary Note 10: Secuer.}
Secuer \cite{wei2022secuer} selects anchor cells by random sampling and k-means, links every cell to its nearest anchors, clusters the resulting bipartite graph spectrally and assigns the clusters by k-means; its authors report clustering of a simulated dataset of 10 million cells. We ran it as in its paper, on the 50 principal components of 2,000 HVGs (the scanpy embedding of Table~\ref{tab:main}), with default parameters (1,000 anchors, 5 nearest anchors) and the annotated $k$, for random seeds 1, 2 and 3 (Supplementary Table~6). Its AMI was 0.06 to 0.24 below that of KQE-2 on every atlas, with the largest gaps on the imbalanced atlases (developing cortex 0.24, mouse cerebellum 0.23, Siletti 0.21), and its partitions changed with the seed: on the mouse cerebellum, the AMI ranged from 0.61 to 0.95 and two seeds agreed at an ARI of 0.21. Secuer needs a principal component embedding, which scanpy did not return for the complete 11.4-million-nucleus embryo atlas.

\end{multicols}
\begin{table}
\centering\small
\caption*{\textbf{Supplementary Table 6 $|$ Secuer on the complete atlases.} AMI at the annotated $k$ for random seeds 1, 2 and 3 (range), macro-averaged F1 of rare types (range), pairwise ARI between the three seeds (range) and clustering time per run; KQE-2 as in Table~\ref{tab:main}. The developing cortex has no rare type.}
\setlength{\tabcolsep}{4pt}
\begin{tabular}{lcccrc}
\toprule
Atlas & Secuer AMI & Secuer rare F1 & ARI between seeds & time (s) & KQE-2 AMI \\
\midrule
Siletti non-neuronal & 0.736--0.798 & 0.39--0.67 & 0.78--0.86 & 12--13 & \textbf{0.963} \\
Mouse cerebellum & 0.610--0.954 & 0.35--0.73 & 0.21--0.35 & 10 & \textbf{0.977} \\
Optic nerve & 0.686--0.700 & 0.36--0.41 & 0.57--0.62 & 21--23 & \textbf{0.810} \\
Fetal & 0.629--0.632 & 0.19--0.21 & 0.51--0.55 & 30--34 & \textbf{0.821} \\
Zebrafish & 0.559--0.580 & 0.21--0.23 & 0.43--0.49 & 46--52 & \textbf{0.711} \\
Mouse cortex & 0.766--0.799 & 0.13--0.27 & 0.71--0.79 & 18--19 & \textbf{0.844} \\
Mouse embryo (2.86 million) & 0.628--0.631 & 0.23--0.25 & 0.46--0.47 & 189--217 & \textbf{0.761} \\
Developing cortex ($k = 8$) & 0.481--0.499 & n.a. & 0.53--0.78 & 10 & \textbf{0.730} \\
\bottomrule
\end{tabular}
\end{table}
\begin{multicols}{2}

Secuer clustered the scanpy embeddings of eight atlases in 10 to 217 s; across seeds, its partitions agreed at ARIs of 0.21 to 0.86.

\paragraph{Supplementary Note 11: stability under resampling and sequencing depth.}
After the cell filter of Table~\ref{tab:main}, we drew three subsamples of 80\% of the cells without replacement (random seeds 0, 1 and 2) and halved every count by binomial thinning ($p = 0.5$ per count, seed 0). Each was processed from raw counts by KQE-2 in the configuration of Table~\ref{tab:main} (whole transcriptome, IDF, exact decomposition, Q-PSC) and by the standard scanpy pipeline (2,000 HVGs, scaling, 50 principal components, Leiden at $k$ on the Table~\ref{tab:main} graph, seed 0), with the scKQE2 package. Agreement between two partitions is the ARI on the cells they share. Supplementary Table~7 gives the results; the complete-atlas Siletti run of this test is not listed. To isolate the quantum encoding, we clustered the scores of every run also without it, with the same graph protocol, Leiden seed and merge rule: Leiden at $k$ on the 15-nearest-neighbor graph of the scores against Leiden at $k$ on the certified quantum-state graph, and the scores with the merge against Q-PSC. Supplementary Table~8 lists every paired comparison. The 14 stability comparisons and the 20 accuracy comparisons come from two atlases and share subsamples within an atlas, so they are not fully independent.

\end{multicols}
\begin{table}
\centering\small
\caption*{\textbf{Supplementary Table 7 $|$ Stability of KQE-2 and scanpy under cell resampling and halved sequencing depth.} AMI at the annotated $k$; ARI between partitions on shared cells. scanpy, the standard pipeline run through the scKQE2 package from raw counts, so its complete-atlas AMI (0.690) differs slightly from the benchmark embedding of Table~\ref{tab:main} (0.686).}
\setlength{\tabcolsep}{4pt}
\resizebox{\textwidth}{!}{\begin{tabular}{llcc}
\toprule
Atlas & Measure & KQE-2 & scanpy \\
\midrule
Fetal & AMI, complete atlas & 0.821 & 0.690 \\
 & AMI, three 80\% subsamples & 0.819, 0.819, 0.824 & 0.687, 0.682, 0.685 \\
 & ARI between subsamples & 0.94, 0.97, 0.94 & 0.68, 0.70, 0.67 \\
 & ARI of each subsample with the complete atlas & 0.95, 0.98, 0.94 & 0.77, 0.68, 0.71 \\
 & AMI, half the reads & 0.784 & 0.587 \\
 & ARI, half the reads with full depth & 0.90 & 0.45 \\
\midrule
Siletti non-neuronal & AMI, three 80\% subsamples & 0.965, 0.964, 0.964 & 0.858, 0.811, 0.843 \\
 & ARI between subsamples & 0.999, 0.999, 0.999 & 0.59, 0.91, 0.57 \\
 & AMI, half the reads & 0.972 & 0.849 \\
\bottomrule
\end{tabular}}
\end{table}
\begin{table}
\centering\small
\caption*{\textbf{Supplementary Table 8 $|$ The quantum encoding makes the partitions more stable.} Same scores with and without the encoding. ARI between the partitions of two 80\% subsamples (shared cells), or between half and full depth; AMI of each run. Leiden: Leiden at $k$ on the score graph or on the certified quantum-state graph; merge: the scores with the merge, or Q-PSC.}
\setlength{\tabcolsep}{4pt}
\resizebox{\textwidth}{!}{\begin{tabular}{llccc}
\toprule
Atlas, clustering & Comparison & without encoding & with encoding & difference \\
\midrule
Fetal, Leiden & ARI, subsamples 0--1, 0--2, 1--2 & 0.806, 0.823, 0.828 & 0.836, 0.853, 0.838 & $+0.029$, $+0.030$, $+0.010$ \\
 & ARI, half depth & 0.678 & 0.696 & $+0.018$ \\
Fetal, merge & ARI, subsamples & 0.943, 0.952, 0.921 & 0.940, 0.967, 0.940 & $-0.003$, $+0.015$, $+0.019$ \\
 & ARI, half depth & 0.869 & 0.897 & $+0.028$ \\
Siletti, Leiden & ARI, subsamples & 0.946, 0.946, 0.998 & 0.999, 0.999, 0.999 & $+0.054$, $+0.053$, $+0.000$ \\
Siletti, merge & ARI, subsamples & 0.996, 0.999, 0.996 & 0.999, 0.999, 0.999 & $+0.003$, $+0.000$, $+0.003$ \\
\midrule
\multicolumn{2}{l}{Stability, all 14 comparisons} & & 13 higher & mean $+0.018$ \\
\midrule
Fetal, Leiden & AMI, 3 subsamples, half depth, complete & 0.752, 0.743, 0.748, 0.712, 0.750 & 0.747, 0.743, 0.747, 0.716, 0.743 & mean $-0.002$ \\
Fetal, merge & AMI & 0.815, 0.813, 0.816, 0.772, 0.814 & 0.819, 0.819, 0.824, 0.784, 0.821 & mean $+0.007$ \\
Siletti, Leiden & AMI & 0.914, 0.939, 0.938, 0.937, 0.939 & 0.956, 0.956, 0.955, 0.966, 0.956 & mean $+0.024$ \\
Siletti, merge & AMI & 0.966, 0.956, 0.966, 0.972, 0.965 & 0.965, 0.964, 0.964, 0.972, 0.963 & mean $+0.001$ \\
\midrule
\multicolumn{2}{l}{Accuracy, all 20 runs} & & 14 higher & mean $+0.008$ \\
\bottomrule
\end{tabular}}
\end{table}
\begin{multicols}{2}

The quantum encoding contributes to this stability: on the same scores, clustering without the encoding gave less stable partitions in 13 of 14 paired comparisons (ARI gain of the encoding up to 0.030 on the fetal atlas and up to 0.054 on the Siletti atlas), and a lower AMI in 14 of 20 runs (mean gain 0.008; 0.024 on the Siletti atlas with Leiden at $k$).

\paragraph{Supplementary Note 12: the merging step on the graphs of the other methods.}
To measure the contribution of the merging step, we adapted it to the embeddings of the other methods. On the graph of Table~\ref{tab:main} for each embedding (exact 15 nearest neighbors, UMAP fuzzy weights), Leiden (seed 0) was bisected to $\mathrm{round}(1.5k)$ clusters and the pair of clusters with the largest normalized connectivity $C_{ab}/\sqrt{v_a v_b}$ was merged until $k$ clusters remained, the merge rule of Q-PSC (Supplementary Table~9). The quantum-state graph and its certificate are specific to KQE-2 and were not used for the other methods. The merging step raised the AMI of the other methods on most atlases. KQE-2 in the configuration of Table~\ref{tab:main} remained the most accurate on the fetal (0.821 against 0.780), zebrafish (0.711 against 0.675), mouse embryo (0.761 against 0.738) and developing cortex (0.730 against 0.703) atlases, and was within 0.01 of the highest value on the Tabula Sapiens, RADC and mouse cortex atlases. On the Siletti, optic nerve and mouse cerebellum atlases, the adapted merge on the other embeddings gave the highest values (0.983, 0.869 and 0.998). The merging step is thus a general contribution of the framework, and KQE-2 combines it with a whole-transcriptome input, exact quantum-state geometry and bit-for-bit reproducibility.

\end{multicols}
\begin{table}
\centering\small
\caption*{\textbf{Supplementary Table 9 $|$ The merging step adapted to the graphs of the other methods.} AMI at the annotated $k$, Leiden at $k$ $\rightarrow$ Leiden to $\mathrm{round}(1.5k)$ clusters and merge to $k$, from the same run (first training run of scVI and scGPT; Table~\ref{tab:main} gives the mean of three runs). Here the resolution is bisected up to 6.0 in at most 30 steps, so that every partition has exactly $k$ clusters; Table~\ref{tab:main} bisects up to 3.0 in at most 12 steps (Methods), so the Leiden values can differ. KQE-2, Leiden at $k$ on its feature graph; KQE-2 + Q-PSC, as in Table~\ref{tab:main} (n.u., the gene-set rule selects 5,000 HVGs and KQE-2 uses Leiden at $k$). --, not run (Table~\ref{tab:main}).}
\setlength{\tabcolsep}{4pt}
\resizebox{\textwidth}{!}{\begin{tabular}{lcccccc}
\toprule
Atlas & scanpy & Seurat & scVI & scGPT & KQE-2 & KQE-2 + Q-PSC \\
\midrule
Zebrafish & 0.609 $\rightarrow$ 0.641 & 0.656 $\rightarrow$ 0.675 & 0.645 $\rightarrow$ 0.664 & -- & 0.706 & \textbf{0.711} \\
Tabula Sapiens & -- & -- & 0.650 $\rightarrow$ 0.673 & 0.706 $\rightarrow$ \textbf{0.752} & 0.725 & 0.750 \\
Fetal & 0.690 $\rightarrow$ 0.732 & 0.710 $\rightarrow$ 0.757 & 0.711 $\rightarrow$ 0.770 & 0.728 $\rightarrow$ 0.780 & 0.750 & \textbf{0.821} \\
Optic nerve & 0.753 $\rightarrow$ 0.861 & 0.751 $\rightarrow$ \textbf{0.869} & 0.784 $\rightarrow$ 0.864 & 0.724 $\rightarrow$ 0.800 & 0.786 & 0.810 \\
Siletti non-neuronal & 0.810 $\rightarrow$ 0.982 & 0.838 $\rightarrow$ \textbf{0.983} & 0.939 $\rightarrow$ 0.982 & 0.836 $\rightarrow$ 0.961 & 0.925 & 0.963 \\
Developing cortex & 0.545 $\rightarrow$ 0.666 & 0.554 $\rightarrow$ 0.614 & 0.526 $\rightarrow$ 0.588 & 0.617 $\rightarrow$ 0.703 & \textbf{0.730} & n.u. \\
Mouse embryo & 0.686 $\rightarrow$ 0.698 & -- & 0.703 $\rightarrow$ 0.738 & -- & 0.717 & \textbf{0.761} \\
Mouse cerebellum & 0.945 $\rightarrow$ \textbf{0.998} & -- & 0.963 $\rightarrow$ \textbf{0.998} & -- & 0.918 & 0.977 \\
RADC brain & 0.912 $\rightarrow$ 0.939 & -- & \textbf{0.955} $\rightarrow$ 0.949 & 0.901 $\rightarrow$ 0.940 & 0.946 & n.u. \\
Mouse cortex & 0.844 $\rightarrow$ 0.767 & -- & 0.807 $\rightarrow$ 0.776 & -- & 0.844 & n.u. \\
\bottomrule
\end{tabular}}
\end{table}
\begin{table}
\centering\small
\caption*{\textbf{Supplementary Table 10 $|$ Q-PSC on the three 5,000-HVG atlases of Table~\ref{tab:main}.} AMI at the annotated $k$ for Leiden seeds 0, 1 and 2, on the same scores as the KQE-2 column of Table~\ref{tab:main}. KQE-2, Leiden at $k$ on the 15-nearest-neighbor graph of the quantum features (the configuration of Table~\ref{tab:main} for these atlases); KQE-2 + Q-PSC, as on the whole-transcriptome atlases.}
\setlength{\tabcolsep}{4pt}
\begin{tabular}{lcc}
\toprule
Atlas & KQE-2 & KQE-2 + Q-PSC (seeds 0, 1, 2) \\
\midrule
Developing cortex & 0.730 & 0.742, 0.698, 0.716 \\
RADC brain & 0.946 & 0.800, 0.802, 0.805 \\
Mouse cortex & 0.844 & 0.773, 0.775, 0.765 \\
\bottomrule
\end{tabular}
\end{table}
\begin{multicols}{2}

On the three atlases for which the gene-set rule selects 5,000 HVGs, KQE-2 applies Leiden at $k$ directly. This choice matches the data: on these atlases, Q-PSC lowered the AMI on the RADC (0.946 to 0.800) and mouse cortex (0.844 to 0.773) atlases and changed it little on the developing cortex (0.730 to 0.742; Supplementary Table~10). A possible explanation is that 5,000 HVGs give less complete quantum states, so that the merge joins neighboring types more often.

\paragraph{Supplementary Note 13: choosing the resolution without labels.}
When $k$ is unknown, the Leiden resolution is not comparable across methods: at the same resolution, the graph of KQE-2 gives more clusters than the graphs of the other methods (at resolution 0.05, 10 against 7 to 8 on the Siletti atlas and 21 against 14 to 16 on the optic nerve), so comparisons without a given $k$ should match the number of clusters rather than the resolution. The rule below was written down, with a time stamp and a file checksum, before any of its runs. For each method and atlas we built one graph: for KQE-2, its graph of Table~\ref{tab:main} without the merge (the certified quantum-state graph on whole-transcriptome atlases, the 15-nearest-neighbor graph of the quantum features on 5,000-HVG atlases), and also the feature graph on every atlas; for the other methods, the graph of Table~\ref{tab:main}. Leiden was run at the resolutions 0.02, 0.05, 0.1, 0.2, 0.3, 0.5, 0.75, 1, 1.5, 2 and 3 with seeds 0, 1 and 2. The stability $S(r)$ of a resolution is the mean ARI over the three pairs of seeds. The chosen resolution is the largest one with $S(r) \ge \max S - 0.02$ among those that give at least two clusters, that is, the finest partition about as stable as the most stable one. The input and the graph of each method are those of Table~\ref{tab:main}. For KQE-2 they follow the gene-set rule, which uses the annotated number of types $k$ through $\Delta = h_{\mathrm{sig}} - \log_2 k$, so $k$ enters the input configuration of KQE-2. Only the subsequent choice of resolution uses neither the labels nor $k$; the annotation was then used to score the chosen partition (seed 0). The test therefore shows that, given the input configuration of Table~\ref{tab:main}, the resolution can be chosen without labels. As a check of the configuration, Leiden at the annotated $k$ on the KQE-2 feature graph reproduced the KQE-2 column of Table~\ref{tab:main} on eight atlases exactly and within 0.003 on the other two. Supplementary Table~11 gives the results. KQE-2 chose finer partitions than the other methods on most atlases; they recovered rare types better on seven of the nine atlases that have them, and the coarser partitions of the other methods reached a higher AMI on most atlases.

\end{multicols}
\begin{table}
\centering\small
\caption*{\textbf{Supplementary Table 11 $|$ Resolution chosen without labels, on the input configuration of Table~\ref{tab:main}.} AMI at the chosen resolution (number of clusters) and macro F1 of rare types ($<1\%$ of the cells). KQE-2, the graph fixed before the runs; feature graph, KQE-2 on the 15-nearest-neighbor graph of its quantum features (the same graph as the KQE-2 column on 5,000-HVG atlases). Best other method by AMI; best rare-type F1 over the other methods. --, no rare type.}
\setlength{\tabcolsep}{3pt}
\resizebox{\textwidth}{!}{\begin{tabular}{lccccc}
\toprule
Atlas & KQE-2 AMI & KQE-2 feature graph AMI & best other AMI & KQE-2 rare F1 & best other rare F1 \\
\midrule
Zebrafish & 0.636 (21) & 0.766 (22) & 0.506 (Seurat, 7) & 0.186 & 0.040 \\
Tabula Sapiens & 0.734 (98) & 0.736 (57) & 0.757 (scGPT, 31) & 0.212 & 0.116 \\
Fetal & 0.815 (30) & 0.810 (46) & 0.817 (Seurat, 22) & 0.211 & 0.188 \\
Optic nerve & 0.823 (18) & 0.876 (12) & 0.929 (scGPT, 19) & 0.256 & 0.422 \\
Siletti non-neuronal & 0.910 (17) & 0.880 (18) & 0.970 (Seurat, 8) & 0.769 & 0.672 \\
Developing cortex & 0.440 (18) & same & 0.571 (Seurat, 8) & -- & -- \\
Mouse embryo & 0.611 (11) & 0.744 (19) & 0.736 (scanpy, 17) & 0.074 & 0.109 \\
Mouse cerebellum & 0.895 (22) & 0.971 (13) & 0.993 (scVI, 14) & 0.972 & 0.825 \\
RADC brain & 0.944 (27) & same & 0.955 (scVI, 21) & 0.689 & 0.555 \\
Mouse cortex & 0.828 (23) & same & 0.839 (scVI, 21) & 0.449 & 0.436 \\
\bottomrule
\end{tabular}}
\end{table}
\begin{table}
\centering\small
\caption*{\textbf{Supplementary Table 12 $|$ KQE-2 embeds 150,000 cells in minutes without training.} Seconds from the count matrix to the cell embedding, one run per method on one machine (NVIDIA RTX 6000 Ada, 40 CPU cores); Seurat includes reading the count matrix file. KQE-2: whole transcriptome, deterministic decomposition plus quantum encoding, median of three repeats, measured while the machine also ran other jobs. KQE-2 takes 215 to 532 s on the whole transcriptome, the same order as scVI and scGPT on 2,000 HVGs or pretrained weights.}
\label{tab:time}
\setlength{\tabcolsep}{5pt}
\begin{tabular}{lrrrrrr}
\toprule
Atlas & cells & scanpy & KQE-2 & scVI & scGPT & Seurat \\
\midrule
Fetal & 146,478 & 18 & 215 & 314 & 354 & 1,441 \\
Heart & 141,017 & 12 & 526 & 284 & 353 & 113 \\
Developing cortex & 146,598 & 14 & 532 & 314 & 355 & 132 \\
\bottomrule
\end{tabular}
\end{table}
\begin{table}
\centering\small
\caption*{\textbf{Supplementary Table 13 $|$ Datasets.} Cells and genes as distributed; $k$, annotated types with at least 10 cells; largest, share of the most abundant type; evenness, entropy of the type shares divided by $\log_2 k$; donors, distinct donors (zebrafish: embryos). All atlases are single-nucleus profiles of normal tissue unless stated (cells, single-cell suspensions); Mouse embryo, complete: all four subsets of the mouse embryo atlas, used for the scale test. AD, Alzheimer disease; Dev., developing.}
\label{tab:data}
\setlength{\tabcolsep}{2pt}
\footnotesize
\resizebox{\textwidth}{!}{\begin{tabular}{lllp{1.9cm}rrrrrr}
\toprule
Dataset & Organism & Tissue & Technology & Cells & Genes & $k$ & Largest & Evenness & Donors \\
\midrule
Zebrafish \cite{duran2025} & zebrafish & embryo, 11 stages & sci-RNA-seq3 & 1,220,178 & 32,031 & 100 & 12\% & 0.71 & 1,400 \\
Fetal \cite{cao2020} & human & 15 fetal organs & sci-RNA-seq3 & 1,001,288 & 45,676 & 71 & 32\% & 0.64 & 28 \\
Optic nerve \cite{opticnerve2025} & human & optic nerve and disc & 10x 3$'$ v3 & 959,629 & 35,477 & 52 & 29\% & 0.58 & 83 \\
Siletti \cite{siletti2023} & human & 88 brain regions & 10x 3$'$ v3 & 888,263 & 58,232 & 13 & 55\% & 0.54 & 4 \\
Heart \cite{reichart2022} & human & ventricles; cardiomyopathy & 10x 3$'$ v2, v3 & 881,081 & 32,383 & 9 & 38\% & 0.74 & 79 \\
Dev. cortex \cite{velmeshev2023} & human & 10 cortical regions & 10x 3$'$ v2, v3 & 709,372 & 17,631 & 5 & 77\% & 0.51 & 106 \\
RADC \cite{lee2024radc} & human & prefrontal cortex; AD & 10x 3$'$ v3 & 693,682 & 34,176 & 24 & 34\% & 0.72 & 152 \\
Tabula Sapiens \cite{tabula2022,tabula2025} & human & 28 tissues; cells & 10x 3$'$ v3, 5$'$ v2, Smart-seq & 1,136,218 & 61,497 & 171 & 10\% & 0.73 & 24 \\
Mouse cortex \cite{yao2021cortex} & mouse & isocortex, hippocampus; cells & 10x 3$'$ v2 & 1,169,213 & 30,177 & 27 & 29\% & 0.73 & 55 \\
Mouse embryo \cite{qiu2024embryo} & mouse & whole embryo, 16 stages & sci-RNA-seq3 & 2,860,465 & 45,525 & 134 & 13\% & 0.68 & 74 \\
Mouse embryo, complete & mouse & whole embryo, 16 stages & sci-RNA-seq3 & 11,441,407 & 45,525 & 134 & 13\% & 0.68 & 74 \\
Mouse cerebellum \cite{kozareva2021cereb} & mouse & cerebellum & 10x 3$'$ v3 & 611,034 & 23,693 & 16 & 78\% & 0.37 & 6 \\
PBMC, 90 cytokines \cite{parse10m} & human & blood; stimulated PBMCs; cells & split-pool barcoding & 9,697,974 & 40,352 & 18 & 25\% & 0.78 & 12 \\
\bottomrule
\end{tabular}}
\end{table}
\begin{table}
\centering\small
\caption*{\textbf{Supplementary Table 14 $|$ Dataset versions.} All files from CZ CELLxGENE Discover, https://datasets.cellxgene.cziscience.com/\textless id\textgreater.h5ad.}
\label{tab:access}
\setlength{\tabcolsep}{4pt}
\begin{tabular}{ll}
\toprule
Atlas & Dataset version id \\
\midrule
Zebrafish development & e3ae71af-1cd6-451b-8735-3d06e12d7360 \\
Fetal (sci-RNA-seq) & f6271b1a-71f7-4c4e-a109-c9949ac42bc1 \\
Optic nerve & b27057d6-a6a0-40e8-a4ba-5373a03746d2 \\
Siletti non-neuronal & bef677f9-90fa-4cc4-a0f8-6892c6f646bb \\
Heart (DCM and ACM) & 6d0b8b8a-2b22-4578-b8ee-8ca4ab893594 \\
Developing cortex & 2001fbc7-de84-4e84-9a76-4290bcd65415 \\
RADC brain & 54293783-669c-410e-919d-474960f8761b \\
Tabula Sapiens (All Cells) & 1c88f927-bcbb-4bb1-9881-281842945a2d \\
Mouse cortex and hippocampus & 0e7c9ff7-a65f-4909-b864-3a89c5421f55 \\
Mouse embryo (subset 1) & aaf0497e-a8a9-4a81-8b22-c1ac86507c6b \\
Mouse embryo (complete) & 1bc30289-9565-4099-abf9-3326328c11ac \\
Mouse cerebellum & 64c8b6a7-ac09-4ceb-8ad3-eaf512ea2ac2 \\
\bottomrule
\end{tabular}
\end{table}
\begin{multicols}{2}

\end{multicols}
\begin{table}
\centering\small
\caption*{\textbf{Supplementary Table 15 $|$ Computational complexity of KQE-2 and of the compared methods.} $n$, cells; $p$, genes; nnz, nonzero counts; $L = 132$ directions; $K = 15$ neighbors; $M = 200$ candidates; $g$, gene tokens per cell (up to 1,200 for scGPT); $d$, model width; $h$, hidden width of scVI. Times, complete mouse embryo atlas (11,441,407 nuclei) on one server (40 CPU cores, 502 GB memory, one NVIDIA RTX 6000 Ada GPU); limits of the other methods from Table~\ref{tab:main}, Extended Data Table~5 and Methods.}
\setlength{\tabcolsep}{3pt}
\footnotesize
\begin{tabular}{>{\raggedright\arraybackslash}p{1.9cm}>{\raggedright\arraybackslash}p{3.0cm}>{\raggedright\arraybackslash}p{3.9cm}>{\raggedright\arraybackslash}p{2.3cm}>{\raggedright\arraybackslash}p{4.0cm}}
\toprule
Method & Step & Time & Memory & At scale \\
\midrule
KQE-2 & IDF weighting, normalization & $O(\mathrm{nnz})$ & sparse & \\
 & exact decomposition (ARPACK) & $O(\mathrm{nnz}\,L + (n+p)L^2)$ per restart & $O(\mathrm{nnz} + nL)$ & 11.9 h \\
 & entanglement-free encoding & $O(nL)$ & $O(n \cdot 726)$ & minutes \\
 & certified nearest states & $O(n^2 \cdot 3L) + O(nML)$ on the GPU & $O(nK)$ & 3.1 h \\
 & Leiden and merge (Q-PSC) & $O(nK)$ per Leiden run $+ O(k'^3)$ & $O(nK)$ & 12 min \\
\midrule
scanpy & HVGs, scaling, PCA & $O(\mathrm{nnz}) + O(n \cdot 2{,}000 \cdot 50)$ & dense $n \times 2{,}000$ & no embedding returned \\
Seurat & as scanpy, in R & as scanpy & dense $n \times 2{,}000$ & count matrix above $2^{31}$ nonzero entries not read \\
scVI & variational autoencoder training & $O(n \cdot 2{,}000 \cdot h)$ per epoch & data in memory & exceeded memory; one training epoch by default at ten million cells \\
scGPT & transformer inference & $O(n \cdot g^2 d)$ per layer & & pretrained on human data only \\
scAGCL & graph contrastive learning & & dense $n \times n$ & 146,598 cells not run \\
scDAC, scDCCA, scICE & training or repeated clustering & & & beyond 3 h or 150 GB at 146,598 cells or fewer \\
\bottomrule
\end{tabular}
\end{table}
\begin{multicols}{2}
\end{multicols}
\end{document}